\pdfoutput=1

\documentclass[11pt]{article}

\usepackage[final]{acl}
\usepackage{romanbar}
\usepackage{booktabs}
\usepackage{subcaption} 
\usepackage{times}
\usepackage{latexsym}
\usepackage{amssymb}
\usepackage{amsmath}
\usepackage{caption} 
\usepackage{booktabs}
\usepackage{multirow}
\usepackage[T1]{fontenc}

\usepackage[utf8]{inputenc}

\usepackage{microtype}
\usepackage{subcaption}
\usepackage{inconsolata}

\usepackage{graphicx}

\newcommand{\vpara}[1]{\vspace{0.05in}\noindent \textbf{#1 }}

\newtheorem{definition}{\textbf{Definition}}

\title{P$^2$ Law: Scaling Law for Post-Training After Model Pruning}

\author{
    Xiaodong Chen\textsuperscript{\rm 2,3}$^*$,\ Yuxuan Hu\textsuperscript{\rm 2,3}\thanks{Xiaodong Chen and Yuxuan Hu have equal contribution.},\ Xiaokang Zhang\textsuperscript{\rm 2,3},\ Yanling Wang\textsuperscript{\rm 4} \\
    \  {\bf Cuiping Li}\textsuperscript{\rm 1,2},\ {\bf Hong Chen}\textsuperscript{\rm 1,2},\ {\bf Jing Zhang\textsuperscript{\rm 1,2}}\thanks{Corresponding author.}\\
    \textsuperscript{\rm 1}Engineering Research Center of Database and Business Intelligence, MOE, China \\
    \textsuperscript{\rm 2}School of Information, Renmin University of China,Beijing, China\\
    \textsuperscript{\rm 3}Key Laboratory of Data Engineering and Knowledge Engineering, MOE, China \\
    \textsuperscript{\rm 4} Zhipu AI, China\\
    \{chenxiaodong,huyuxuan1999,zhang-jing,zhang2718,licuiping,chong\}@ruc.edu.cn \\
    \ wangyl@zgclab.edu.cn
}

\begin{document}
\maketitle
\begin{abstract}
Pruning has become a widely adopted technique for reducing the hardware requirements of large language models (LLMs). To recover model performance after pruning, post-training is commonly employed to mitigate the resulting performance degradation. 
While post-training benefits from larger datasets, once the dataset size is already substantial, increasing the training data provides only limited performance gains. To balance post-training cost and model performance, it is necessary to explore the optimal amount of post-training data.
Through extensive experiments on the Llama-3 and Qwen-2.5 series models, pruned using various common pruning methods, we uncover the scaling \textbf{Law} for \textbf{P}ost-training after model \textbf{P}runing, referred to as the P$^2$ Law. This law identifies four key factors for predicting the pruned model’s post-training loss: the model size before pruning, the number of post-training tokens, the pruning rate, and the model’s loss before pruning. Moreover, P$^2$ Law can generalize to larger dataset sizes, larger model sizes, and higher pruning rates, offering valuable insights for the post-training of pruned LLMs.

\end{abstract}

\section{Introduction}

Large language models (LLMs) based on the Transformer architecture~\cite{2017Attention} have been applied across diverse domains and tasks. However, as LLMs grow in size, their hardware demands increase substantially, limiting their practical deployment in real-world scenarios. To address this challenge, researchers have focused on developing compact models through model pruning techniques~\cite{pruning-han2016} that maintain high performance while reducing hardware requirements.

\begin{figure}[t]
\centering
\includegraphics[width=0.45\textwidth]{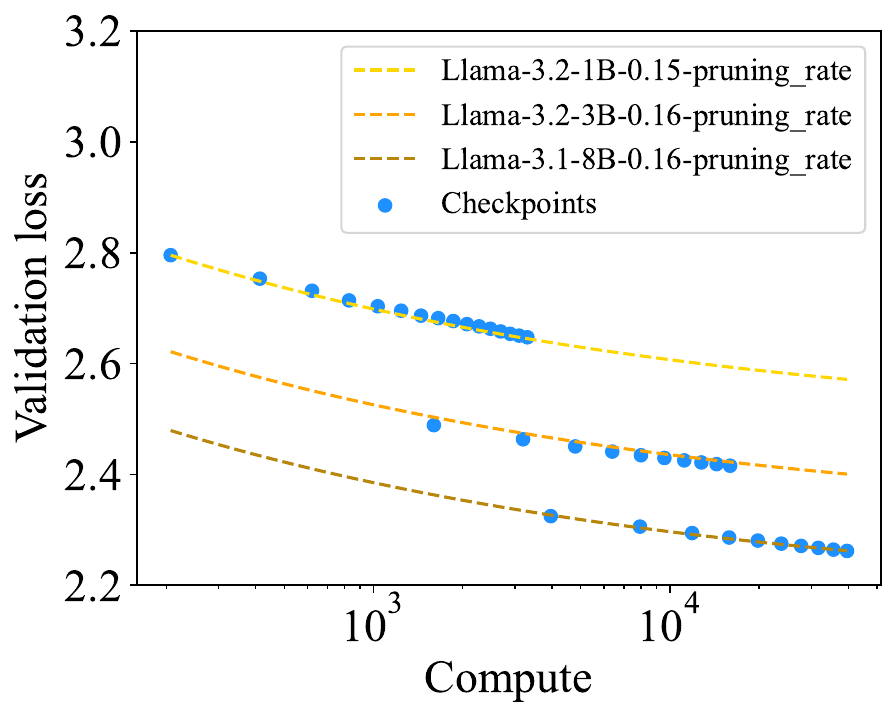}
\caption{Loss curves derived by P$^2$ Law and the actual checkpoints of Llama-3 series models pruned by depth pruning with a pruning rate of approximately 15\%. Compute ($C$) denotes the computational cost, which is calculated by $C=6ND$~\cite{Kaplan2020Scaling}, where $N$ denotes the model size after pruning, and $D$ denotes the number of post-training tokens.}
\label{fig:first}
\end{figure}

Model pruning can be broadly categorized into unstructured pruning ~\cite{sparsegpt-frantar2023, RIA-zhang2024, wanda-sun2024} and structured pruning~\cite{llm-streamline-chen2024, sp3-hu2024, pat-liu2024, compact-llm-muralidharan2024, ma2023llmpruner, slicegpt-ashkboos2024, shortgpt-men2024}. Unstructured pruning removes individual elements from weight matrices, producing sparse matrices while preserving satisfactory model performance. However, the introduced structural irregularities make this approach hardware-unfriendly and hinder its ability to accelerate computation.
To mitigate this problem, semi-structured pruning, a variant of unstructured pruning, leverages specific hardware support~\cite{2:4sparsity-mishra2021} to achieve acceleration but may result in greater performance degradation compared to unstructured pruning.
In contrast, structured pruning removes entire components, such as attention heads or layers, effectively reducing the model size but often with a higher performance drop compared to other pruning methods.

To effectively leverage hardware-friendly models pruned using semi-structured or structured pruning methods, post-training~\cite{slicegpt-ashkboos2024, llm-streamline-chen2024, yang2024laco, ma2023llmpruner, kim2024shortened} serves as an essential step after model pruning to mitigate the performance degradation.
For example, LLM-Pruner~\cite{ma2023llmpruner} utilizes 50,000 instruction data samples for fine-tuning, whereas Shortened Llama~\cite{kim2024shortened} uses 627B tokens of pre-training data for continual pre-training of the pruned LLMs. 
In general, compared to fine-tuning with a small dataset, continual pre-training with a large dataset is a more effective way to fully recover performance, but it demands substantial hardware resources.
Given the significant hardware demands, a question is raised: \textbf{is it truly necessary to use a vast amount of data for performance recovery?} 
LLM-Streamline~\cite{llm-streamline-chen2024} answers the question by demonstrating that using large amounts of data for post-training only slightly improves performance compared to using a suitably sized amount.
Hence, this raises another question: \textbf{whether a scaling law can be established to predict the optimal amount of post-training data required after model pruning for resource efficiency?}



To address the problem, we conduct pilot experiments on the Llama-3~\cite{llama3-dubey2024} and Qwen-2.5 series models~\cite{qwen2.5}, applying both typical structured and semi-structured pruning methods.
In specific, we observe several trends in the post-training loss curves, allowing us to identify the necessary conditions that the scaling \textbf{Law} for \textbf{P}ost-training after model \textbf{P}runing (P$^2$ Law) must satisfy. Building on the Chinchilla scaling law~\cite{Hoffmann2022scaling} proposed for pre-training and the identified conditions, we define multiple parameterizations of our P$^2$ Law and select the most suitable parameterization.
To assess the fit of different parameterizations to P$^2$ Law, we introduce a new metric named Average Slope Difference (ASD).
As scaling laws are used to find the suitable training data size by balancing cost and performance, focusing on the slope of the predicted loss curve rather than the predicted loss values, the ASD metric is designed to measure the slope discrepancy between predicted and actual loss curves.
Finally, P$^2$ Law is parameterized as,
\begin{equation}
\small
\begin{split}
    \mathcal{L}(N_0, D, \rho, \mathcal{L}_0) =  \mathcal{L}_0 + (\frac{1}{\rho})^{\gamma}(\frac{1}{N_0})^{\delta} 
    (\frac{N_{C}}{N_0^{\alpha} }
   + \frac{D_C}{D^{\beta}} + E) 
\end{split}
\label{Eq.scaling_law}
\end{equation}
where $N_C$, $D_C$, $E$, $\alpha$, $\beta$, $\gamma$, $\delta$ are constants, $N_0$ denotes the model size before pruning, $D$ denotes the number of post-training tokens, $\rho$ denotes the pruning rate, $\mathcal{L}_0$ denotes the model's loss before pruning, and $\mathcal{L}$ denotes the pruned model's post-training loss.

In this paper, we conduct a series of experiments to validate the P$^2$ Law.
Taking Llama-3 series models pruned by depth pruning with a pruning rate of approximately 15\% as an example, Figure~\ref{fig:first} illustrates that P$^2$ Law accurately fits the actual post-training losses of the pruned model checkpoints, where compute ($C$) represents the computational cost calculated as $C=6ND$~\cite{Kaplan2020Scaling}, $N$ is the model size after pruning, and $D$ is the number of post-training tokens.
Utilizing the post-training loss curves derived by P$^2$ Law, we can accurately predict that the computational cost required for the post-training loss of Llama-3.2-1B to start decreasing gently is approximately $10^4$. This predicted size of post-training data provides a good balance between cost and performance.
Furthermore, we evaluate the generalization ability of the P$^2$ Law, demonstrating that P$^2$ Law can effectively generalizes to larger dataset sizes, larger model sizes, and higher pruning rates.

Overall, this work makes the following contributions:
\begin{itemize}
    \item We conduct extensive studies to uncover the P$^2$ Law, the first scaling law for post-training after pruning, helping balance post-training cost and pruned LLM performance.

    \item We propose ASD, an effective metric for the evaluation of parameterizations of scaling laws for the post-training of pruned LLMs.
    
    \item  We demonstrate that the P$^2$ Law generalizes effectively to larger dataset sizes, larger models, and higher pruning rates, offering valuable insights for optimizing pruned LLMs across diverse settings.
\end{itemize}

\section{Preliminary\label{sec: Preliminary}}

In this section, we present the preliminary of this work, including various pruning methods and the post-training method.



\subsection{Pruning\label{sec: pruning}}
We utilize three common pruning methods to prune LLMs, including two structured pruning methods (depth pruning~\cite{llm-streamline-chen2024, song2024slebstreamliningllmsredundancy, gromov2024unreasonable} and width pruning~\cite{slicegpt-ashkboos2024, sp3-hu2024, pat-liu2024}) and a hardware-friendly variant of unstructured pruning method known as 2:4 semi-structured pruning~\cite{wanda-sun2024, sparsegpt-frantar2023, RIA-zhang2024}. 

\vpara{Depth Pruning.} Depth pruning is a structured pruning method that removes entire Transformer layers from LLMs. Specifically, depth pruning involves estimating the importance of each Transformer layers in LLMs and then removing those layers with the lowest importance. 

\vpara{Width Pruning.} Width pruning is another structured pruning method that reduces the number of embedding channels in LLMs. This method involves measuring the importance of embedding channels and pruning the least important ones. 

\vpara{2:4 Semi-Structured Pruning.} Unstructured pruning removes individual unimportant elements from the weight matrices, producing sparse matrices. 2:4 semi-structured pruning is a variant of unstructured pruning, with a sparse pattern of 2:4. In this pattern, every four elements in the weight matrices are grouped together, with two of the elements in each group set to zero. This semi-structured sparsity can be efficiently accelerated by hardware. We utilize SparseGPT~\cite{sparsegpt-frantar2023}, a well-known 2:4 semi-structured pruning method, to prune LLMs. 

For more details about the pruning methods used in this paper, please refer to the Appendix~\ref{sec: pruning methods}.

\subsection{Post-Training\label{sec: post-training}} 
After the pruning, we conduct post-training on the pruned LLMs to mitigate the performance decline. For LLMs pruned using depth or width pruning, we train all parameters of the pruned LLMs. For sparse LLMs derived from 2:4 semi-structured pruning, inspired by LoRS~\cite{hu2025lorsefficientlowrankadaptation}, we combine the updated weight from each training iterate with the sparse mask during the post-training process to ensure the model's sparsity, further post-training details about the 2:4 semi-structured pruning is provided in Appendix~\ref{sec: 2:4}.

\section{Experiments for Finding Necessary Conditions Satisfied by P$^2$ Law}
In this section, we conduct experiments on six LLMs from the Llama-3 and Qwen-2.5 series, covering various model sizes and using depth pruning, width pruning, and 2:4 semi-structured pruning.

First, we detail the pruning settings and post-training settings in Section~\ref{sec:setting}. Next, we describe multiple trends observed in the post-training loss curves in Section~\ref{sec: trends}. Finally, in Section~\ref{sec: conditions}, we identify several necessary conditions that the P$^2$ Law must satisfy based on the observed trends. 

\begin{figure*}[t]
    \centering
    \begin{minipage}{\linewidth}
        \centering
        \begin{subfigure}[b]{0.3\linewidth}
            \includegraphics[width=\linewidth]{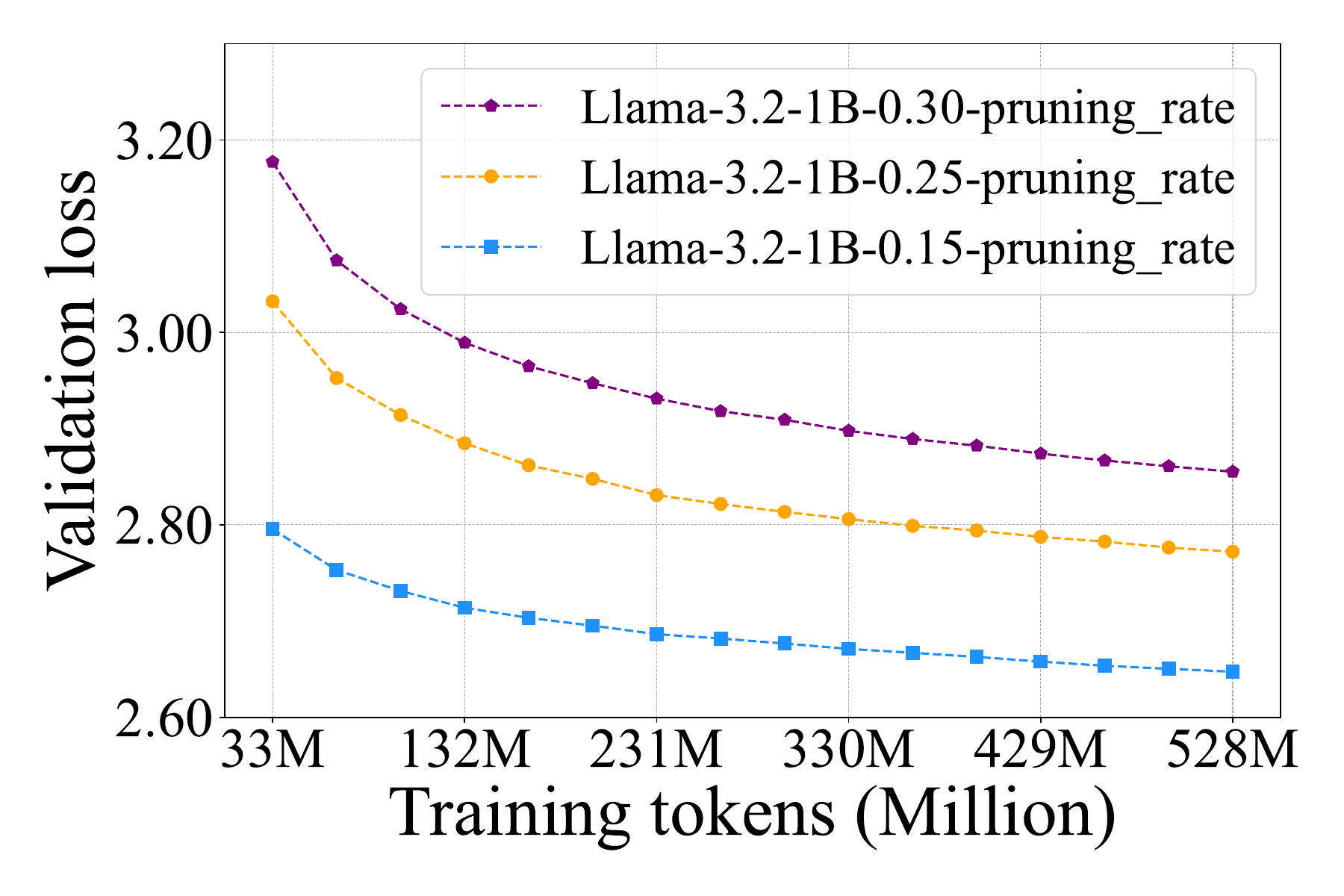}
            \caption{Post-training loss curves of Llama-3.2-1B pruned by depth pruning with different pruning rates.}
            \label{fig:1b_depth}
        \end{subfigure}
        \hfill
        \begin{subfigure}[b]{0.3\linewidth}
            \includegraphics[width=\linewidth]{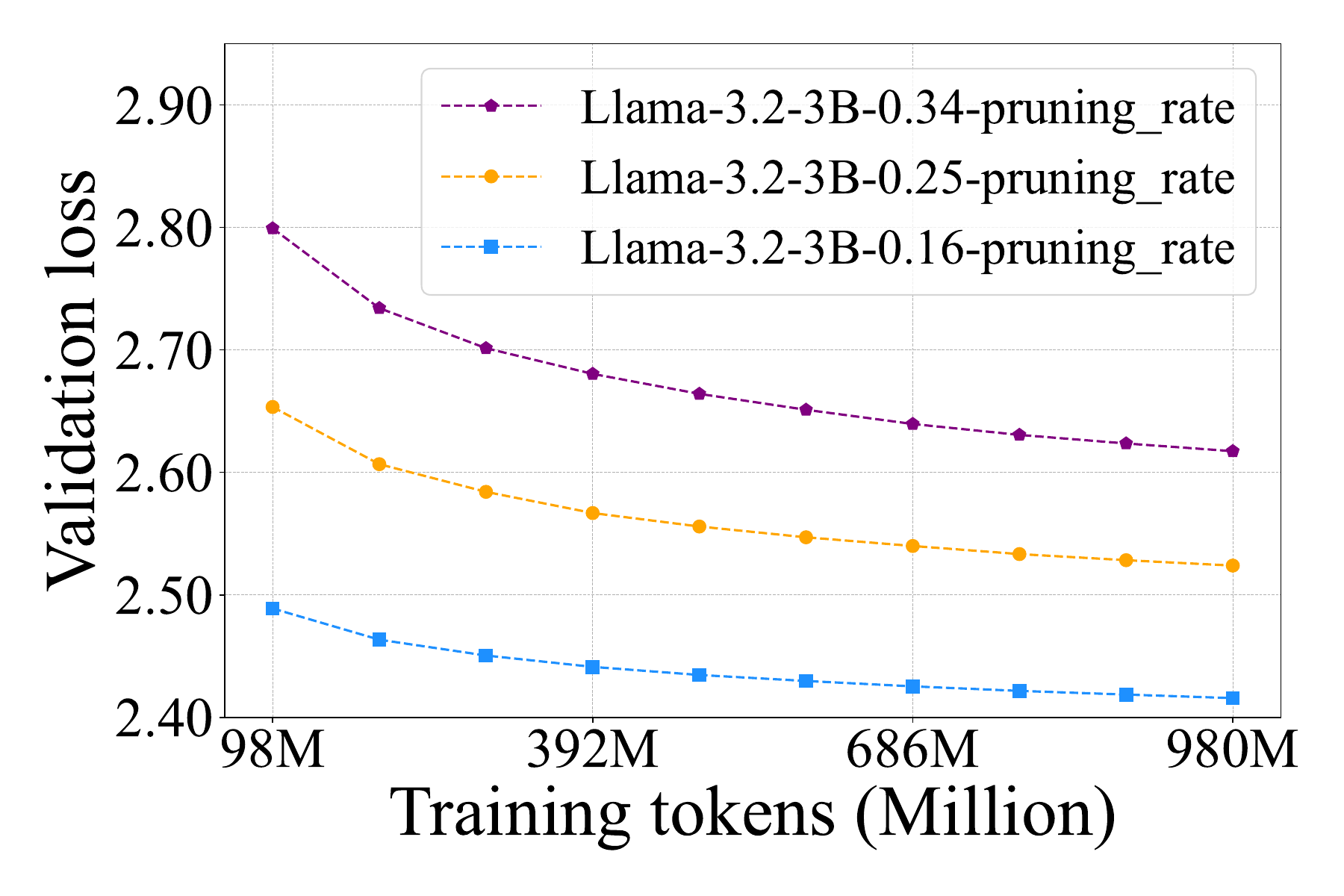}
            \caption{Post-training loss curves of Llama-3.2-3B pruned by depth pruning with different pruning rates.}
            \label{fig:3b_depth}
        \end{subfigure}
        \hfill
        \begin{subfigure}[b]{0.3\linewidth}
            \includegraphics[width=\linewidth]{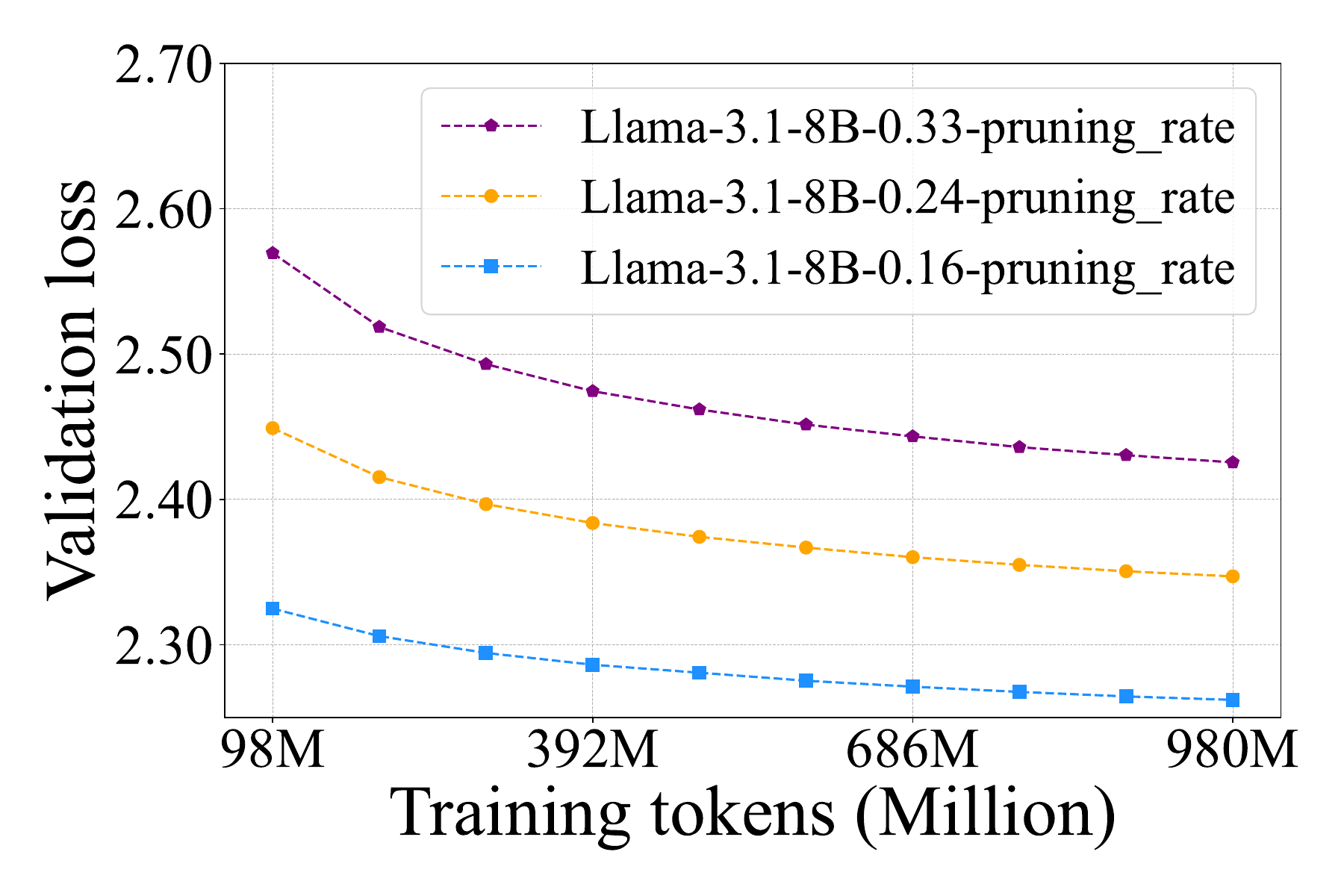}
            \caption{Post-training loss curves of Llama-3.1-8B pruned by depth pruning with different pruning rates.}
            \label{fig:8b_depth}
        \end{subfigure}
    \end{minipage}
    \caption{Post-training loss curves of Llama-3 series models pruned by depth pruning with different pruning rates.}
    \label{fig:1-8b_depth}
\end{figure*}

\begin{figure*}[t]
\centering
\begin{minipage}{0.45\textwidth}
  \centering
  \includegraphics[width=\textwidth]{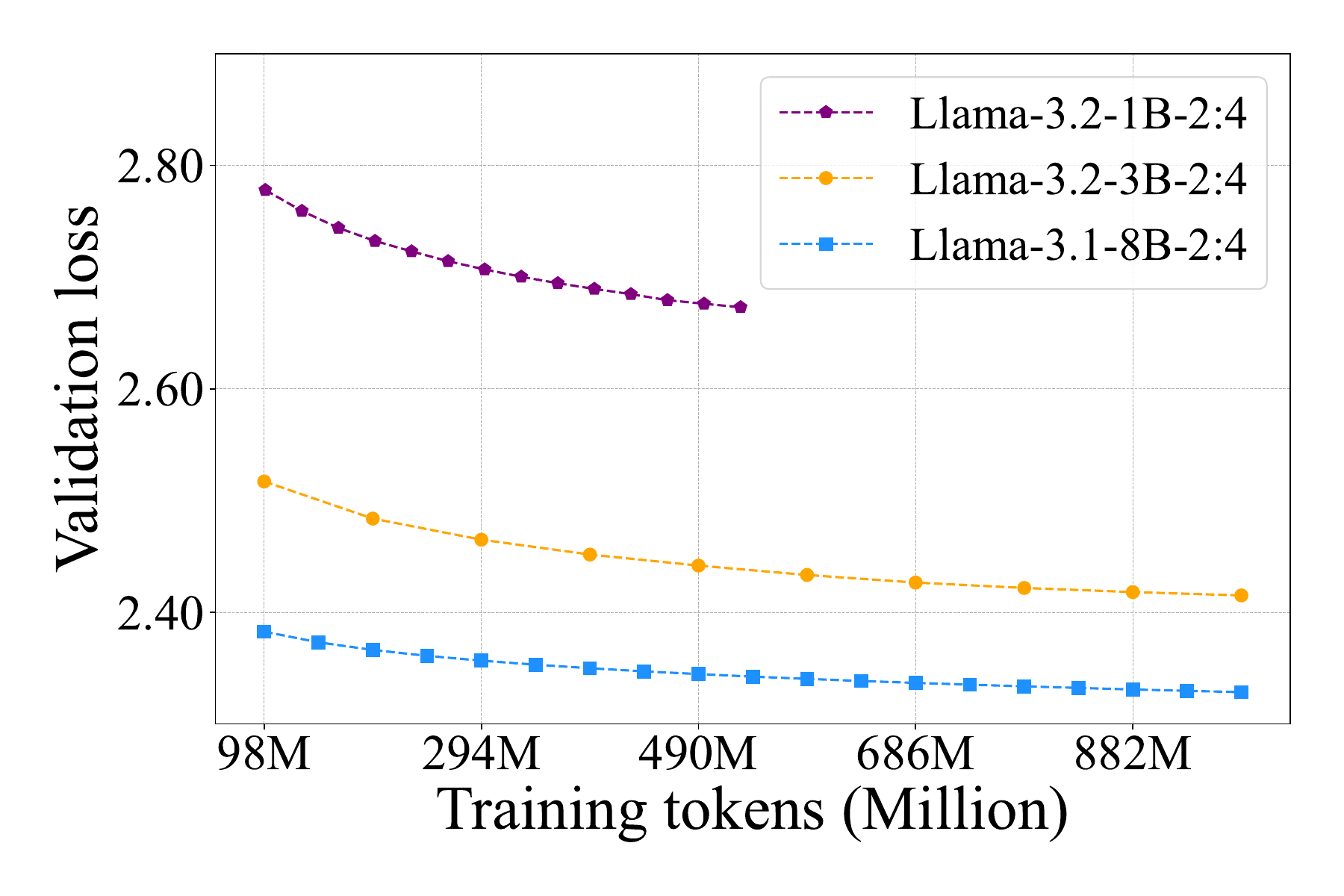}
  \caption{Post-training loss curves of Llama-3 series models pruned by 2:4 semi-structured pruning.}
  \label{fig:1_8bsemi}
\end{minipage}
\hfill
\begin{minipage}{0.45\textwidth}
  \centering
  \includegraphics[width=\textwidth]{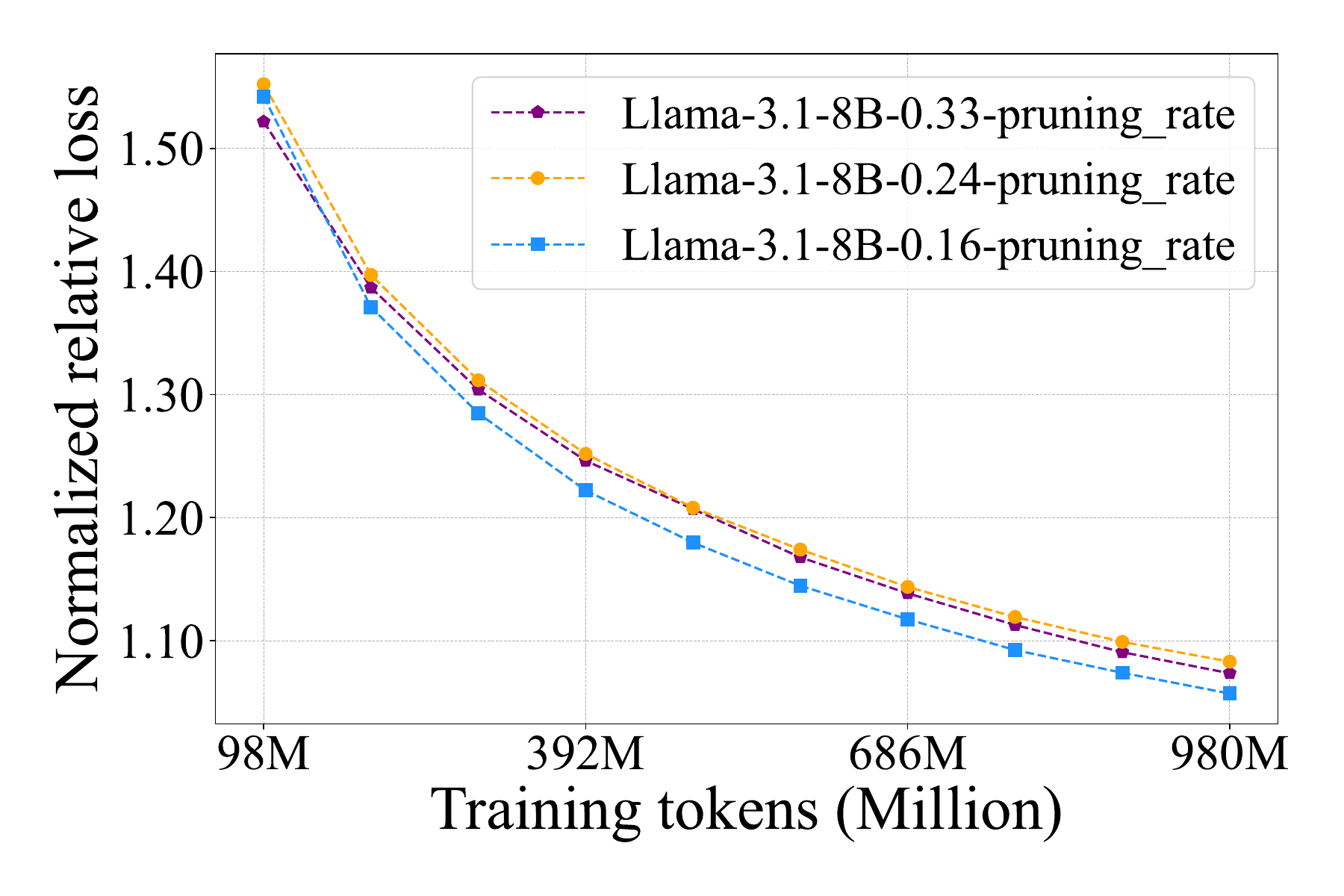}
  \caption{Normalized relative post-training loss curves of Llama-3.1-8B pruned by depth pruning.}
  \label{fig:relative loss}
\end{minipage}
\end{figure*}

\begin{table}[t]
\centering
\renewcommand\arraystretch{0.5}
\resizebox{0.9\linewidth}{!}{
\begin{tabular}{c|c|c}
\toprule
             & Depth pruning  & Width pruning  \\ \midrule
Llama-3.2-1B & 15\%,25\%,30\% & 15\%,25\%,35\% \\ \midrule
Llama-3.2-3B & 16\%,25\%,34\% & 15\%,25\%,35\% \\ \midrule
Llama-3.1-8B & 16\%,24\%,33\% & 15\%,25\%,35\% \\ \midrule
Qwen-2.5-0.5B& 15\%,21\%,27\% & 15\%,25\%,35\% \\ \midrule
Qwen-2.5-1.5B& 15\%,24\%,33\% & 15\%,25\%,35\% \\ \midrule
Qwen-2.5-3B  & 17\%,25\%,32\% & 15\%,25\%,35\% \\ \bottomrule
\end{tabular}
 }
	\caption{Pruning rates used for depth pruning and width pruning on different LLMs.}
	\label{tab: pruning_rates}
\end{table}

\subsection{Settings\label{sec:setting}}
We conduct experiments on six LLMs from the Llama-3 and Qwen-2.5 series, including Llama-3.2-1B, Llama-3.2-3B, Llama-3.1-8B, Qwen-2.5-0.5B, Qwen-2.5-1.5B and Qwen-2.5-3B.

\vpara{Pruning.} 
The pruning rates used for depth pruning and width pruning are shown in Table~\ref{tab: pruning_rates}. 
The pruning processes have been introduced in Appendix~\ref{sec: pruning methods}.
We randomly select 1,024 samples from the pre-training dataset SlimPajama for pruning.



\vpara{Post-Training.} For Llama-3.2-3B, Qwen-2.5-3B and Llama-3.1-8B, we randomly select 1B tokens from SlimPajama for post-training. For Llama-3.2-1B, Qwen-2.5-0.5B, Qwen-2.5-1.5B, we randomly select 0.5B tokens from SlimPajama for post-training. During the post-training process, we set the learning rate to 2e-5 and the batch size to 262k tokens. All post-training processes are conducted on 4 Nvidia A800-80G GPUs and 4 Nvidia A6000-48G GPUs. The entire training process takes a total of 500 hours. For more details about batch size and learning rate settings, please refer to the Appendix~\ref{sec:lr and bs}.

\subsection{Trends of the Post-Training Loss Curves\label{sec: trends}}
To better explore the trends of the post-training loss curves, we define:
\begin{definition}
\textbf{Relative post-training loss $\Delta \mathcal{L}$.} The relative post-training loss is the difference between the pruned model's post-training loss $\mathcal{L}$ and the model's loss $\mathcal{L}_0$ before pruning.
\begin{equation}
    \begin{split}
    \Delta \mathcal{L} =\mathcal{L} - \mathcal{L}_0
    \end{split}
    \end{equation}
\end{definition}

\begin{definition}
\textbf{Normalized relative post-training loss $\Delta \mathcal{L}_{norm}$.} The normalized relative post-training loss is defined as the ratio of the relative post-training loss $\Delta \mathcal{L}$ to a power-law function of the pruning rate $\rho$.
\begin{equation}
\begin{split}
    \Delta \mathcal{L}_{norm} = \frac{\Delta \mathcal{L}}{(\frac{1}{\rho})^\gamma}
\end{split}
\end{equation}
where $\gamma$ is a constant.
\end{definition}

In Figures~\ref{fig:1-8b_depth} and~\ref{fig:1_8bsemi}, we present the post-training loss curves for the Llama-3 series models pruned by depth pruning and 2:4 semi-structured pruning. 
Additional post-training loss curves (exhibiting similar trends) are shown in Figures~\ref{fig:1-8b_width},~\ref{fig:1-8b_depth_qwen},~\ref{fig:1-8b_width_qwen}, and~\ref{fig:1_8bsemi_qwen} in the Appendix~\ref{sec:addition loss curves}. 
By analyzing the post-training loss curves, we observe the following trends:
\begin{itemize}
\item \textbf{Trend 1: Smaller LLMs exhibit faster decreases in post-training loss.} For instance, as shown in Figure~\ref{fig:1_8bsemi}, with 2:4 semi-structured pruning, the post training loss curve of Llama-3.1-8B is much flatter compared to those of Llama-3.2-3B and Llama-3.2-1B. The same trend is observed under both depth pruning and width pruning, as depicted in Figure~\ref{fig:1-8b_depth} and Figure~\ref{fig:1-8b_width}.
This suggests that smaller LLMs exhibit faster decreases in post-training loss.

\item \textbf{Trend 2: Relative post-training loss $\Delta \mathcal{L}$ follows a power-law relationship with the pruning rate $\rho$.} As shown in Figure~\ref{fig:relative loss}, with depth pruning, the normalized relative post-training loss curves of Llama-3.1-8B at various pruning rates nearly overlap. This can be formally expressed as:
\begin{equation}
\begin{split}
    \frac{\Delta \mathcal{L}^{(0.33)}}{(\frac{1}{0.33})^{\gamma}} \approx \frac{\Delta \mathcal{L}^{(0.24)}}{(\frac{1}{0.24})^{\gamma}} \approx \frac{\Delta \mathcal{L}^{(0.16)}}{(\frac{1}{0.16})^{\gamma}}
\end{split}
\label{eq:pruning_rate}
\end{equation}
where $\Delta \mathcal{L}^{(0.33)}$, $\Delta \mathcal{L}^{(0.24)}$, and $\Delta \mathcal{L}^{(0.16)}$ represent the relative post-training loss of Llama-3.1-8B pruned by depth pruning with pruning rates $\rho$ of 0.33, 0.24, and 0.16, respectively. This demonstrates that the pruning rate and the relative post-training loss are governed by a power-law relationship.
\end{itemize}

\subsection{Necessary Conditions Satisfied by P$^2$ Law\label{sec: conditions}}
Based on the aforementioned trends, we identify three fundamental conditions for the P$^2$ Law:
\begin{itemize}
    \item \textbf{Condition 1.} The post-training loss $\mathcal{L}$ decreases as the number of post-training tokens $D$ increases:
    \begin{equation}
    \begin{split}
    \frac{\partial \mathcal{L}}{\partial D} < 0
    \end{split}
    \end{equation}
    \item \textbf{Condition 2.} As derived from Trend 1 in Section~\ref{sec: trends}, under similar pruning rates, the post-training loss curves of smaller LLMs decrease faster as the number of post-training tokens $D$ increases:
    \begin{equation}
    \begin{split}
    \frac{\partial}{\partial N_0}(\frac{\partial \mathcal{L}}{\partial D}) = \frac{\partial^2 \mathcal{L}}{\partial N_0 \partial D} > 0
    \end{split}
    \end{equation}
    where $N_0$ is the model size before pruning.
    \item \textbf{Condition 3.} From Eq.\ref{eq:pruning_rate} in Trend 2, the relative post-training loss $\Delta \mathcal{L}$ follows a power-law relationship with the pruning rate $\rho$: 
    \begin{equation}
    \begin{split}
    \Delta \mathcal{L} \propto (\frac{1}{\rho})^{\gamma}
    \end{split}
    \end{equation}
\end{itemize}

An ideal P$^2$ Law should satisfy aforementioned three conditions.
Additionally, the P$^2$ Law should also satisfy the condition that when the pruning rate $\rho$ is 0, the relative post-training loss $\Delta \mathcal{L}$ is 0, which is a necessary condition for Condition 3.

\section{P$^2$ Law\label{sec:scaling_law}}
In this section, we aim to parameterize the P$^2$ Law according to the above three necessary conditions.
In Section~\ref{sec: metric}, we introduce the metric for assess the quality of different candidate parametrizations. Next, based on the Chinchilla scaling law, we define multiple parameterizations for our P$^2$ Law and select the most suitable one in Section~\ref{sec: parameterizations}. Finally, in Section~\ref{sec:Generalization}, we demonstrate the generalization ability of the P$^2$ Law from three perspectives: dataset size, model size, and pruning rate.

\subsection{Metric for Accessing Law Fitting\label{sec: metric}}
Following prior work~\cite{que2024d}, we utilize both $R^2$~\cite{fisher1922mathematical} and Huber loss~\cite{huber1992robust} to evaluate different parameterizations of scaling law. The $R^2$ value, reflecting the proportion of variance explained, trends toward 1 as the fit becomes more robust. Huber loss, a robust loss function, blends the characteristics of mean squared error and mean absolute error, making it less sensitive to outliers. The Huber loss is a positive number, and a lower Huber loss suggests a better fit.

\begin{figure}[t]
\centering
\includegraphics[width=0.45\textwidth]{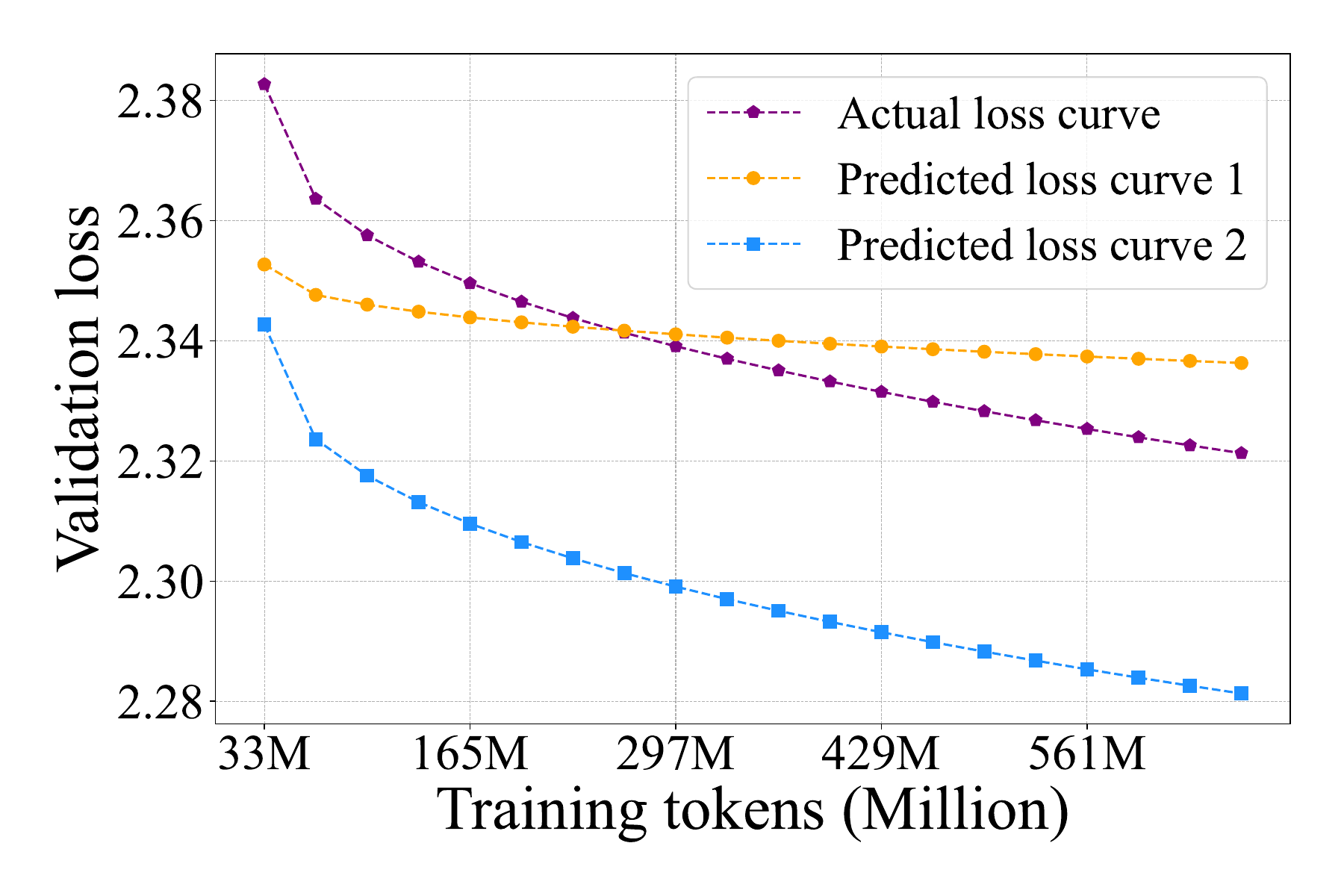}
\caption{An example showcasing the advantages of ASD. The ASD of predicted loss curve 2 is lower because its slope is closer to that of the actual loss curve.}
\label{fig:example}
\end{figure}

Scaling laws are often used to determine the optimal amount of training data by balancing computational cost and model performance.
For instance, as shown in Figure~\ref{fig:example}, there is one actual loss curve and two predicted loss curves. Traditional metrics like $R^2$ and Huber loss indicate that predicted curve 1 better matches the actual curve. However, the convergence trend predicted by curve 1 deviates significantly from the actual convergence trend. While curve 1 predicts that the loss has flattened, the actual loss continues to decrease.
On the other hand, while predicted curve 2 deviates more from the actual curve in terms of absolute values, its slope is consistently closer to the actual curve. This makes its prediction of the flattening point much more accurate.
To address this issue, we propose a new metric called Average Slope Difference (ASD), which measures the difference between the slope of the loss curve predicted by the scaling law and the slope of the actual loss curve. ASD is formally defined as:
\begin{align}
\text{ASD} = \frac{1}{N}\sum_{i=2}^{N}\left|(y_{i} - y_{i-1}) - (\hat{y_{i}} - \hat{y}_{i-1})\right|
\end{align}
where ${y_i}$ represents the loss of $N$ points uniformly sampled from the actual loss curve as the number of post-training tokens increases, and $\hat{y_{i}}$ represents the corresponding loss values on the curve predicted by the scaling law. Since the early parts of the loss curve during post-training do not represent true convergence, we only sample points from the latter half of the training process. A smaller ASD value indicates that the predicted loss curve's slope more closely matches the slope of the actual loss curve.

\begin{table*}[t]
\centering
\renewcommand\arraystretch{0.8}
\resizebox{\textwidth}{!}{
\begin{tabular}{c|c|ccc|ccc|ccc}
\toprule
 \multirow{2}{*}{LLM} & \multirow{2}{*}{Parameterizations}   & \multicolumn{3}{c|}{Depth pruning}      & \multicolumn{3}{c|}{Width pruning}      & \multicolumn{3}{c}{2:4 semi-structured pruning}        \\
 &    & $R^2$     & Huber loss & ASD      & $R^2$     & Huber loss & ASD      & $R^2$     & Huber loss & ASD      \\
 \midrule
\multirow{3}{*}{Llama-3 series}  
 & $\mathcal{L}_1$ & \textbf{0.9717} & \textbf{0.000016} & \textbf{0.000619} & \textbf{-1.2985} & \textbf{0.000177} & \textbf{0.000592} & \textbf{0.8126} & \textbf{0.000056} & \textbf{0.001466} \\
 & $\mathcal{L}_2$ & 0.9300 & 0.000045   &0.001150  &-2.5578 & 0.000450   &0.001419  & 0.7797 & 0.000079   &0.002294  \\
 & $\mathcal{L}_3$ & 0.7737 & 0.000118   &0.000827  &-4.5905 & 0.000776   &0.001754 & -0.2555 & 0.000493   &0.002054  \\
 \midrule
\multirow{3}{*}{Qwen-2.5 series} 
 & $\mathcal{L}_1$ & \textbf{0.9781} & \textbf{0.000011} & \textbf{0.000524} & \textbf{0.9891} & \textbf{0.000010} & \textbf{0.000648} & \textbf{0.9995} & \textbf{0.000000} & \textbf{0.000191} \\
 & $\mathcal{L}_2$ & 0.9423 & 0.000031   &0.000879  & 0.9803 & 0.000027   &0.000712  & 0.9867 & 0.000010   &0.000753  \\
 & $\mathcal{L}_3$ & 0.8855 & 0.000075   &0.001270  & 0.9824 & 0.000024   &0.000733 & 0.9930 & 0.000005   &0.000491 \\
 \bottomrule
\end{tabular}
}
\caption{Evaluation of three parameterizations for P$^2$ Law fitting.}
\label{tab:Evaluation of Fitting Results}
\end{table*}

\begin{figure*}[t]
    \centering
    \begin{minipage}{\linewidth}
        \centering
        \begin{subfigure}[b]{0.3\linewidth}
            \includegraphics[width=\linewidth]{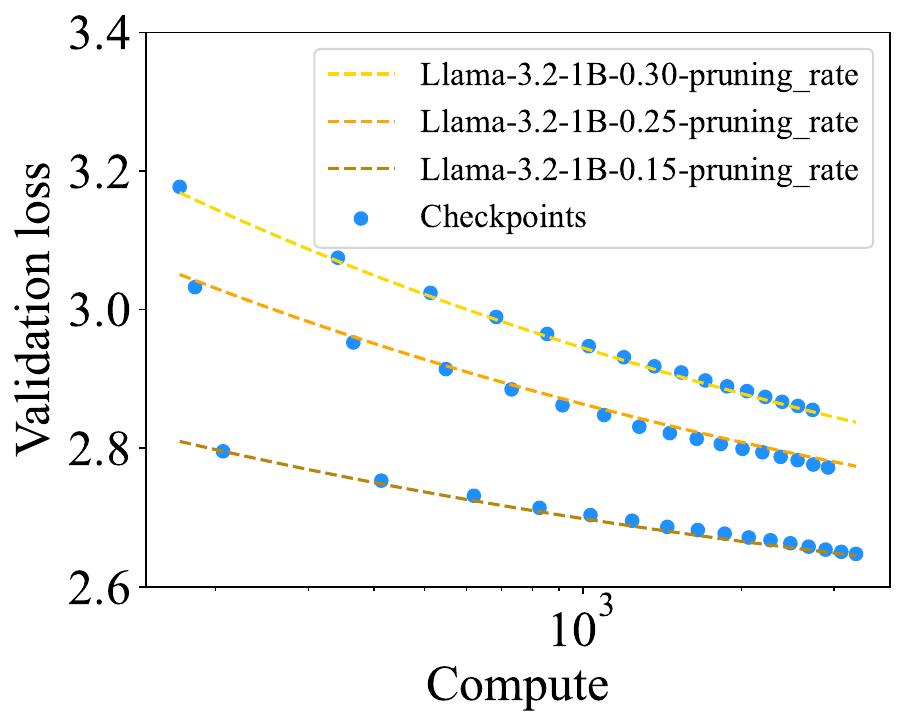}
            \caption{Loss curves derived by P$^2$ Law and the actual checkpoints of Llama-3.2-1B pruned by depth pruning.}
            \label{fig:depth_fitted_llama1b}
        \end{subfigure}
        \hfill
        \begin{subfigure}[b]{0.3\linewidth}
            \includegraphics[width=\linewidth]{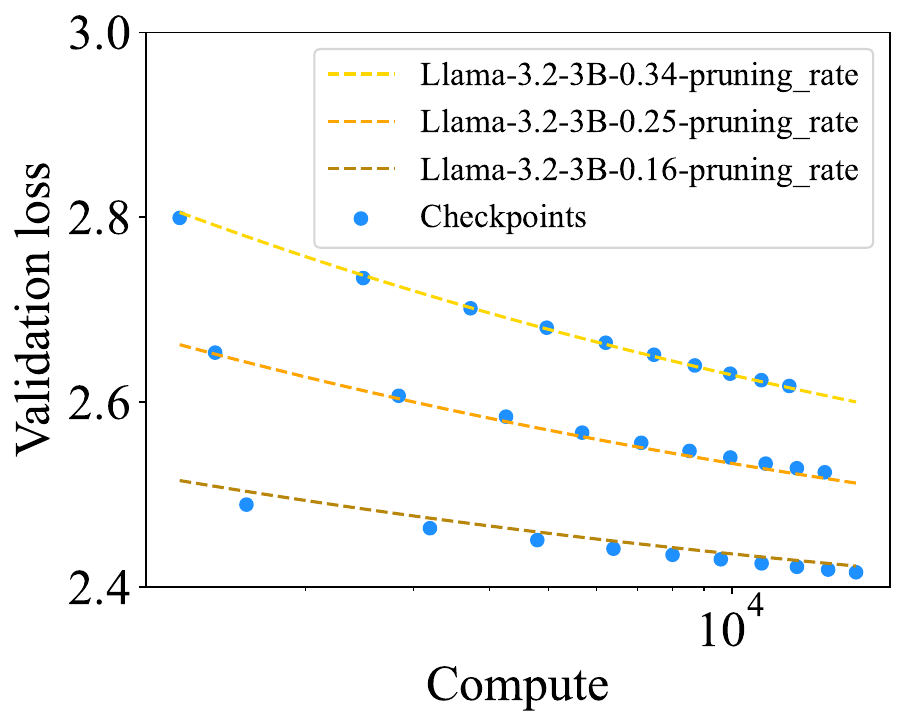}
            \caption{Loss curves derived by P$^2$ Law and the actual checkpoints of Llama-3.2-3B pruned by depth pruning.}
            \label{fig:depth_fitted_llama3b}
        \end{subfigure}
        \hfill
        \begin{subfigure}[b]{0.3\linewidth}
            \includegraphics[width=\linewidth]{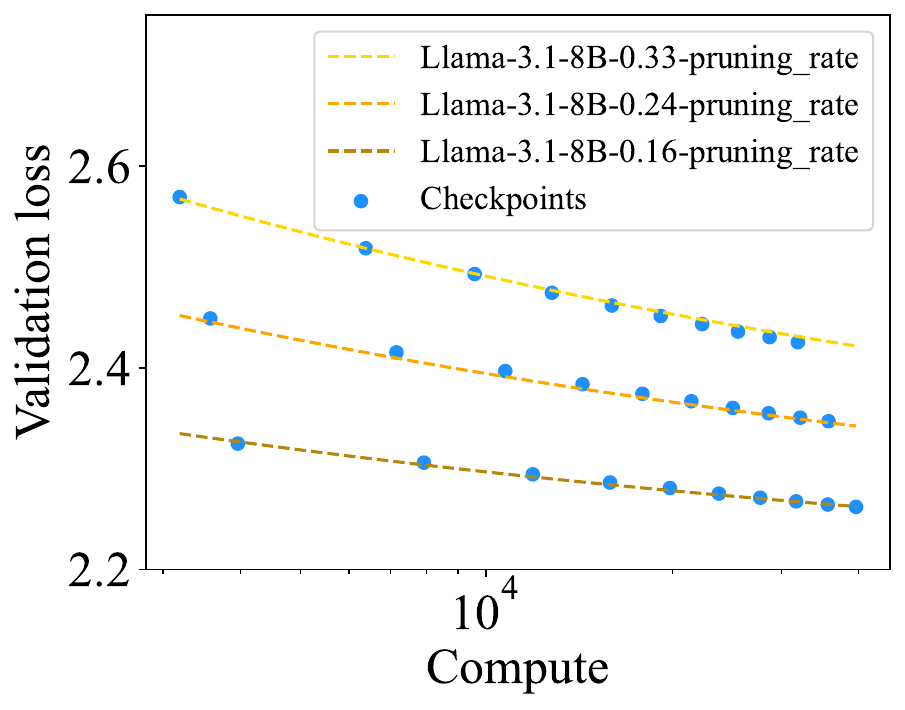}
            \caption{Loss curves derived by P$^2$ Law and the actual checkpoints of Llama-3.1-8B pruned by depth pruning.}
            \label{fig:depth_fitted_llama8b}
        \end{subfigure}
    \end{minipage}
    \caption{Loss curves derived by P$^2$ Law and the actual checkpoints of Llama-3 series models pruned by depth pruning.}
    \label{fig:fitted_llama_depth}
\end{figure*}

\subsection{Derivation of P$^2$ Law\label{sec: parameterizations}}
Previous efforts have explored scaling laws for pre-training of LLMs, with Chinchilla scaling~\cite{Hoffmann2022scaling} being a superior work, and we choose it as the foundational parameterization for our P$^2$ Law.
The Chinchilla scaling law describes the relationship between model performance and key factors such as model size, the number of pre-training tokens, and the computational resources used during the pre-training process. It is formally defined as follows:
\begin{equation}
\begin{split}
\mathcal{L}(N, D) = \frac{N_C}{N^{\alpha}} + \frac{D_C}{D^{\beta}} + E
\end{split}
\end{equation}
where $N_C$, $D_C$, $E$, $\alpha$, and $\beta$ are constants, $N$ represents the model size, $D$ denotes the number of pre-training tokens and $\mathcal{L}$ represents the model's loss.
Compared to the OpenAI scaling law~\cite{Kaplan2020Scaling}, the Chinchilla scaling law demonstrates superior performance (detailed in Appendix~\ref{sec:comparison}). Therefore, we adopt the Chinchilla scaling law as the foundational parameterization for our P$^2$ Law.
Combining the pruning rate $\rho$ and the model's loss $\mathcal{L}_0$ before pruning, we define the following three candidate parameterizations:
\begin{equation*}
\small
\begin{split}
    \mathcal{L}_1(N_0, D, \rho, \mathcal{L}_0) = \mathcal{L}_0 + (\frac{1}{\rho})^{\gamma}(\frac{1}{N_0})^{\delta} 
     (\frac{N_{C}}{N_0^{\alpha}} 
   + \frac{D_C}{D^{\beta}} + E) 
\end{split}
\end{equation*}
\begin{equation*}
\small
\begin{split}
    \mathcal{L}_2(N_0, D, \rho, \mathcal{L}_0) =  \mathcal{L}_0 + (\frac{1}{\rho})^{\gamma} 
    (\frac{N_{C}}{N_0^{\alpha} }
   + \frac{D_C}{D^{\beta}} + E) 
\end{split}
\end{equation*}
\begin{equation*}
\small
\begin{split}
    \mathcal{L}_3(N_0, D, \rho, \mathcal{L}_0) =  \mathcal{L}_0 + (\frac{1}{\rho})^{\gamma}(\frac{1}{N_0})^{\delta}
    (\frac{D_C}{D^{\beta}} + E) 
\end{split}
\end{equation*}
where $N_C$, $D_C$, $E$, $\alpha$, $\beta$, $\gamma$ and $\delta$ are constants, $N_0$ denotes the model size before pruning, $D$ denotes the number of post-training tokens and $\mathcal{L}_1, \mathcal{L}_2, \mathcal{L}_3$ denote the pruned model's post-training loss. Additionally, since there is no pruning rate in the 2:4 semi-structured pruning, P$^2$ Law for the 2:4 semi-structured pruning does not need to satisfy Condition 3. As a result, both the pruning rate and the loss before pruning are omitted and we adjust the parameterizations to:
\begin{equation}
\begin{split}
    \mathcal{L}_1(N_0, D) =  (\frac{1}{N_0})^{\delta} 
    (\frac{N_{C}}{N_0^{\alpha}}
   + \frac{D_C}{D^{\beta}} + E) 
\end{split}
\end{equation}
\begin{equation}
\begin{split}
    \mathcal{L}_2(N_0, D) =  
    (\frac{N_{C}}{N_0^{\alpha} }
   + \frac{D_C}{D^{\beta}} + E) 
\end{split}
\end{equation}
\begin{equation}
\begin{split}
    \mathcal{L}_3(N_0, D) = (\frac{1}{N_0})^{\delta}
    (\frac{D_C}{D^{\beta}} + E) 
\end{split}
\end{equation}

\begin{table*}[t]
\centering
\renewcommand\arraystretch{0.8}
\resizebox{\textwidth}{!}{
\begin{tabular}{c|c|ccc|ccc|ccc}
\toprule
 \multirow{2}{*}{LLM} & \multirow{2}{*}{Generalization}   & \multicolumn{3}{c|}{Depth pruning}      & \multicolumn{3}{c|}{Width pruning}      & \multicolumn{3}{c}{2:4 semi-structured pruning}        \\
 &    & $R^2$     & Huber loss & ASD      & $R^2$     & Huber loss & ASD      & $R^2$     & Huber loss & ASD      \\
 \midrule
\multirow{3}{*}{Llama-3 series}  
 & Dataset size &0.9725&0.000016&0.001001&0.9270&0.000019&0.000674&0.8244&0.000063&0.002561 \\
 & Model size &-0.5441&0.000745&0.001321&-&-&-&-&-&- \\
 & Pruning rate &0.9676&0.000059&0.000879&0.9707&0.000056&0.001123&-&-&- \\
 \midrule
\multirow{3}{*}{Qwen-2.5 series} 
 & Dataset size &0.9780&0.000012&0.001026&0.9896&0.000010&0.001116&0.9940&0.000000&0.000299 \\
 & Model size &-0.8786&0.000763&0.001573&0.8627&0.000137&0.001772&-&-&- \\
 & Pruning rate &0.9660&0.000043&0.000920&0.9704&0.000095&0.001003&-&-&- \\
 \bottomrule
\end{tabular}
}
\caption{Evaluation of generalization results from the perspectives of dataset size, model size, and pruning rate.}
\label{tab:Evaluation of Generalization Results}
\end{table*}

\begin{figure*}[t]
\centering
\begin{minipage}{\linewidth}
\centering
\begin{subfigure}[b]{0.3\linewidth}
\includegraphics[width=\linewidth]{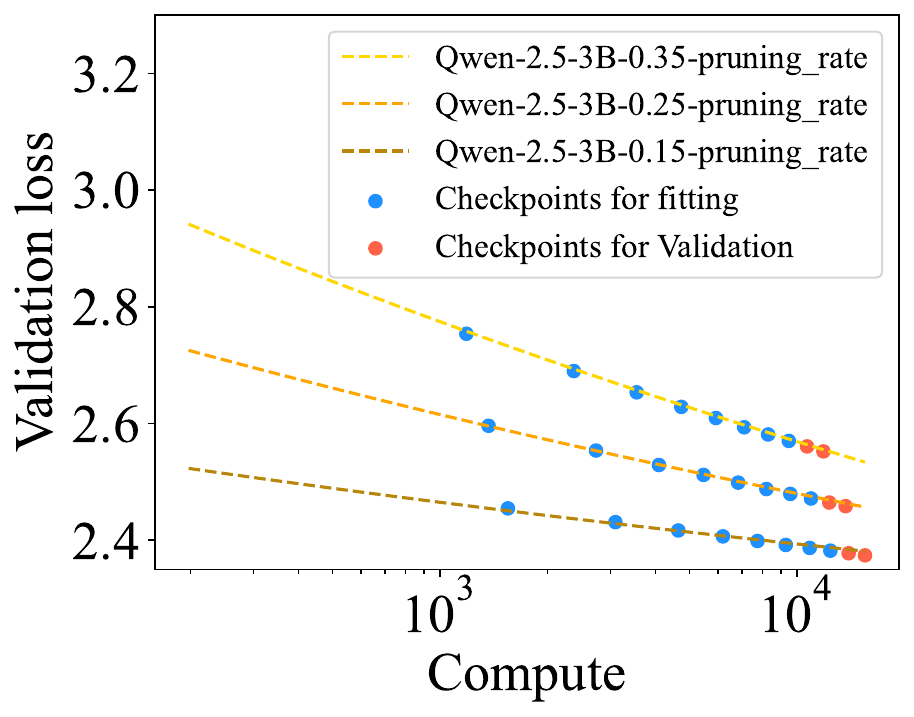}
\caption{Loss curves fitted with the P$^2$ Law using the first 80\% of checkpoints; the remaining 20\% are used for validation.}
\label{fig:generalization dataset}
\end{subfigure}
\hspace{0.03\textwidth}  
\begin{subfigure}[b]{0.3\linewidth}
\includegraphics[width=\linewidth]{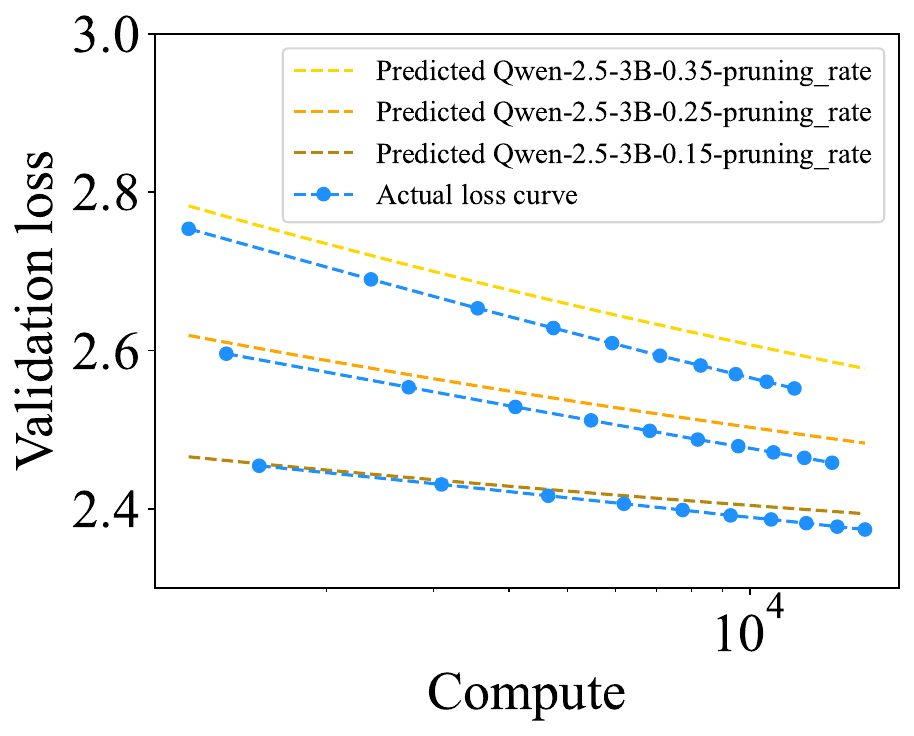}
\caption{P$^2$ Law  is fitted using checkpoints from smaller LLMs and used to predict the loss curves of larger LLMs.}
\label{fig:generalization model}
\end{subfigure}%
\hspace{0.03\textwidth}  
\begin{subfigure}[b]{0.3\linewidth}
\includegraphics[width=\linewidth]{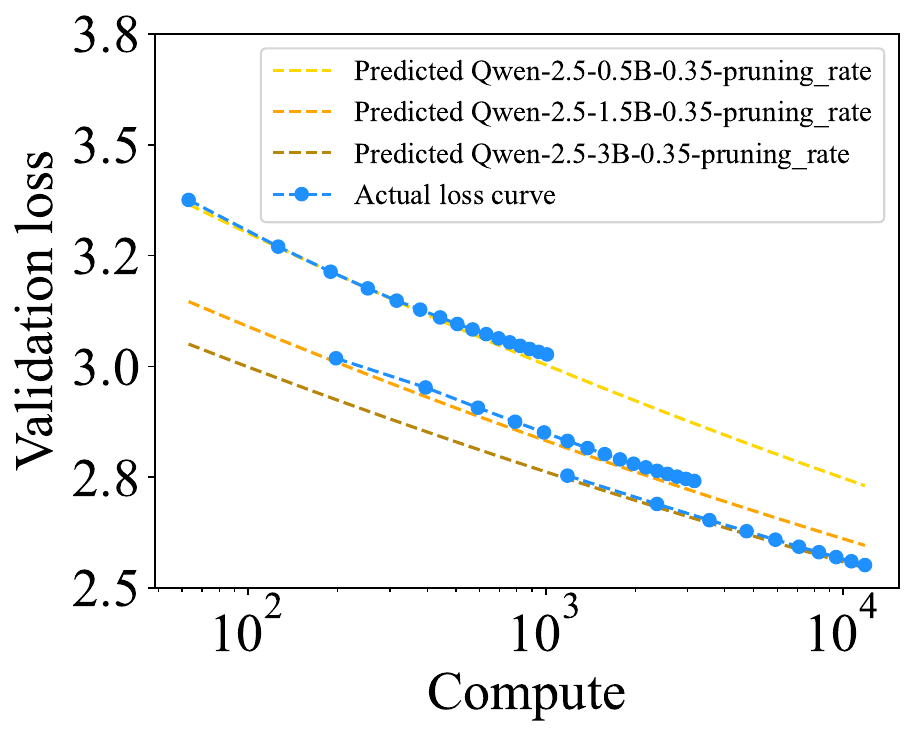}
\caption{P$^2$ Law is fitted using checkpoints from smaller pruning rates and used to predict the loss curves of larger ones.}
\label{fig:generalization pruning rate}
\end{subfigure}%
\end{minipage}
\caption{Generalization of the P$^2$ Law for Qwen-2.5 series models pruned by width pruning. }
\hfill
\end{figure*}

We utilize all checkpoints to fit the three candidate parameterizations through Levenberg-Marquardt method~\cite{more2006levenberg}, and the specific parameter values (i.e., the values of $N_C$, $D_C$, $E$, $\alpha$, $\beta$, $\gamma$, and $\delta$) for the fitted $\mathcal{L}_1$, $\mathcal{L}_2$, and $\mathcal{L}_3$ are provided in Table~\ref{tab:fit parameter} in Appendix~\ref{sec:fit parameters}.
As shown in Table~\ref{tab:Evaluation of Fitting Results}, $\mathcal{L}_1$ significantly outperforms $\mathcal{L}_2$ and $\mathcal{L}_3$ in terms of the $R^2$, Huber loss, and ASD metrics. Additionally, as shown in Table~\ref{tab:condition 2} in Appendix~\ref{sec:fit parameters}, after our calculation and verification, $\mathcal{L}_2$ and some fitted $\mathcal{L}_3$ fails to satisfy Condition 2. In contrast, all of the fitted $\mathcal{L}_1$ satisfy all three conditions. \textbf{Based on the experimental results, we select $\mathcal{L}_1$ as the parameterization for our P$^2$ Law.}

In Figure~\ref{fig:fitted_llama_depth}, we show the loss curves $\mathcal{L}_1$ derived by P$^2$ Law alongside the actual checkpoints of the Llama-3 series models pruned by depth pruning, where the compute ($C$) is approximated using the empirical formula $C = 6ND$~\cite{Kaplan2020Scaling}, and $N$ denotes the model size after pruning. Additional loss curve derived by P$^2$ Law are shown in Figure~\ref{fig:fitted_llama_width},~\ref{fig:fitted_qwen_depth},~\ref{fig:fitted_qwen_width} and~\ref{fig:fitted_semi} in Appendix~\ref{sec:addition fitting result}. As shown in these figures, the loss curve derived by P$^2$ Law accurately aligns with all actual checkpoints, under all the three pruning methods, except for the width pruning on Llama-3.1-8B. 
As shown in Figure~\ref{fig:8b_depth} and~\ref{fig:8b_width}, for Llama-3.1-8B, we observe that depth pruning outperforms width pruning at similar pruning rates, which contrasts with the observations in other cases. This suggests that width pruning on Llama-3.1-8B may lead to anomalous performance, making our law unsuitable for this special scenario.
We elaborate on this anomalous performance of width pruning on Llama-3.1-8B further in Appendix~\ref{sec:llama patterns}.

\subsection{Generalization of P\texorpdfstring{$^2$}{} Law\label{sec:Generalization}}
In this section, we explore the generalization ability of P$^2$ Law from three perspectives: dataset size, model size and pruning rate.

\subsubsection{Settings}
We begin by outlining the settings of generalization experiments as follows:

\vpara{Dataset Size.} The fitting setting follows the same setting as described in Section~\ref{sec: parameterizations}, with the only difference being that the first 80\% of the checkpoints recorded during each training process are used to fit the P$^2$ Law, and the remaining 20\% for validation.

\vpara{Model size.} We fit the P$^2$ Law using checkpoints from smaller LLMs and validate it on checkpoints from larger LLMs, while maintaining the pruning rate during both fitting and prediction. Taking the Qwen-2.5 series models as an example, we fit the P$^2$ Law using all checkpoints from Qwen-2.5-0.5B and Qwen-2.5-1.5B, and subsequently validate it with the actual checkpoints of Qwen-2.5-3B across three pruning rates. Due to the limited number of available actual loss curves for 2:4 semi-structured pruning, we did not conduct experiments for this pruning method.

\vpara{Pruning Rate.} We fit the P$^2$ Law using checkpoints from lower pruning rates and validate it using checkpoints from higher pruning rates, while keeping the model size constant during both fitting and prediction.
Taking width pruning of the Qwen-2.5 series models as an example, we fit the P$^2$ Law using checkpoints from these models at lower pruning rates (0.15 and 0.25) and then validate it with the actual checkpoints at a higher pruning rate of 0.35.
Since there is no pruning rate in the 2:4 semi-structured pruning, we only explore the generalization ability on pruning rates under depth pruning and width pruning.

Due to the anomaly of width pruning on Llama-3.1-8B (see Section~\ref{sec: parameterizations}), we exclude this model from generalization experiments.

\subsubsection{Experimental Results} 

\vpara{Dataset Size Generalization.}
The evaluation results are shown in Table~\ref{tab:Evaluation of Generalization Results}, and the loss curves of Qwen-2.5-3B (pruned by width pruning) derived by P$^2$ Law are illustrated in Figure~\ref{fig:generalization dataset}. Additional loss curves derived by P$^2$ Law are provided in Figures~\ref{fig:llama_depth_data},~\ref{fig:llama_width_data},~\ref{fig:qwen_depth_data}, and~\ref{fig:qwen_width_data} in Appendix~\ref{sec: addition generalization results}. The results in Table~\ref{tab:Evaluation of Generalization Results} show that the loss curves derived by P$^2$ Law accurately matches the validation checkpoints, indicating that the P$^2$ Law generalizes well to larger dataset sizes.

\vpara{Model Size Generalization.}
The evaluation results are presented in Table~\ref{tab:Evaluation of Generalization Results}, and the loss curves of Qwen-2.5-3B predicted by P$^2$ Law (pruned by width pruning) are visualized in Figure~\ref{fig:generalization model}. 
Additional loss curves predicted by P$^2$ Law are shown in Figure~\ref{fig:model_size} in Appendix~\ref{sec: addition generalization results}. As shown in Table~\ref{tab:Evaluation of Generalization Results}, the P$^2$ Law fitted on smaller LLMs performs poorly in $R^2$ and Huber loss when applied to larger models, indicating challenges in generalizing to larger, unseen models. However, the low ASD suggests it still captures the slope of the actual loss curve. This trend is also seen in Figure~\ref{fig:model_size}, where despite a gap between predicted and actual loss curves, the predicted and actual loss curves align in their downward trend after training stabilizes. This suggests P$^2$ Law fitted from smaller LLMs can still predict the optimal computation cost point for larger LLMs, confirming its generalization feasibility.

\vpara{Pruning Rate Generalization.}
We present the generalization evaluations in Table~\ref{tab:Evaluation of Generalization Results} and illustrate the loss curves of Qwen-2.5 series models predicted by P$^2$ Law (pruned by width pruning) in Figure~\ref{fig:generalization pruning rate}. 
Additional loss curves predicted by P$^2$ Law are provided in Figure~\ref{fig:pruning_rate_llama} and~\ref{fig:pruning_rate_qwen} in Appendix~\ref{sec: addition generalization results}.
As shown in the Figure~\ref{fig:generalization pruning rate} and Table~\ref{tab:Evaluation of Generalization Results}, the values of different metrics indicate that the actual loss curves closely align with the predicted loss curves, suggesting that the P$^2$ Law generalizes well to higher pruning rates.

\section{Related Work}

\subsection{Model Pruning}

Model pruning can be categorized into unstructured pruning and structured pruning. 

\vpara{Unstructured Pruning.} Unstructured pruning methods~\cite{sparsegpt-frantar2023, RIA-zhang2024, wanda-sun2024} compress LLMs by removing individual unimportant elements from the weight matrices, producing sparse ones. However, it is often hardware-inefficient and only speeds up LLMs when a specific sparsity pattern, such as 2:4 sparsity~\cite{2:4sparsity-mishra2021}, is applied. 
The approach which employ the 2:4 sparsity is known as semi-structured pruning. 


\vpara{Structured Pruning.} Structured pruning methods for LLMs can be divided into two categories: depth pruning~\cite{llm-streamline-chen2024, song2024slebstreamliningllmsredundancy, gromov2024unreasonable, shortgpt-men2024}, which aims to reduce the number of layers in the LLMs, and width pruning~\cite{slicegpt-ashkboos2024, sp3-hu2024, pat-liu2024, ma2023llmpruner}, which aims to reduce the embedding channels, the number of attention heads, or the intermediate size of the FFN. 

\subsection{Scaling Law}
The OpenAI scaling law~\cite{Kaplan2020Scaling} and the Chinchilla scaling law~\cite{Hoffmann2022scaling} are the most popular scaling laws in the pre-training of LLMs, both of which establishe a power-law relationship between model performance, model size, the number of pre-training tokens, and the computational resources used during pre-training.

We are the first to investigate the scaling law for the post-training after model pruning, and we propose the P$^2$ Law as a scaling law for this process. Compared to the OpenAI scaling law, the Chinchilla scaling law demonstrates superior performance (detailed in Appendix~\ref{sec:comparison}). Therefore, we adopt the Chinchilla scaling law as the foundational parameterization for our P$^2$ Law. 


\section{Conclusion}

In this paper, we conduct post-training experiments on models from the Llama-3 and Qwen-2.5 series, covering various sizes and employing both typical structured and semi-structured pruning methods. Through extensive experiments, we identify the P$^2$ Law --- the first scaling law for post-training after model pruning. Further experiments validate the effectiveness of the P$^2$ Law and demonstrate its generalization to larger dataset sizes, larger model sizes, and higher pruning rates, offering valuable insights for resource allocation in the post-training of pruned LLMs.

\section*{Limitation}
Due to constraints in GPU resources, the experiments conducted in this paper are restricted to LLMs with fewer than 8B parameters. Given the substantial increase in experimental costs for larger-scale models—for instance, training a 70B LLM with 1B tokens on 4 A800 GPUs would require approximately 1,000 hours—we intend to expand our experiments to larger models as soon as sufficient computational resources become available. This will enable us to further validate the applicability of the P$^2$ Law across a broader range of model parameter scales.

\section*{Acknowledgments}
This work is supported by the National Key Research \& Develop Plan (2023YFF0725100) and the National Natural Science Foundation of China (62322214, U23A20299, U24B20144, 62172424, 62276270).


\bibliography{custom}

\clearpage

\appendix
\section{License}
Our research is grounded in the SlimPajama training dataset, which is distributed under the Apache 2.0 license. This license allows for the free use, modification, reproduction, and distribution of the software, both for personal and commercial purposes. Consistent with open science practices, we will make our training data publicly available upon acceptance of this work. The data will be released under the CC BY-SA 4.0 license, which enables reuse and redistribution, provided that derivative works adhere to the same licensing terms

\section{Details of Pruning Methods\label{sec: pruning methods}}
\subsection{Depth Pruning\label{sec: depth pruning}}
Following the existing depth pruning methods~\cite{shortgpt-men2024,llm-streamline-chen2024, yang2024laco}, we estimate the layer importance using cosine similarity and prune layers with lower importance. Specifically, we randomly select $N$ samples from the pre-training data. We then record the hidden states generated by the LLMs for these samples and compute the cosine similarity between the input and output hidden states of each layer. Assuming that the input hidden states of layer $i$ are represented by $\boldsymbol{x}^{(i)}$, the importance score (IS) of layer $i$ is computed as:
\begin{flalign}
    \text{IS}^{\text{layer}, i} = \frac{1}{N}\sum_{j=1}^N\left( \frac{1}{L}\sum_{k=1}^{L} \frac{\boldsymbol{x}^{(i)}_{j,k} \cdot \boldsymbol{x}^{(i + 1)}_{j,k}}{\Vert \boldsymbol{x}^{(i)}_{j,k} \Vert \cdot \Vert \boldsymbol{x}^{(i + 1)}_{j,k} \Vert} \right)
\end{flalign}
where $\boldsymbol{x}^{(i)}_j$, $\boldsymbol{x}^{(i + 1)}_j \in \mathbb{R}^{d \times L}$ denotes the input and output hidden states of the $j$-th sample respectively, $L$ denotes the sequence length and $d$ denotes the hidden size.
Given the number of pruned layers $n$ determined by the target sparsity, we remove the $n$ layers corresponding to the top-$n$ highest cosine similarities for pruning.

\subsection{Width Pruning\label{sec: width pruning}}
Following the approaches of Wanda~\cite{wanda-sun2024} and MINITRON~\cite{compact-llm-muralidharan2024}, we utilize activation-based metrics for width pruning. Specifically, we randomly select $N$ samples from the pre-training data and assess the importance of embedding channels by analyzing the activations generated by the LayerNorm layers. We then prune the least important channels based on this analysis. The formula for calculating the importance score (IS) of embedding channels (emb) is as follows:
\begin{align}
   &\text{IS}^{\text{emb},i} = \frac{1}{N}\sum_{j=1}^{N}\left(\frac{1}{L}\sum_{k=1}^{L}\left| LN(\boldsymbol{x}_{j,k,i}^{LN})\right|\right) &
   \label{eq:width_IS}
\end{align}
where $\boldsymbol{x}_{j,k,i}^{LN}$ denotes the input of the $i$-th channel of the $k$-th token in the $j$-th sample at the LayerNorm layer, $L$ denotes the sequence length, and $LN$ denotes the Layer Normalization operaten. Given a specific sparsity, we calculate the number of embedding channels that need to be pruned, and then remove the channels with the lowest importance.

\subsection{2:4 Semi-Structured Pruning\label{sec: 2:4}}
Unstructured pruning removes individual unimportant elements from the weight matrices, producing sparse matrices. When the sparsity structure follows a specific pattern, such as 2:4 sparsity~\cite{2:4sparsity-mishra2021}, the model can be efficiently accelerated. This approach is known as semi-structured pruning. 
Let $W$ represent the weight matrix of a linear layer of an LLM, $x$ represent the input of the linear layer. The object of semi-structured pruning is to learn a sparsity mask $M$ and an updated weight $\Delta W$ so that the dense matrix $W$ is transformed into a sparse matrix $\tilde{W}$:
\begin{flalign}
& min\ \ \Vert Wx - \tilde{W}x \Vert \notag \\
& s.t.\quad\tilde{W} = M\cdot(W + \Delta W)
\end{flalign}
where $W \in \mathbb{R}^{d_{out} \times d_{in}}$, $M\in \{0,1\}^{d_{out} \times d_{in}}$, $\Delta W\in \mathbb{R}^{d_{out} \times d_{in}}$ and $x \in \mathbb{R}^{d_{in}}$.

We randomly select 1,024 data samples from the pre-training dataset SlimPajama for pruning. and use SparseGPT~\cite{sparsegpt-frantar2023} to optimize the aforementioned objectives. 

In the post-training process, We train this 2:4 sparse model pruned by SparseGPT. Inspired by LoRS~\cite{hu2025lorsefficientlowrankadaptation}, during the post-training process, we combine the updated weight $\Delta\tilde{W}^t$ from each training iterate $t$ with the mask $M$ to obtain the weight after update $\tilde{W}^{t}$, ensuring the model's sparsity:
\begin{align}
\tilde{W}^{t} = \tilde{W}^{t-1} + M\cdot\Delta\tilde{W}^{t}
\end{align}


\section{Batch Size and Learning Rate Settings\label{sec:lr and bs}}
Previous research indicates that the relationship between batch size and the number of model parameters is very weak~\cite{mccandlish2018empirical}. Furthermore, OpenAI Scaling Law also utilize the same batch size for models with varying parameter counts. As a result, we apply a consistent and commonly used batch size of 262k tokens across models of different scales. Regarding the learning rate, OpenAI suggests that the optimal learning rate follows a logarithmic relationship with the size of the model parameters~\cite{Kaplan2020Scaling}. Based on their provided formula, the optimal learning rate for 8B models is calculated to be 2e-3, while for 0.5B models, it is 1.8e-3, indicating a minimal difference. Furthermore, our experiments reveal that the optimal learning rate for post-training of models ranging from 0.5B to 8B is approximately 2e-5. Therefore, we adopt a uniform learning rate across models of different scales.

\section{Additional Actual Loss Curves\label{sec:addition loss curves}}
\begin{figure*}[t]
    \centering
    \begin{minipage}{\linewidth}
        \centering
        \begin{subfigure}[b]{0.3\linewidth}
            \includegraphics[width=\linewidth]{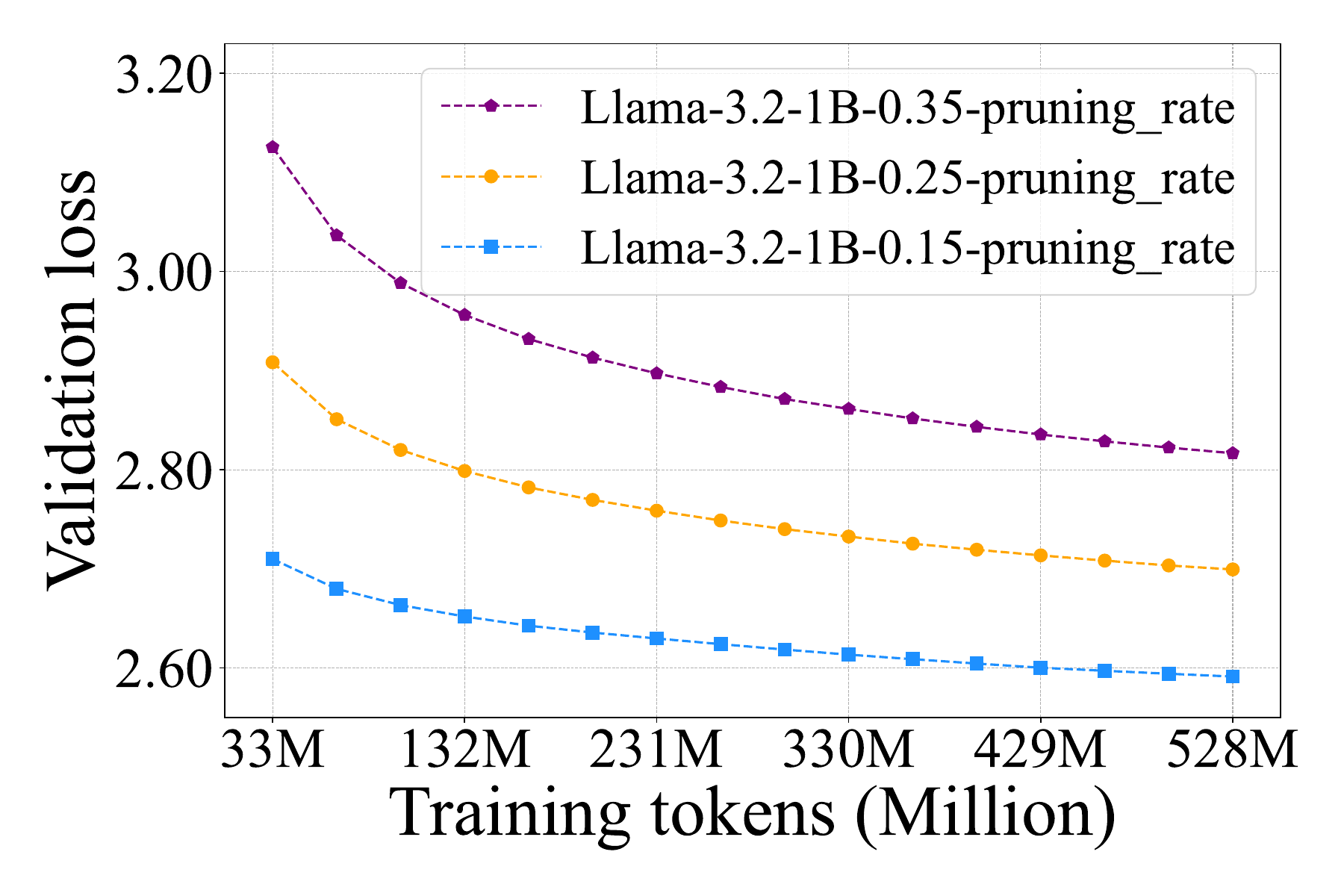}
            \caption{Post-training loss curves of Llama-3.2-1B pruned by width pruning with different pruning rates.}
            \label{fig:1b_width}
        \end{subfigure}
        \hfill
        \begin{subfigure}[b]{0.3\linewidth}
            \includegraphics[width=\linewidth]{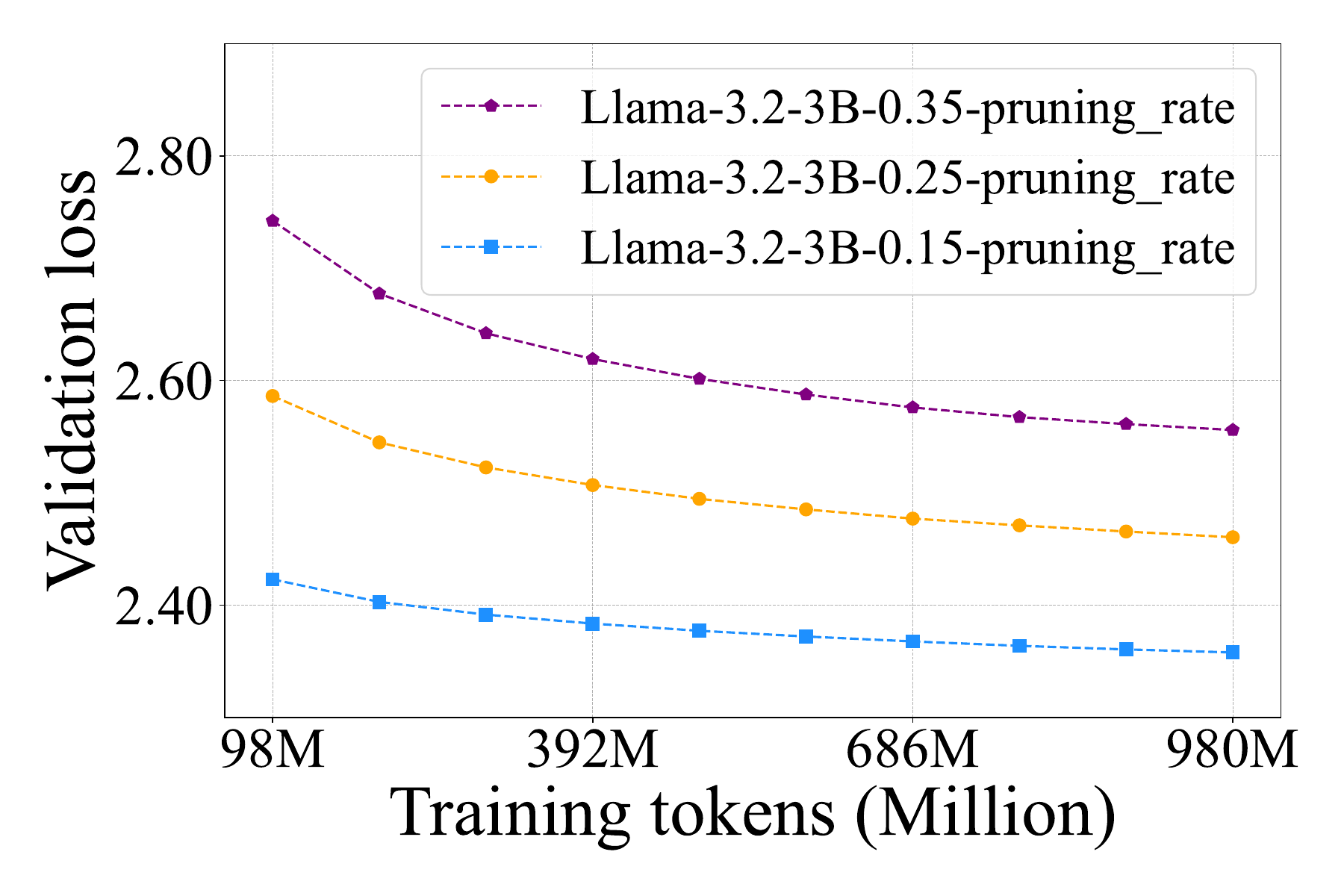}
            \caption{Post-training loss curves of Llama-3.2-3B pruned by width pruning with different pruning rates.}
            \label{fig:3b_width}
        \end{subfigure}
        \hfill
        \begin{subfigure}[b]{0.3\linewidth}
            \includegraphics[width=\linewidth]{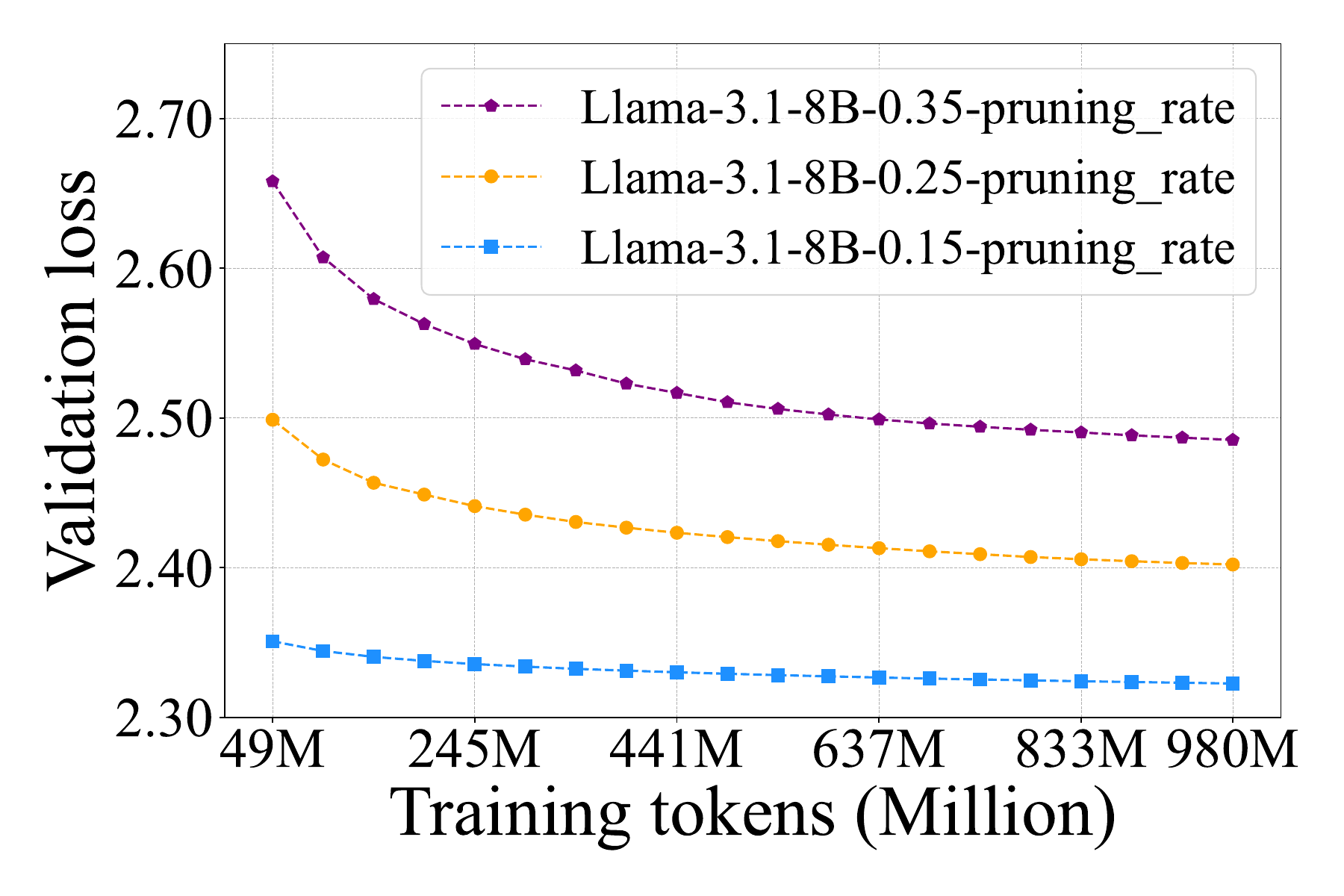}
            \caption{Post-training loss curves of Llama-3.1-8B pruned by width pruning with different pruning rates.}
            \label{fig:8b_width}
        \end{subfigure}
    \end{minipage}
    \caption{Post-training loss curves of Llama-3 series models pruned by width pruning with different pruning rates.}
    \label{fig:1-8b_width}
\end{figure*}

\begin{figure*}[t]
    \centering
    \begin{minipage}{\linewidth}
        \centering
        \begin{subfigure}[b]{0.307\linewidth}
            \includegraphics[width=\linewidth]{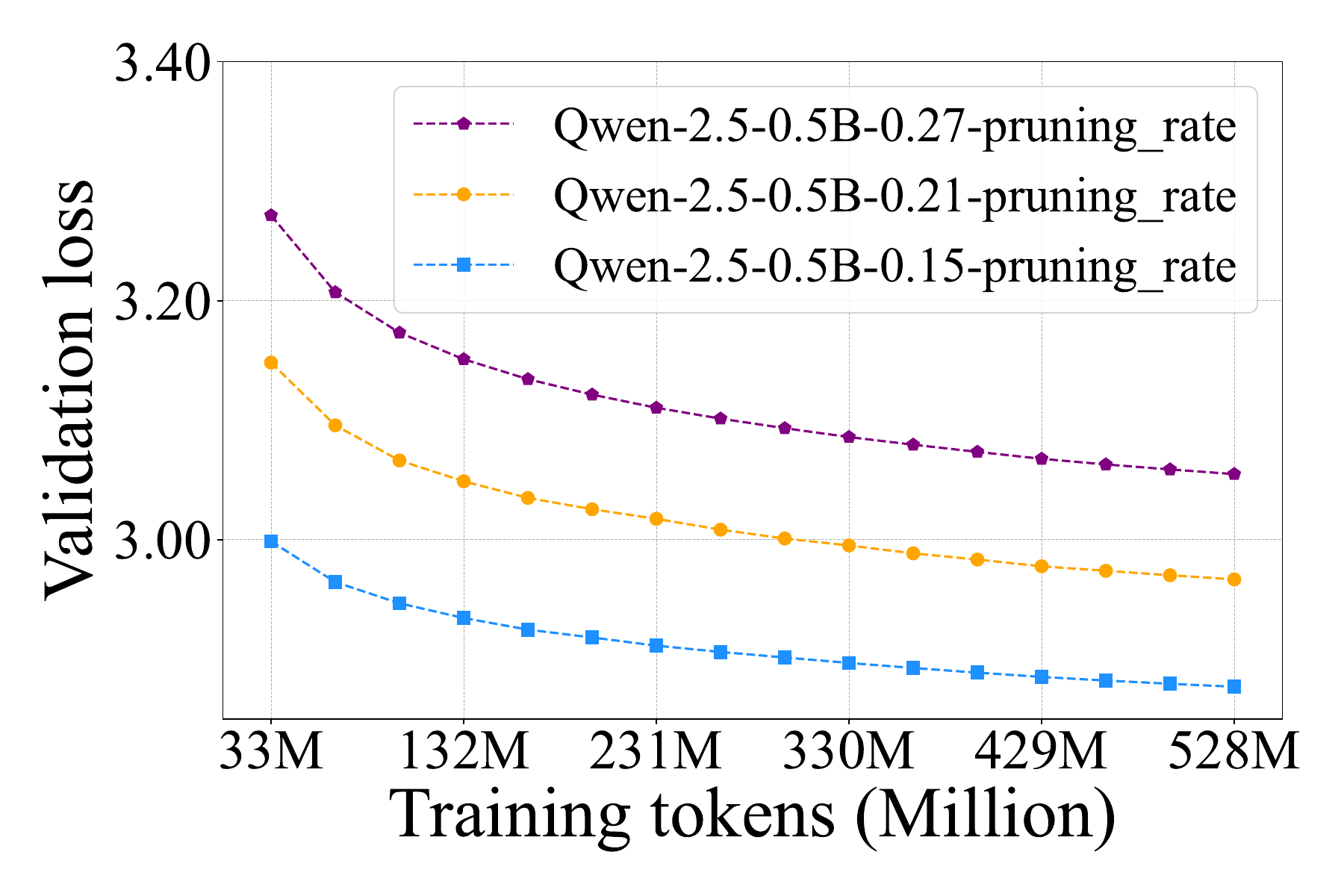}
            \caption{Post-training loss curves of Qwen-2.5-0.5B pruned by depth pruning with different pruning rates.}
            \label{fig:1b_depth_qwen}
        \end{subfigure}
        \hfill
        \begin{subfigure}[b]{0.3\linewidth}
            \includegraphics[width=\linewidth]{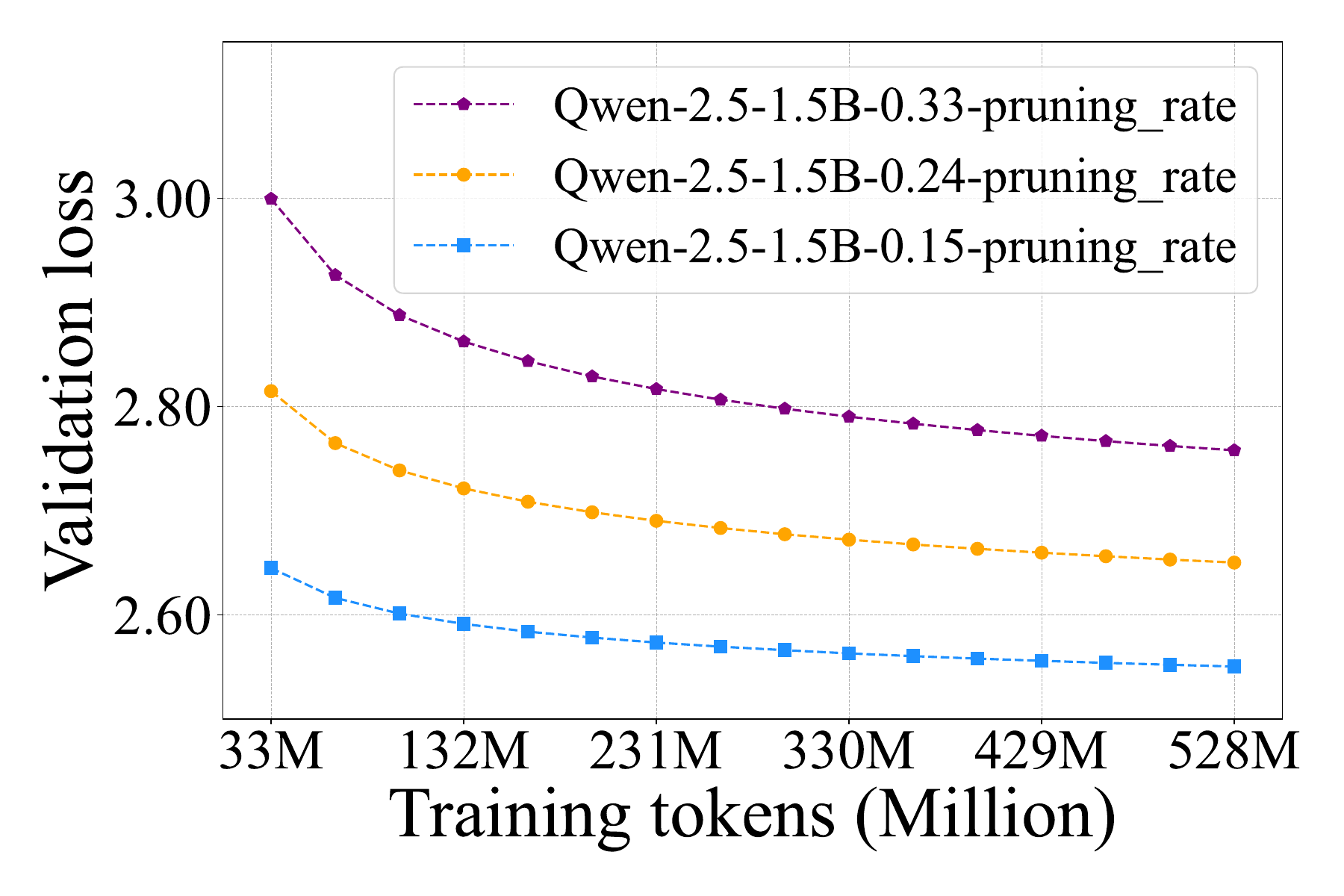}
            \caption{Post-training loss curves of Qwen-2.5-1.5B pruned by depth pruning with different pruning rates.}
            \label{fig:3b_depth_qwen}
        \end{subfigure}
        \hfill
        \begin{subfigure}[b]{0.3\linewidth}
            \includegraphics[width=\linewidth]{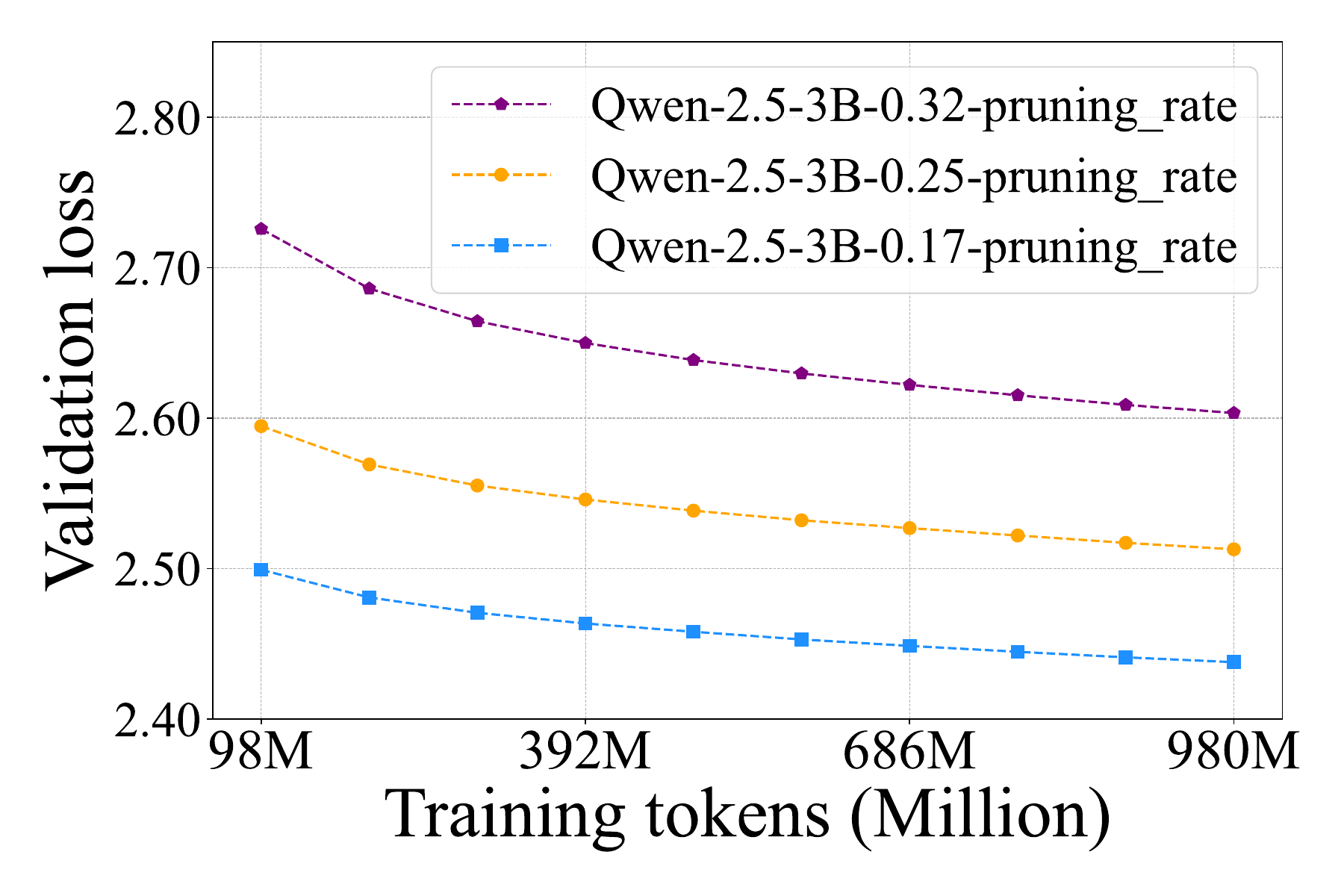}
            \caption{Post-training loss curves of Qwen-2.5-3B pruned by depth pruning with different pruning rates.}
            \label{fig:8b_depth_qwen}
        \end{subfigure}
    \end{minipage}
    \caption{Post-training loss curves of Qwen-2.5 series models pruned by depth pruning with different pruning rates.}
    \label{fig:1-8b_depth_qwen}
\end{figure*}

\begin{figure*}[t]
    \centering
    \begin{minipage}{\linewidth}
        \centering
        \begin{subfigure}[b]{0.3\linewidth}
            \includegraphics[width=\linewidth]{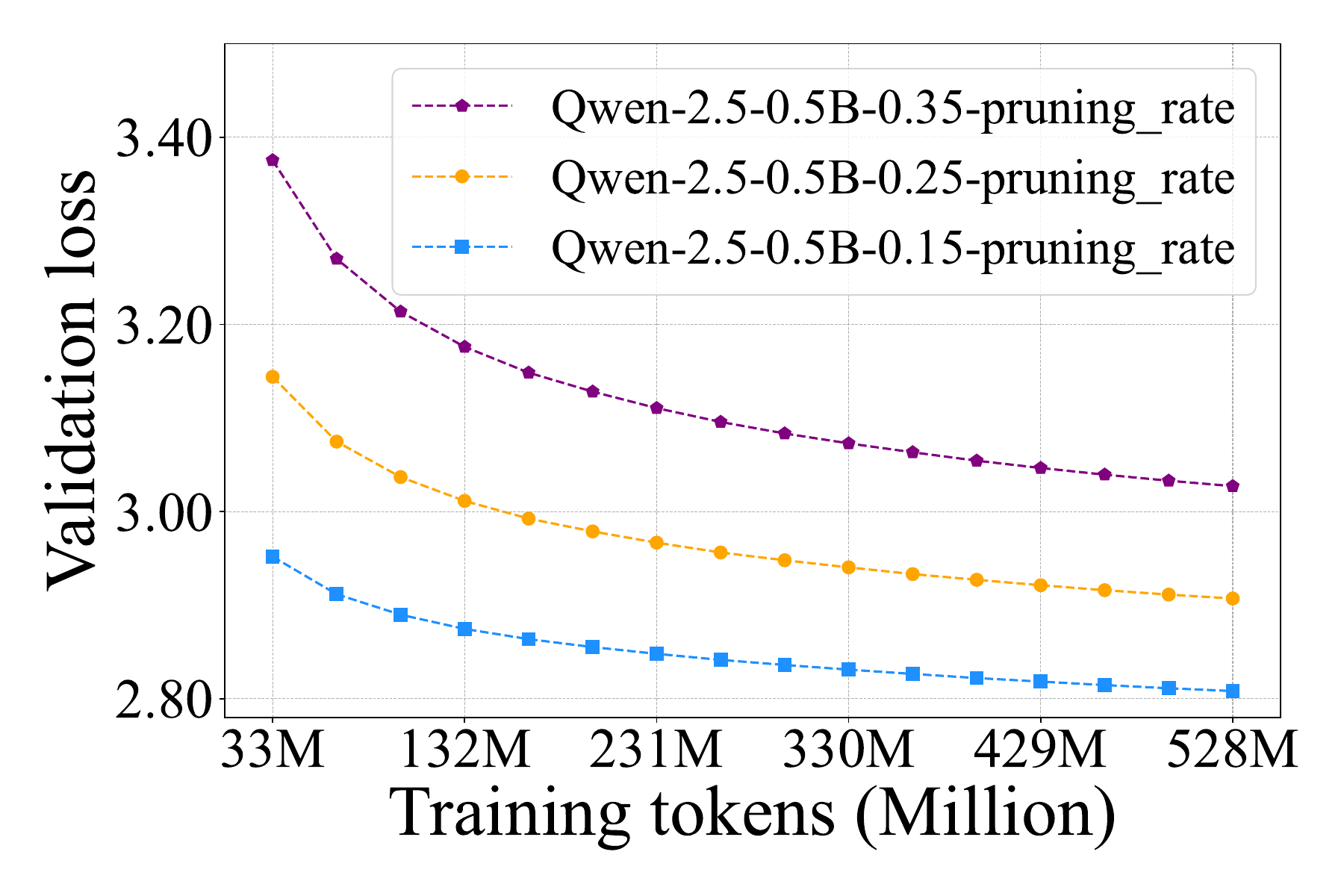}
            \caption{Post-training loss curves of Qwen-2.5-0.5B pruned by width pruning with different pruning rates.}
            \label{fig:1b_width_qwen}
        \end{subfigure}
        \hfill
        \begin{subfigure}[b]{0.307\linewidth}
            \includegraphics[width=\linewidth]{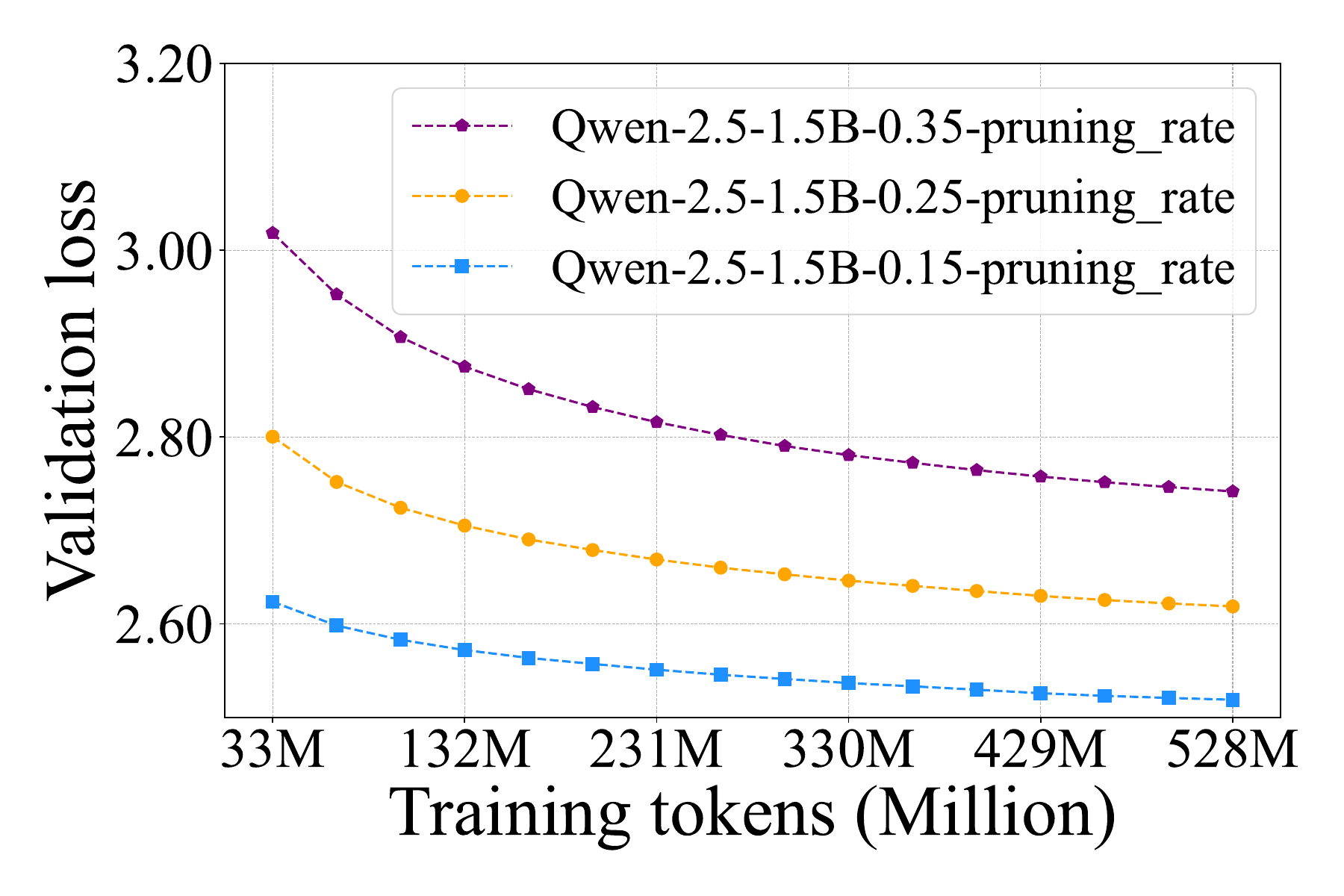}
            \caption{Post-training loss curves of Qwen-2.5-1.5B pruned by width pruning with different pruning rates.}
            \label{fig:3b_width_qwen}
        \end{subfigure}
        \hfill
        \begin{subfigure}[b]{0.307\linewidth}
            \includegraphics[width=\linewidth]{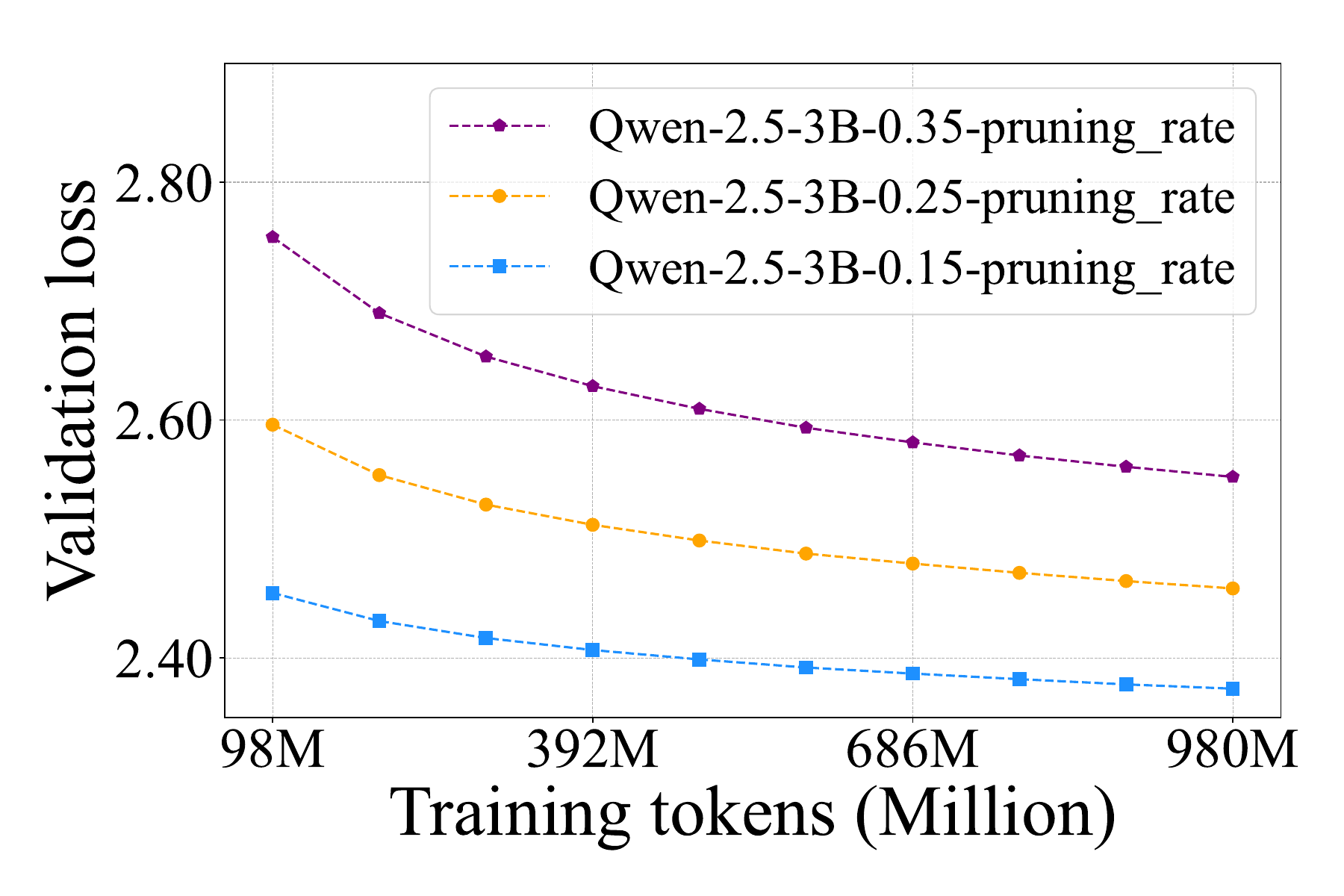}
            \caption{Post-training loss curves of Qwen-2.5-3B pruned by width pruning with different pruning rates.}
            \label{fig:8b_width_qwen}
        \end{subfigure}
    \end{minipage}
    \caption{Post-training loss curves of Qwen-2.5 series models pruned by width pruning with different pruning rates.}
    \label{fig:1-8b_width_qwen}
\end{figure*}

\begin{figure*}[t] 
\centering
\begin{minipage}[t]{0.458\textwidth}
\centering
\includegraphics[width=\linewidth]{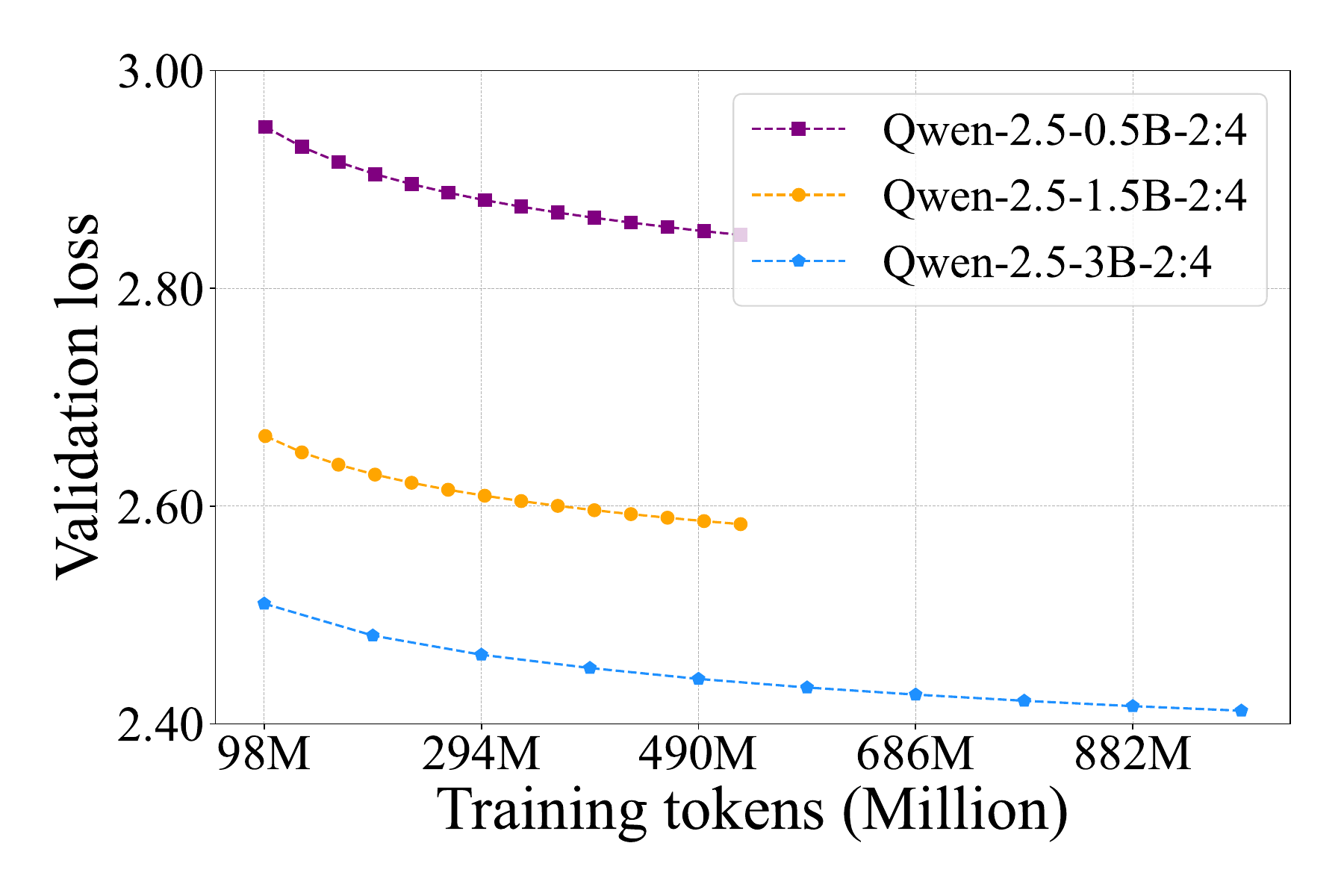}
\caption{Post-training loss curves of Qwen-2.5 series models pruned by 2:4 semi-structured pruning.}
\label{fig:1_8bsemi_qwen}
\end{minipage}%
\hfill 
\begin{minipage}[t]{0.45\textwidth}
\centering
\includegraphics[width=\linewidth]{Figures/semi.pdf}
\caption{Post-training loss curves of Llama-3 series models pruned by 2:4 semi-structured pruning.}
\end{minipage}
\end{figure*}

The additional post-training loss curves for models pruned by width pruning or for the Qwen-2.5 series models are provided in Figures~\ref{fig:1-8b_width},~\ref{fig:1-8b_depth_qwen}, \ref{fig:1-8b_width_qwen} and \ref{fig:1_8bsemi_qwen}.

\section{Comparison with OpenAI Scaling Law\label{sec:comparison}}
Kaplan~\cite{Kaplan2020Scaling} propose OpenAI scaling law as follows:
\begin{equation}
\begin{split}
    \mathcal{L}(N, D) = (\frac{N_{C}}{N^{\alpha}} + \frac{D_C}{D})^{\beta} 
\end{split}
\end{equation}
where $N_C$, $D_c$, $\alpha$ and $\beta$ are constants, $N$ denotes the model size and $D$ denotes the number of pre-training tokens.
We have also defined the following parameterizations based on the OpenAI scaling law:
\begin{equation}
\begin{split}
    \mathcal{L}_4(N_0, D, \rho, \mathcal{L}_0) = \mathcal{L}_0 + (\frac{1}{\rho})^{\gamma}(\frac{1}{N_0})^{\delta} 
    (\frac{N_{C}}{N_0^{\alpha}}
   + \frac{D_C}{D})^{\beta} 
\end{split}
\end{equation}
\begin{equation}
\begin{split}
    \mathcal{L}_5(N_0, D, \rho, \mathcal{L}_0) =  \mathcal{L}_0 + (\frac{1}{\rho})^{\gamma} 
    (\frac{N_{C}}{N_0^{\alpha} }
   + \frac{D_C}{D})^{\beta}
\end{split}
\end{equation}
where $N_C$, $D_C$, $\alpha$, $\beta$, $\gamma$, $\delta$ denotes constants, $N_0$ denotes the model size before pruning, $D$ denotes the number of post-training tokens, $\rho$ denotes pruning rate, $\mathcal{L}_0$ denotes the model's loss before pruning and $\mathcal{L}_4, \mathcal{L}_5$ denote pruned model's post-training loss. 

We utilize all the checkpoints to fit the two parameterizations described above, and the evaluation results are presented in Table~\ref{tab:openai scaling law comparison}. The results show that the performance of these two parameterizations is weaker than that of $\mathcal{L}_1$. Therefore, we adopt the Chinchilla scaling law as the foundational parameterization for our P$^2$ Law.

\section{Parameter Values of Fitted Parameterizations\label{sec:fit parameters}}
We present the parameter values of the fitted $\mathcal{L}_1$, $\mathcal{L}_2$, and $\mathcal{L}_3$ in the Table~\ref{tab:fit parameter}. In addition, we calculate whether $\mathcal{L}_1$, $\mathcal{L}_2$, and $\mathcal{L}_3$ satisfy Condition 2, and the results are shown in the Table~\ref{tab:condition 2}. 

\section{Additional Loss Curves Derived by P$^2$ Law\label{sec:addition fitting result}}
\begin{figure*}[t]
    \centering
    \begin{minipage}{\linewidth}
        \centering
        \begin{subfigure}[b]{0.3\linewidth}
            \includegraphics[width=\linewidth]{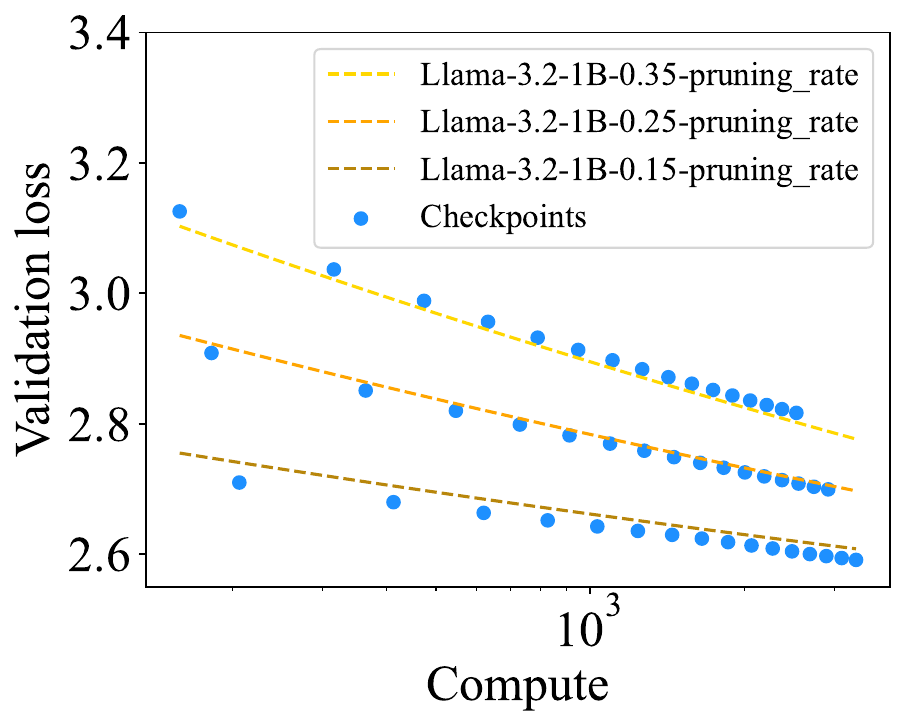}
            \caption{Loss curves derived by P$^2$ Law and the actual checkpoints of Llama-3.2-1B pruned by width pruning.}
            \label{fig:width_fitted_llama1b}
        \end{subfigure}
        \hfill
        \begin{subfigure}[b]{0.3\linewidth}
            \includegraphics[width=\linewidth]{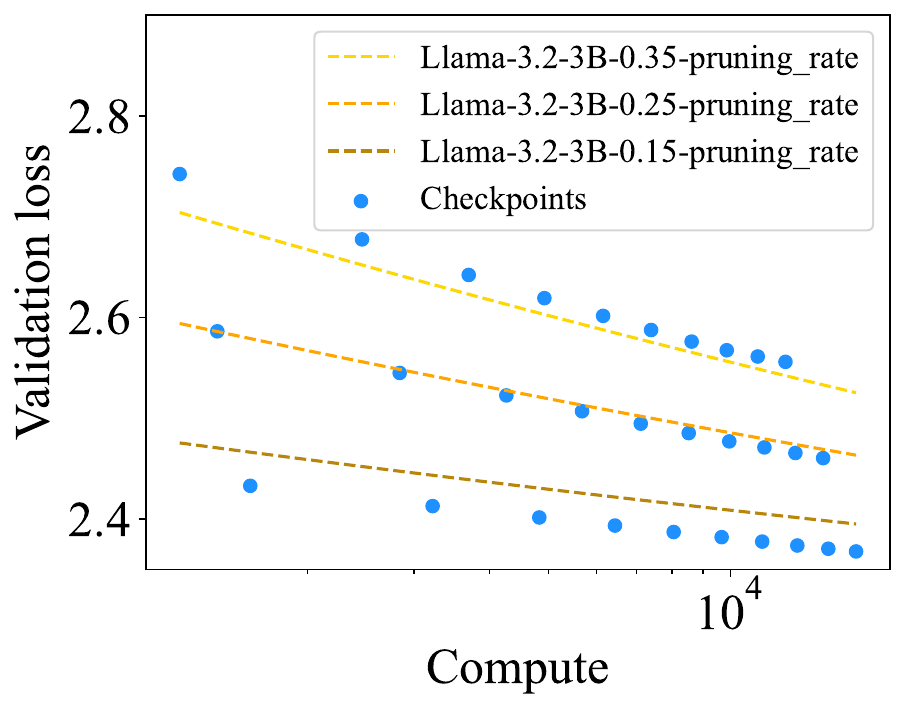}
            \caption{Loss curves derived by P$^2$ Law and the actual checkpoints of Llama-3.2-3B pruned by width pruning.}
            \label{fig:width_fitted_llama3b}
        \end{subfigure}
        \hfill
        \begin{subfigure}[b]{0.301\linewidth}
            \includegraphics[width=\linewidth]{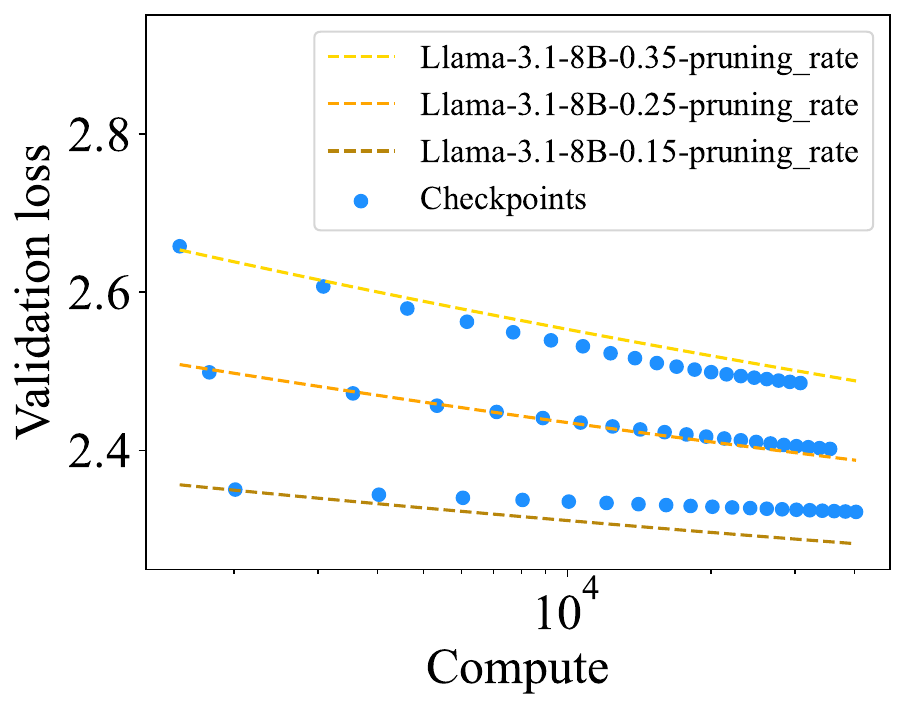}
            \caption{Loss curves derived by P$^2$ Law and the actual checkpoints of Llama-3.1-8B pruned by width pruning.}
            \label{fig:width_fitted_llama8b}
        \end{subfigure}
    \end{minipage}
    \caption{Loss curves derived by P$^2$ Law and the actual checkpoints of Llama-3 series models pruned by width pruning.}
    \label{fig:fitted_llama_width}
\end{figure*}

\begin{figure*}[t]
    \centering
    \begin{minipage}{\linewidth}
        \centering
        \begin{subfigure}[b]{0.3\linewidth}
            \includegraphics[width=\linewidth]{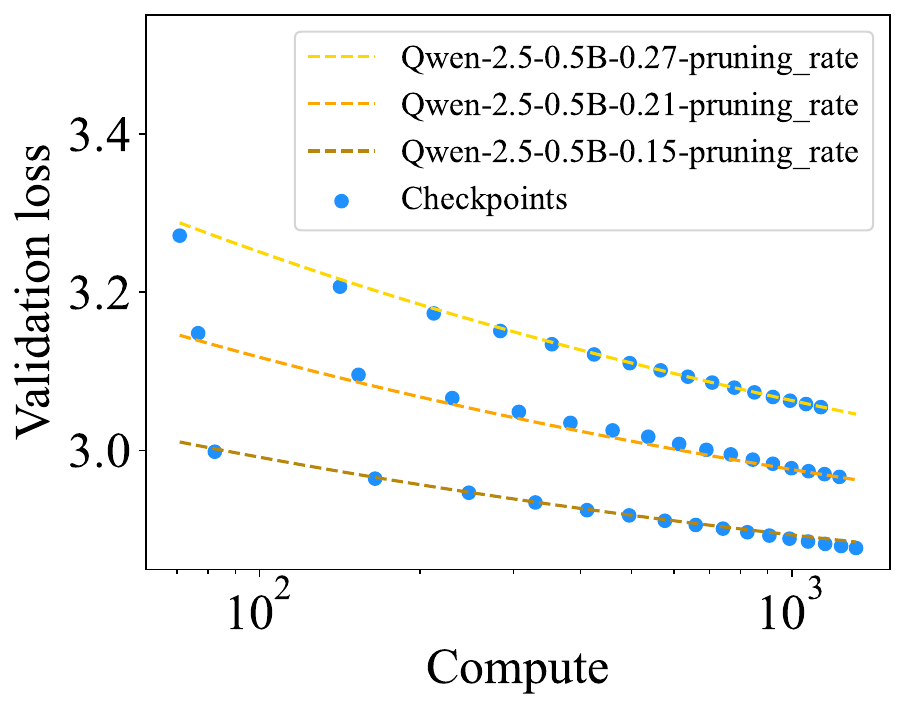}
            \caption{Loss curves derived by P$^2$ Law and the actual checkpoints of Qwen-2.5-0.5B pruned by depth pruning.}
            \label{fig:depth_fitted_qwen0.5b}
        \end{subfigure}
        \hfill
        \begin{subfigure}[b]{0.3\linewidth}
            \includegraphics[width=\linewidth]{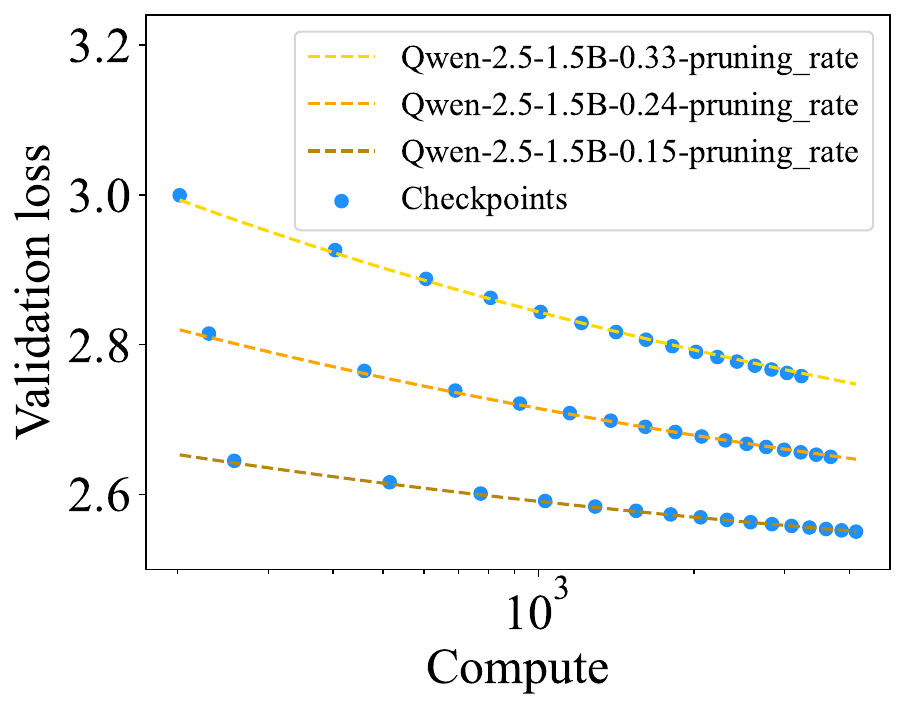}
            \caption{Loss curves derived by P$^2$ Law and the actual checkpoints of Qwen-2.5-1.5B pruned by depth pruning.}
            \label{fig:depth_fitted_qwen1.5b}
        \end{subfigure}
        \hfill
        \begin{subfigure}[b]{0.3\linewidth}
            \includegraphics[width=\linewidth]{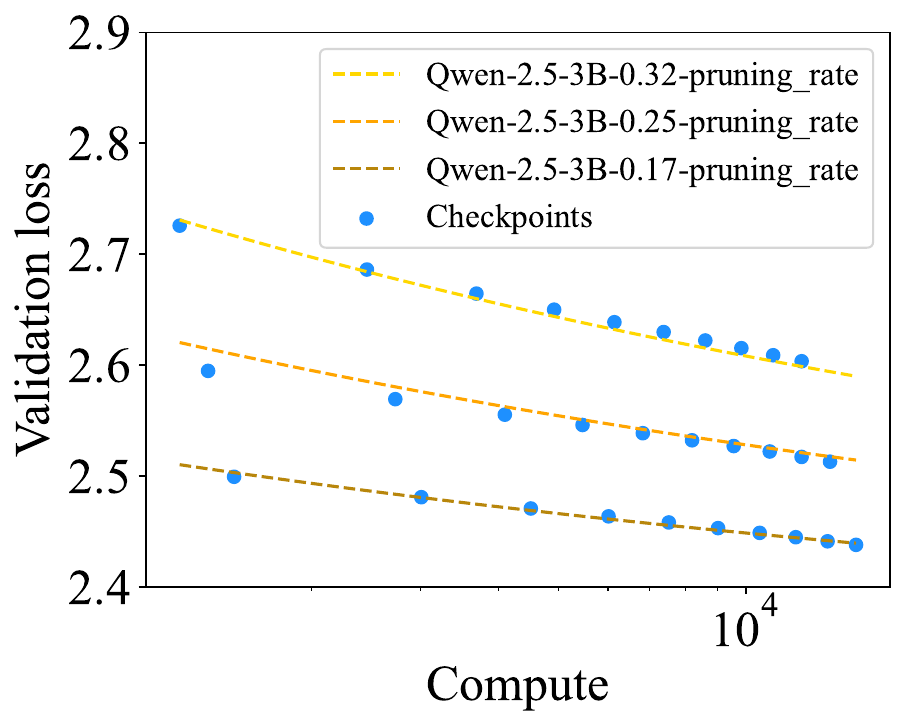}
            \caption{Loss curves derived by P$^2$ Law and the actual checkpoints of Qwen-2.5-3B pruned by depth pruning.}
            \label{fig:depth_fitted_qwen3b}
        \end{subfigure}
    \end{minipage}
    \caption{Loss curves derived by P$^2$ Law and the actual checkpoints of Qwen-2.5 series models pruned by depth pruning.}
    \label{fig:fitted_qwen_depth}
\end{figure*}

\begin{figure*}[t]
    \centering
    \begin{minipage}{\linewidth}
        \centering
        \begin{subfigure}[b]{0.3\linewidth}
            \includegraphics[width=\linewidth]{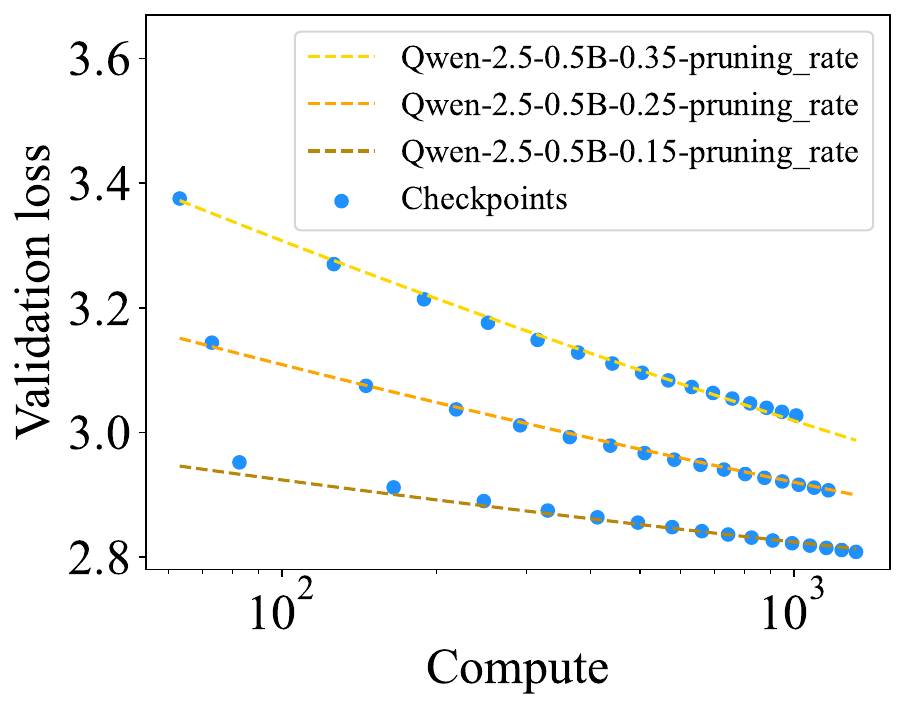}
            \caption{Loss curves derived by P$^2$ Law and the actual checkpoints of Qwen-2.5-0.5B pruned by width pruning.}
            \label{fig:width_fitted_qwen0.5b}
        \end{subfigure}
        \hfill
        \begin{subfigure}[b]{0.3\linewidth}
            \includegraphics[width=\linewidth]{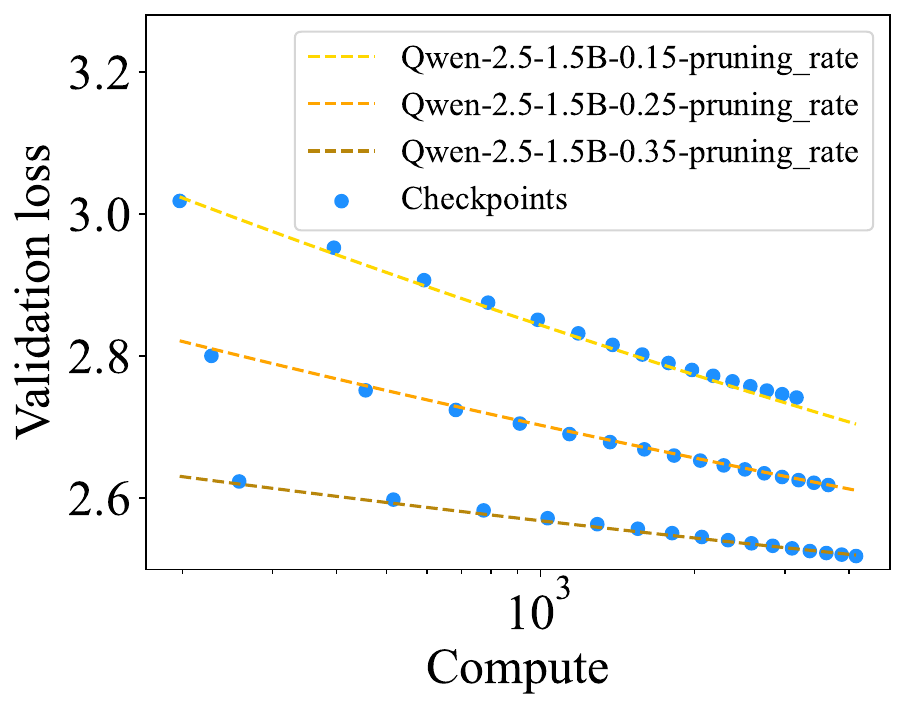}
            \caption{Loss curves derived by P$^2$ Law and the actual checkpoints of Qwen-2.5-1.5B pruned by width pruning.}
            \label{fig:width_fitted_qwen1.5b}
        \end{subfigure}
        \hfill
        \begin{subfigure}[b]{0.301\linewidth}
            \includegraphics[width=\linewidth]{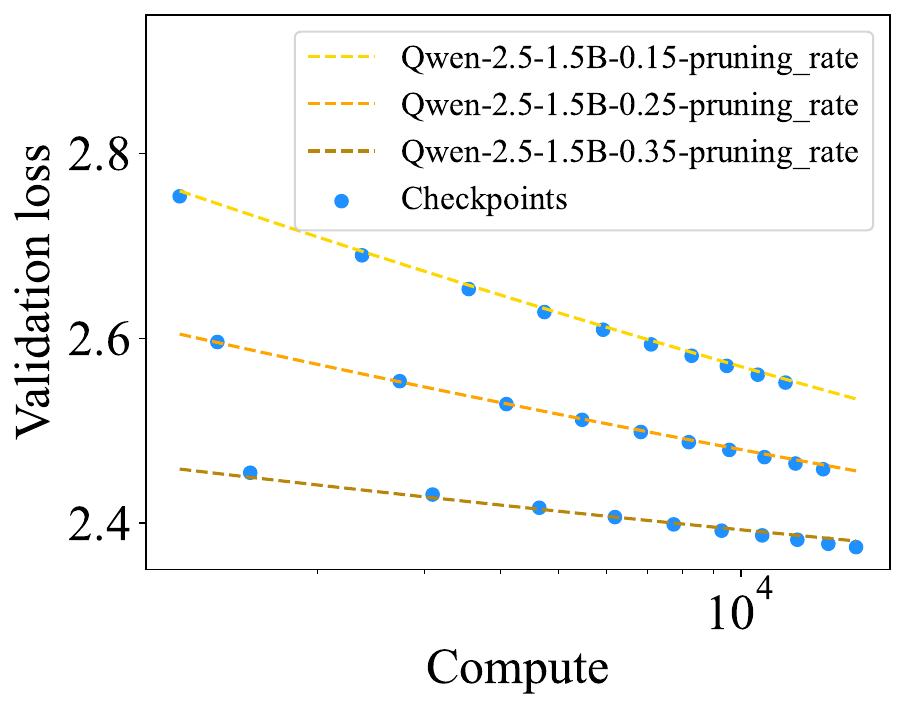}
            \caption{Loss curves derived by P$^2$ Law and the actual checkpoints of Qwen-2.5-3B pruned by width pruning.}
            \label{fig:width_fitted_qwen3b}
        \end{subfigure}
    \end{minipage}
    \caption{Loss curves derived by P$^2$ Law and the actual checkpoints of Qwen-2.5 series models pruned by width pruning.}
    \label{fig:fitted_qwen_width}
\end{figure*}

\begin{figure*}[t]
    \centering
    \begin{minipage}{\linewidth}
        \centering
        \begin{subfigure}[b]{0.45\linewidth}
            \includegraphics[width=\linewidth]{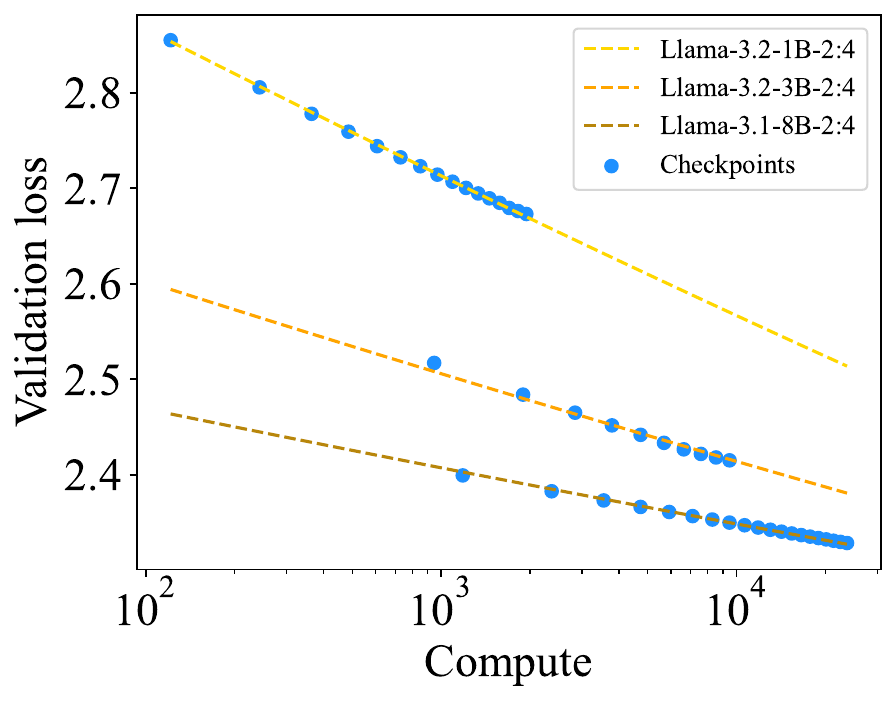}
            \caption{Loss curves derived by P$^2$ Law and the actual checkpoints of Llama-3 series models pruned by 2:4 semi-structured pruning.}
            \label{fig:semi_fitted_llama}
        \end{subfigure}
        \hfill
        \begin{subfigure}[b]{0.455\linewidth}
            \includegraphics[width=\linewidth]{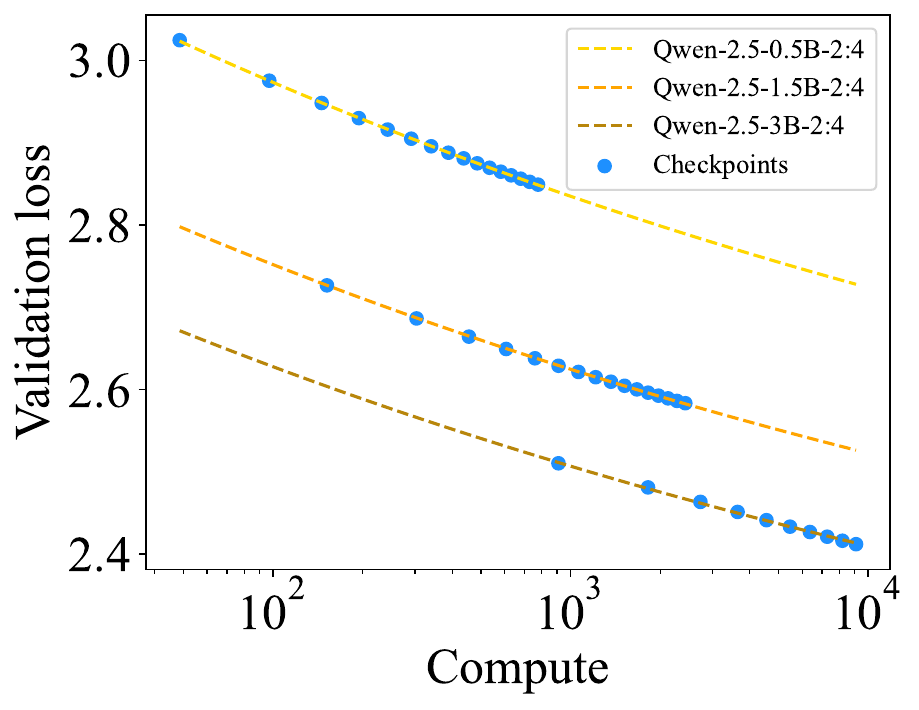}
            \caption{Loss curves derived by P$^2$ Law and the actual checkpoints of Qwen-2.5 series models pruned by 2:4 semi-structured pruning.}
            \label{fig:semi_fitted_qwen}
        \end{subfigure}
    \end{minipage}
    \caption{Loss curves derived by P$^2$ Law and the actual checkpoints of Llama-3 seires and Qwen-2.5 series models pruned by 2:4 semi-structured pruning.}
    \label{fig:fitted_semi}
\end{figure*}

The additional loss curves derived by P$^2$ Law are shown in the Figure~\ref{fig:fitted_llama_width},~\ref{fig:fitted_qwen_depth},~\ref{fig:fitted_qwen_width} and~\ref{fig:fitted_semi}.

\section{Patterns of the Llama-3 Series Models in Terms of Width\label{sec:llama patterns}}
\begin{figure*}[t]
    \centering
    \begin{minipage}{\linewidth}
        \centering
        \begin{subfigure}[b]{0.3\linewidth}
            \includegraphics[width=\linewidth]{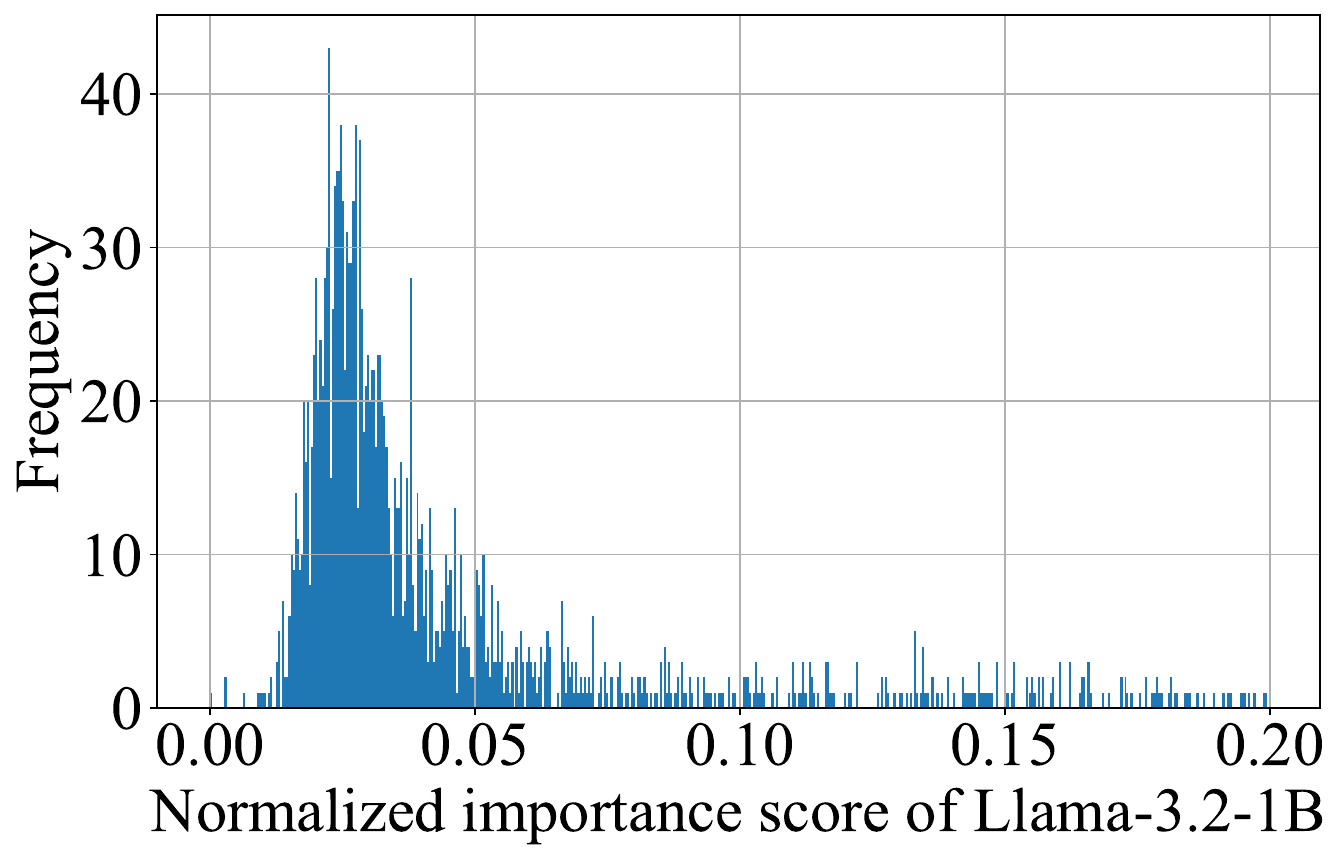}
            \caption{Histogram of the normalized importance scores for the embedding channels of Llama-3.2-1B.}
            \label{fig:1b_Histogram}
        \end{subfigure}
        \hfill
        \begin{subfigure}[b]{0.3\linewidth}
            \includegraphics[width=\linewidth]{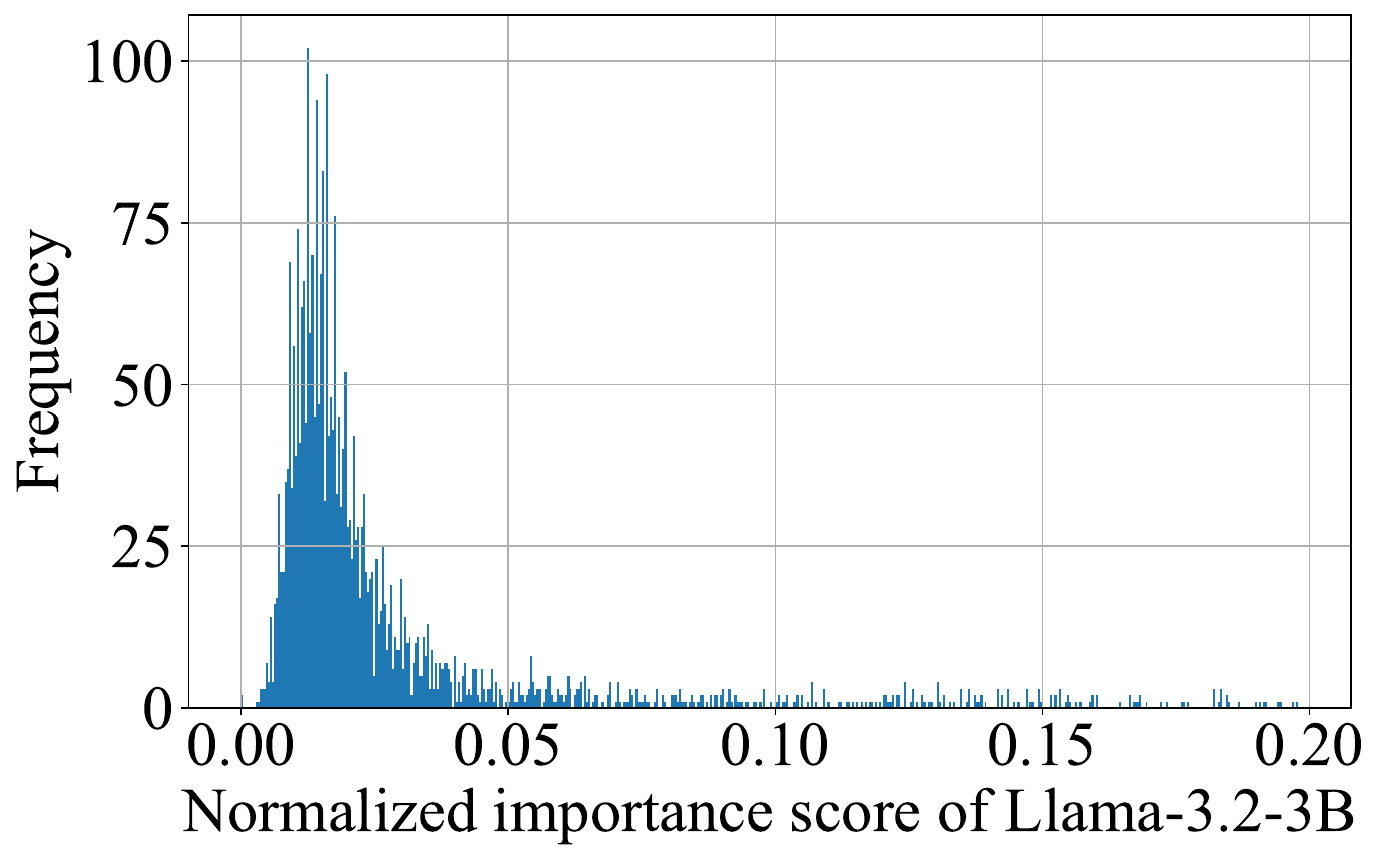}
            \caption{Histogram of the normalized importance scores for the embedding channels of Llama-3.2-3B.}
            \label{fig:3b_Histogram}
        \end{subfigure}
        \hfill
        \begin{subfigure}[b]{0.3\linewidth}
            \includegraphics[width=\linewidth]{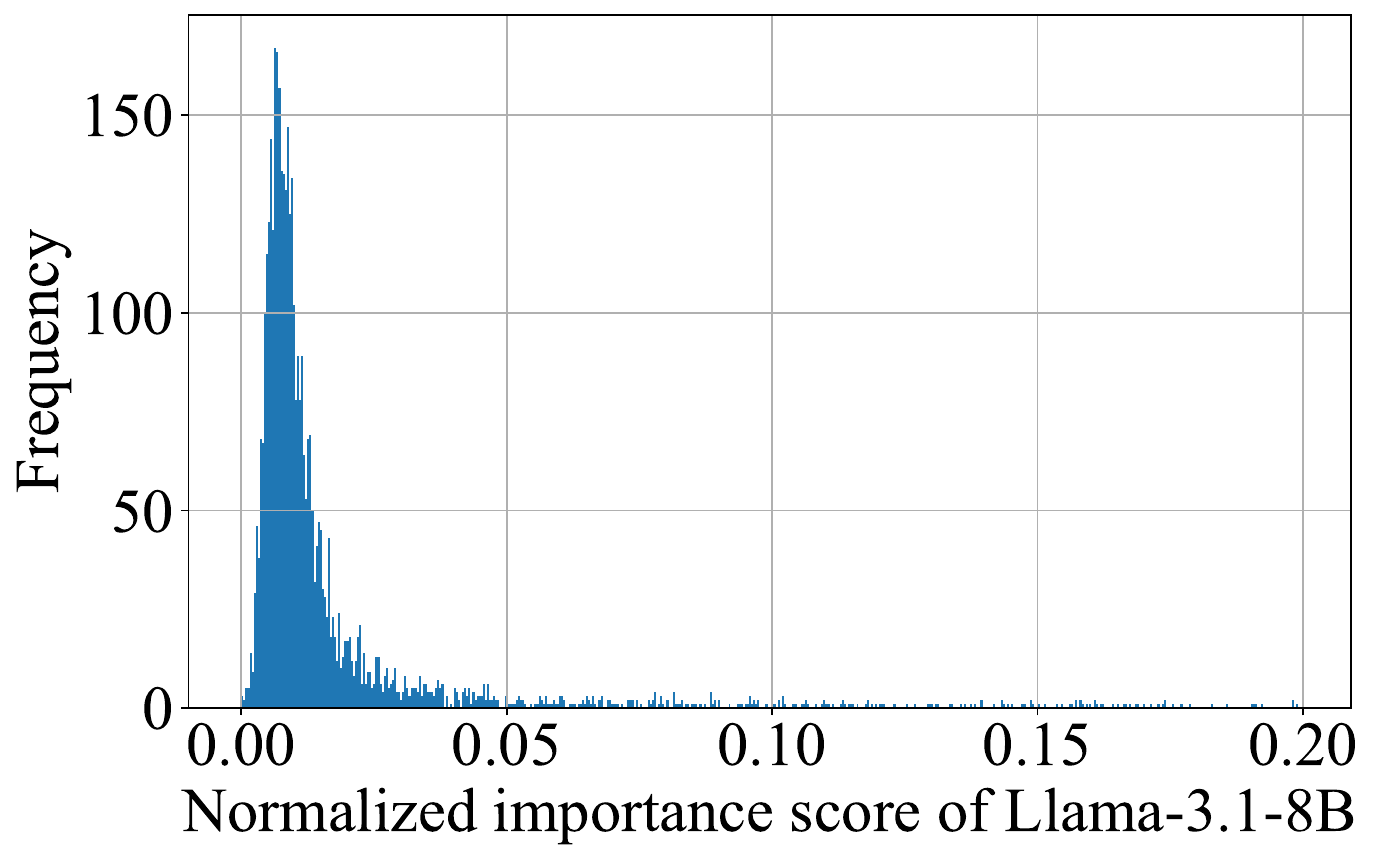}
            \caption{Histogram of the normalized importance scores for the embedding channels of Llama-3.1-8B.}
            \label{fig:8b_Histogram}
        \end{subfigure}
    \end{minipage}
    \caption{Histogram of the normalized importance scores for the embedding channels of Llama-3 series models.}
    \label{fig:1-8b_Histogram}
\end{figure*}

As discussed in Section~\ref{sec: parameterizations}, we observe an anomalous phenomenon in Llama-3.1-8B under width pruning. To investigate this further, we analyze the behavior of the Llama-3 series models with respect to width. Using a random sample of 1024 data points from SlimPajama and applying Eq.\ref{eq:width_IS}, we plot the importance score distributions of the embedding channels for the Llama-3 series models, as shown in Figure~\ref{fig:1-8b_Histogram}. 
For easier comparison, we normalize the Importance score, which is defined as follows:
$$\text{IS}^{\text{emb},i} = \frac{\text{IS}^{\text{emb},i} - \rm min(\text{IS}^{\text{emb},1}, \text{IS}^{\text{emb},2}, ..., \text{IS}^{\text{emb},N_d})}{\rm max(\text{IS}^{\text{emb},1}, \text{IS}^{\text{emb},2}, ..., \text{IS}^{\text{emb},N_d})}$$

where the $N_d$ denotes the number of embedding channels.
Additionally, we remove the extremely high values that represent a very small proportion of the data.
The figure shows that the importance scores of Llama-3.1-8B are more densely distributed compared to those of Llama-3.2-1B and Llama-3.2-3B. This denser distribution may hinder the ability to effectively distinguish less important channels in Llama-3.1-8B based on importance scores, which could potentially explain the observed anomalies in Llama-3.1-8B.

\section{Additional Generalization Loss Curves\label{sec: addition generalization results}}
\begin{figure*}[t]
    \centering
    \begin{minipage}{\linewidth}
        \centering
        \begin{subfigure}[b]{0.3\linewidth}
            \includegraphics[width=\linewidth]{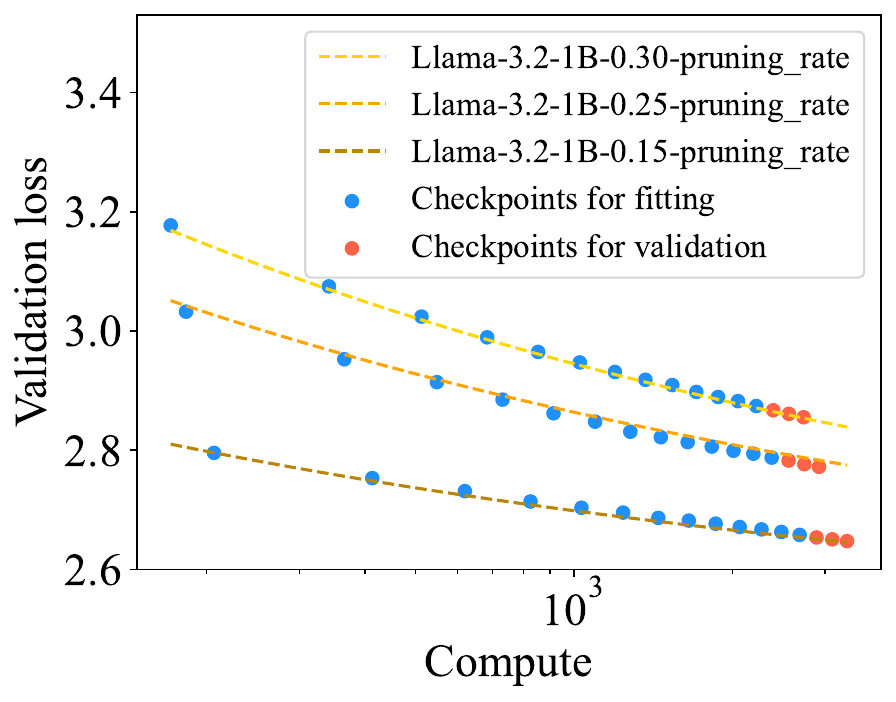}
            \caption{Loss curves fitted with the P$^2$ Law using the first 80\% of checkpoints; the remaining 20\% are used for validation. (Llama-3.2-1B pruned by depth pruning)}
        \end{subfigure}
        \hfill
        \begin{subfigure}[b]{0.3\linewidth}
            \includegraphics[width=\linewidth]{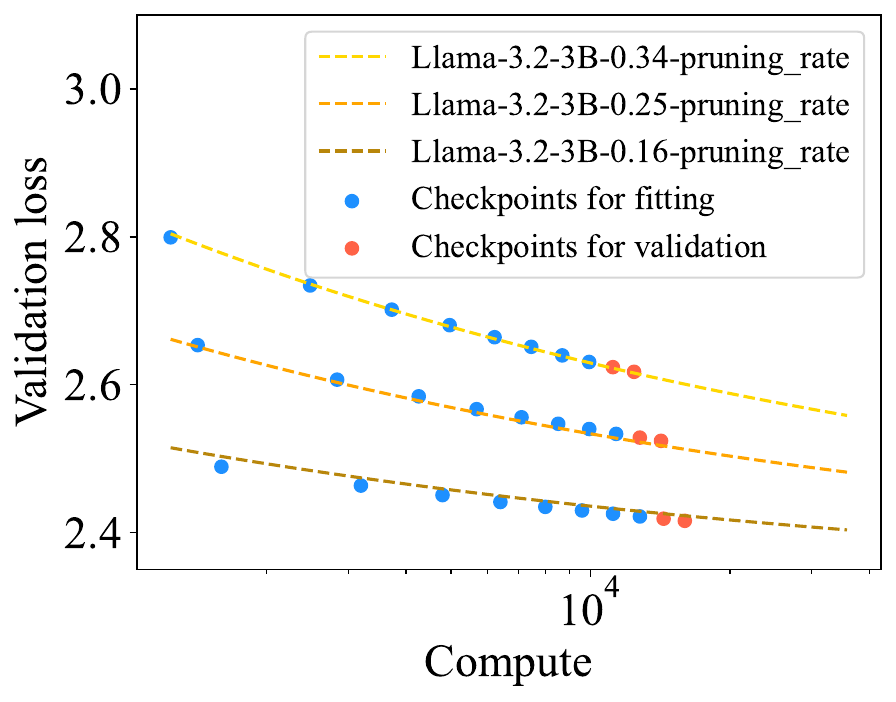}
            \caption{Loss curves fitted with the P$^2$ Law using the first 80\% of checkpoints; the remaining 20\% are used for validation. (Llama-3.2-3B pruned by depth pruning)}
        \end{subfigure}
        \hfill
        \begin{subfigure}[b]{0.301\linewidth}
            \includegraphics[width=\linewidth]{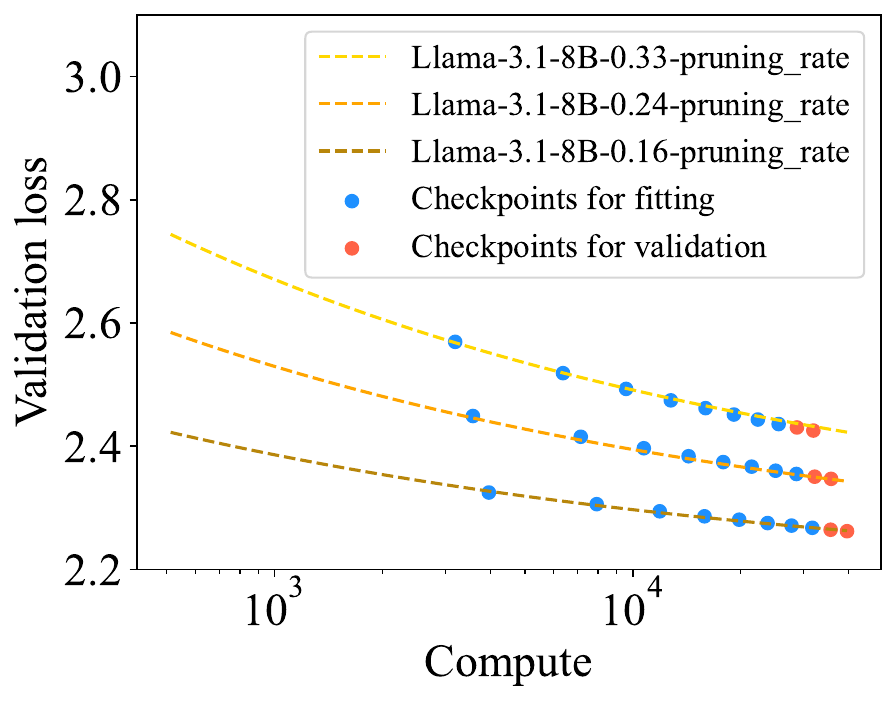}
            \caption{Loss curves fitted with the P$^2$ Law using the first 80\% of checkpoints; the remaining 20\% are used for validation. (Llama-3.1-8B pruned by depth pruning)}
        \end{subfigure}
    \end{minipage}
    \caption{Generalization of the P$^2$ Law for Llama-3 series models pruned by depth pruning on dataset size.}
    \label{fig:llama_depth_data}
\end{figure*}

\begin{figure*}[t]
    \centering
    \begin{minipage}{\linewidth}
        \centering
        \begin{subfigure}[b]{0.45\linewidth}
            \includegraphics[width=\linewidth]{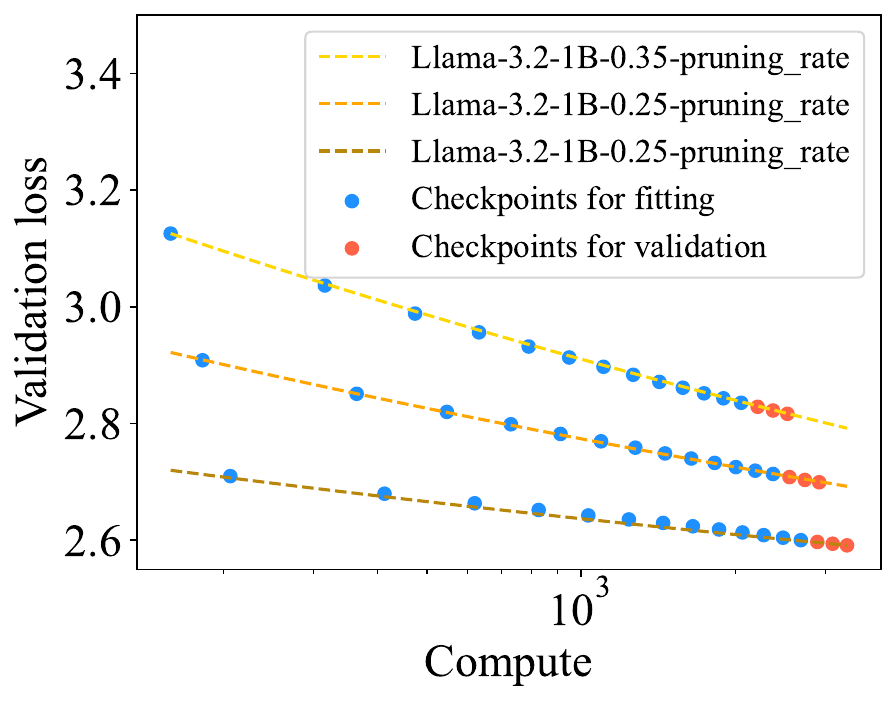}
            \caption{Loss curves fitted with the P$^2$ Law using the first 80\% of checkpoints; the remaining 20\% are used for validation. (Llama-3.2-1B pruned by width pruning)}
        \end{subfigure}
        \hfill
        \begin{subfigure}[b]{0.45\linewidth}
            \includegraphics[width=\linewidth]{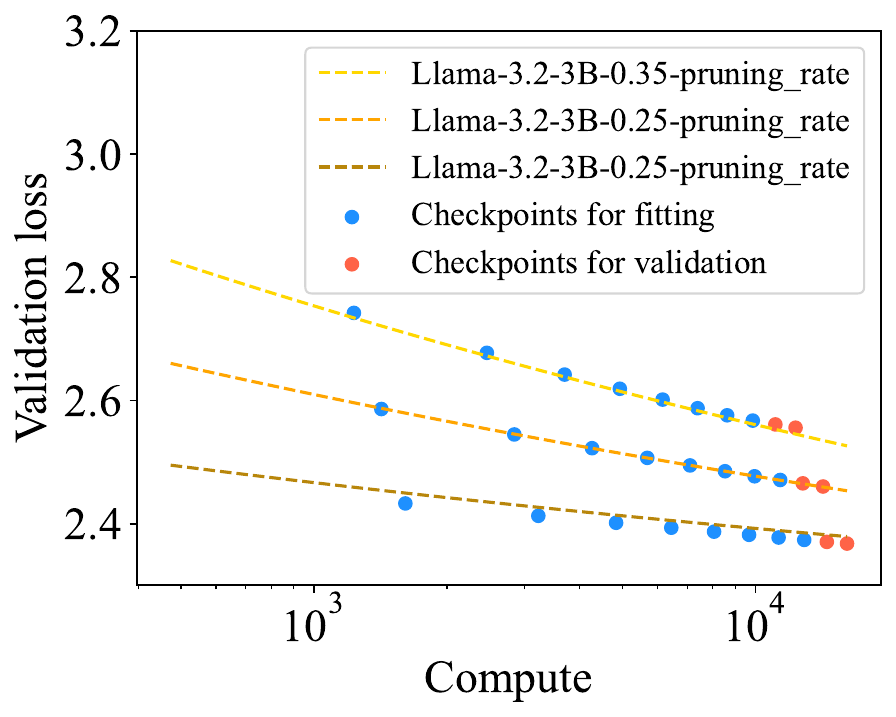}
            \caption{Loss curves fitted with the P$^2$ Law using the first 80\% of checkpoints; the remaining 20\% are used for validation. (Llama-3.2-3B pruned by width pruning)}
        \end{subfigure}
    \end{minipage}
    \caption{Generalization of the P$^2$ Law for Llama-3 series models pruned by width pruning on dataset size.}
    \label{fig:llama_width_data}
\end{figure*}

\begin{figure*}[t]
    \centering
    \begin{minipage}{\linewidth}
        \centering
        \begin{subfigure}[b]{0.3\linewidth}
            \includegraphics[width=\linewidth]{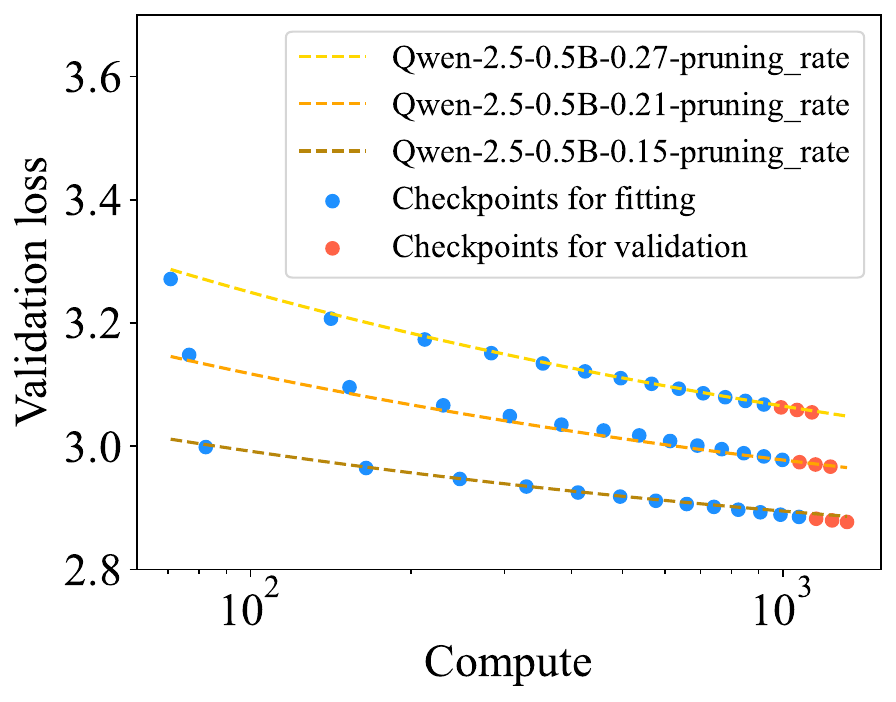}
            \caption{Loss curves fitted with the P$^2$ Law using the first 80\% of checkpoints; the remaining 20\% are used for validation. (Qwen-2.5-0.5B pruned by depth pruning)}
        \end{subfigure}
        \hfill
        \begin{subfigure}[b]{0.3\linewidth}
            \includegraphics[width=\linewidth]{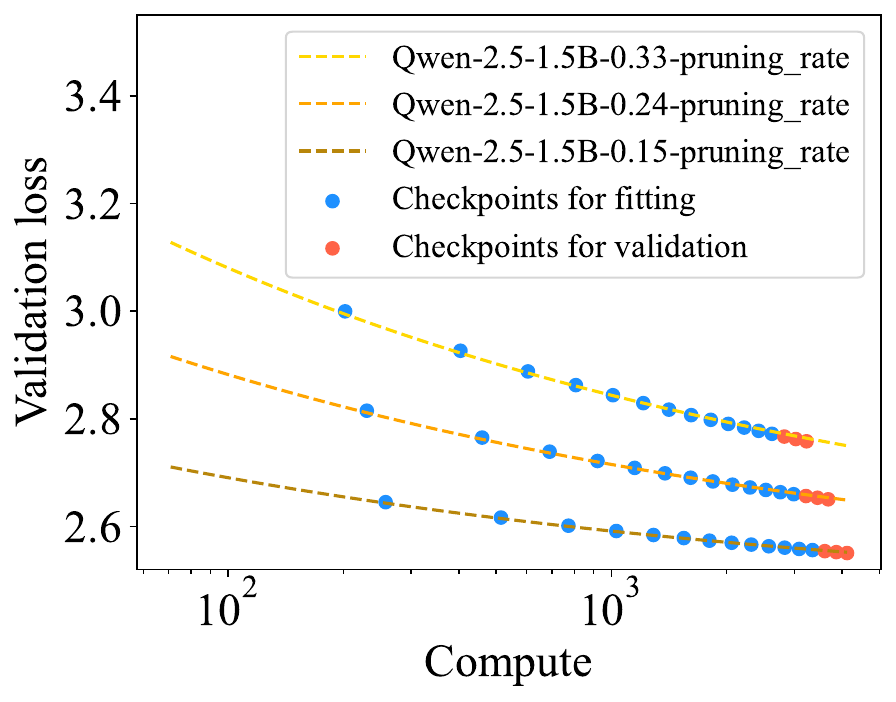}
            \caption{Loss curves fitted with the P$^2$ Law using the first 80\% of checkpoints; the remaining 20\% are used for validation. (Qwen-2.5-1.5B pruned by depth pruning)}
        \end{subfigure}
        \hfill
        \begin{subfigure}[b]{0.301\linewidth}
            \includegraphics[width=\linewidth]{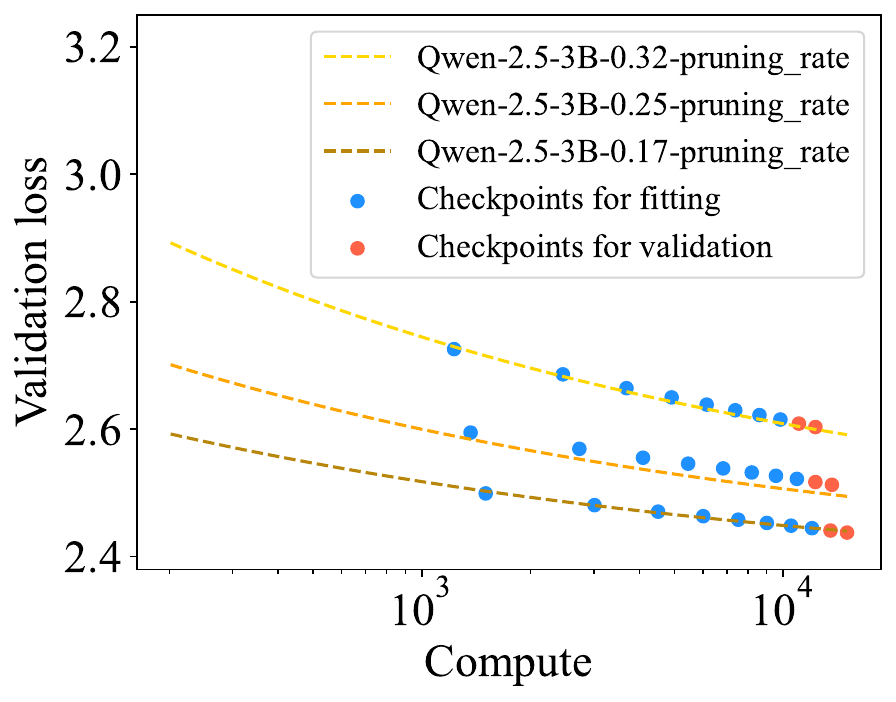}
            \caption{Loss curves fitted with the P$^2$ Law using the first 80\% of checkpoints; the remaining 20\% are used for validation. (Qwen-2.5-3B pruned by depth pruning)}
        \end{subfigure}
    \end{minipage}
    \caption{Generalization of the P$^2$ Law for Qwen-2.5 series models pruned by depth pruning on dataset size.}
    \label{fig:qwen_depth_data}
\end{figure*}

\begin{figure*}[t]
    \centering
    \begin{minipage}{\linewidth}
        \centering
        \begin{subfigure}[b]{0.3\linewidth}
            \includegraphics[width=\linewidth]{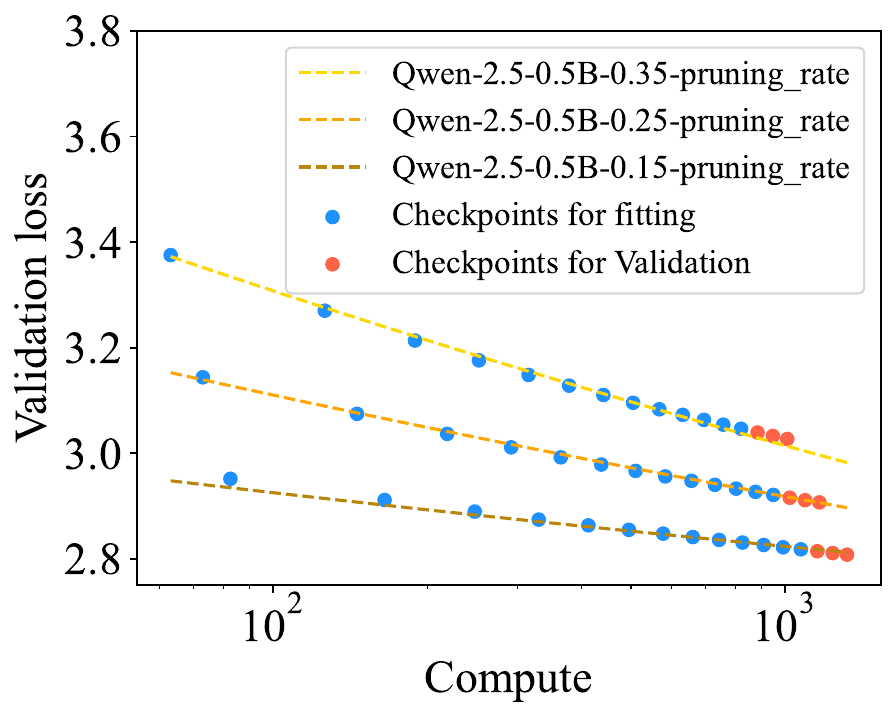}
            \caption{Loss curves fitted with the P$^2$ Law using the first 80\% of checkpoints; the remaining 20\% are used for validation. (Qwen-2.5-0.5B pruned by width pruning)}
        \end{subfigure}
        \hfill
        \begin{subfigure}[b]{0.3\linewidth}
            \includegraphics[width=\linewidth]{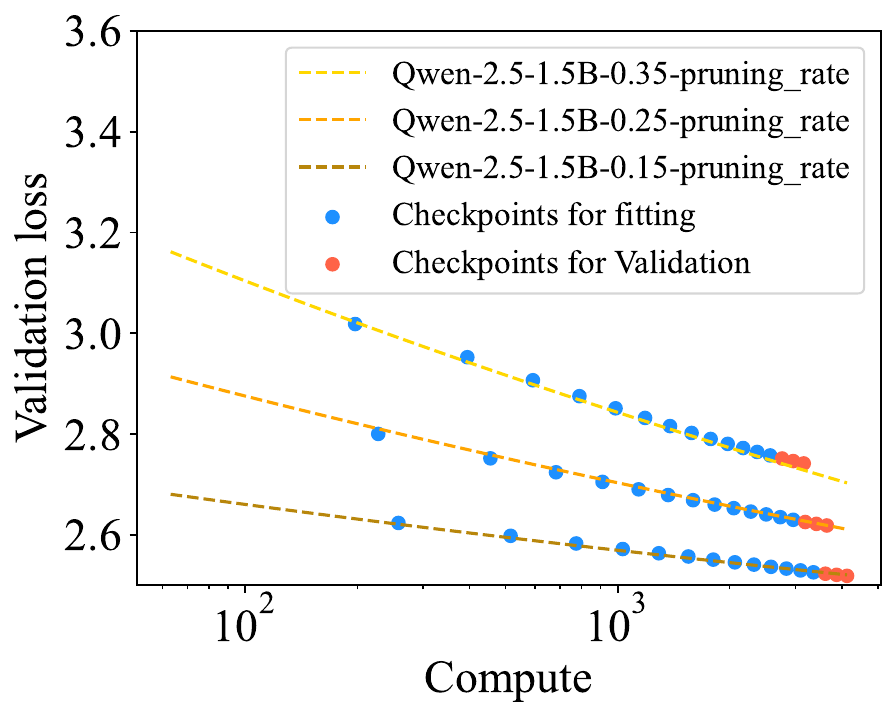}
            \caption{Loss curves fitted with the P$^2$ Law using the first 80\% of checkpoints; the remaining 20\% are used for validation. (Qwen-2.5-1.5B pruned by width pruning)}
        \end{subfigure}
        \hfill
        \begin{subfigure}[b]{0.301\linewidth}
            \includegraphics[width=\linewidth]{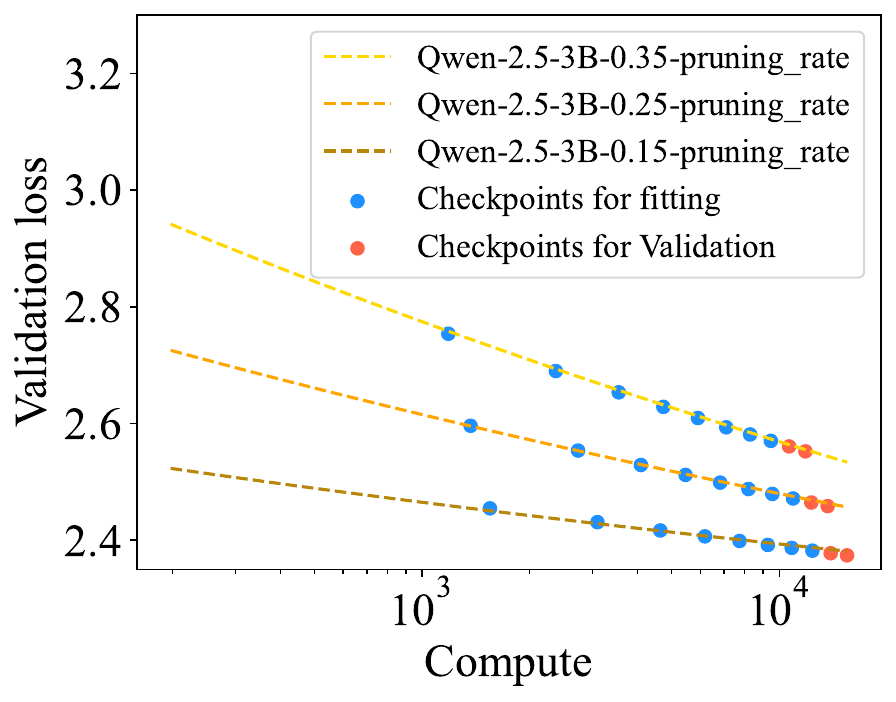}
            \caption{Loss curves fitted with the P$^2$ Law using the first 80\% of checkpoints; the remaining 20\% are used for validation. (Qwen-2.5-3B pruned by width pruning)}
        \end{subfigure}
    \end{minipage}
    \caption{Generalization of the P$^2$ Law for Qwen-2.5 series models pruned by width pruning on dataset size.}
    \label{fig:qwen_width_data}
\end{figure*}

\begin{figure*}[t]
    \centering
    \begin{minipage}{\linewidth}
        \centering
         \begin{subfigure}[b]{0.45\linewidth}
            \includegraphics[width=\linewidth]{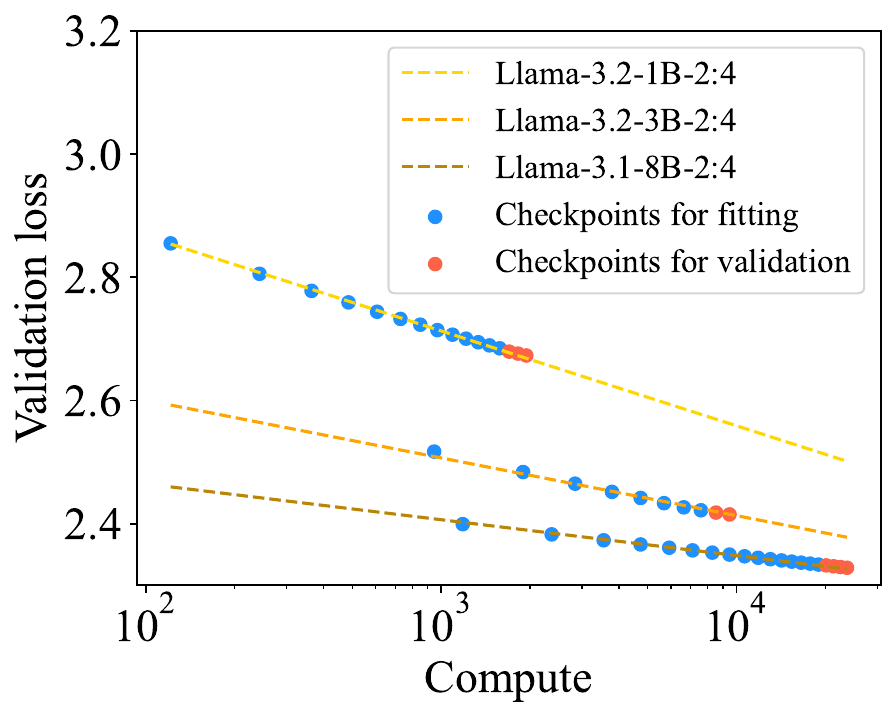}
            \caption{Loss curves fitted with the P$^2$ Law using the first 80\% of checkpoints; the remaining 20\% are used for validation. (Llama-3 series models pruned by 2:4 semi-structured pruning)}
        \end{subfigure}
        \hfill
        \begin{subfigure}[b]{0.45\linewidth}
            \includegraphics[width=\linewidth]{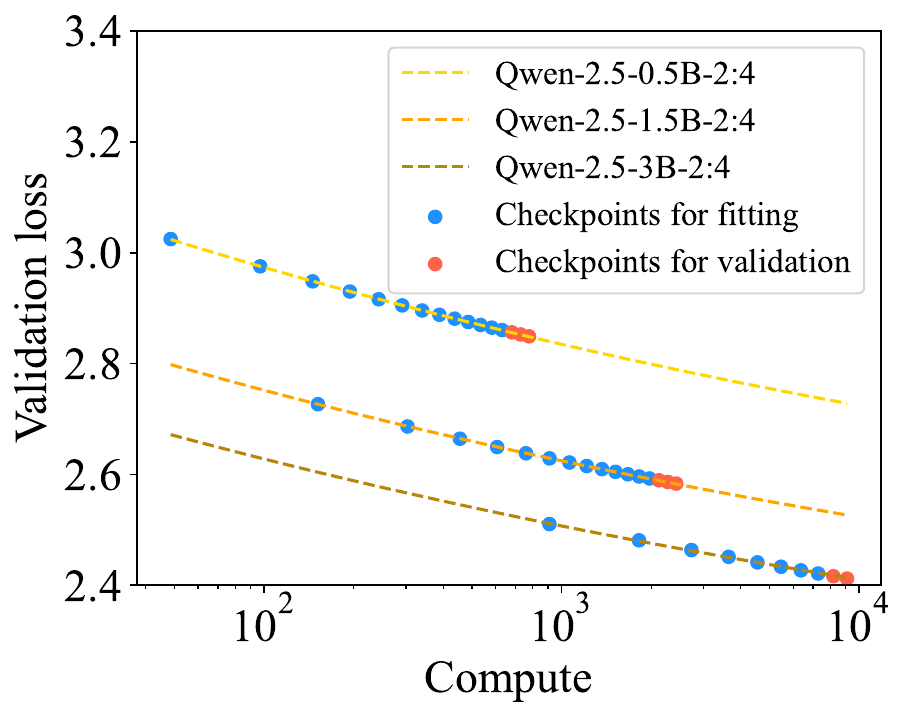}
            \caption{Loss curves fitted with the P$^2$ Law using the first 80\% of checkpoints; the remaining 20\% are used for validation. (Qwen-2.5 series models pruned by 2:4 semi-structured pruning)}
        \end{subfigure}
    \end{minipage}
    \caption{Generalization of the P$^2$ Law for models pruned by 2:4 semi-structured pruning on dataset size.}
    \label{fig:semi_data}
\end{figure*}

\begin{figure*}[t]
    \centering
    \begin{minipage}{\linewidth}
        \centering
        \begin{subfigure}[b]{0.3\linewidth}
            \includegraphics[width=\linewidth]{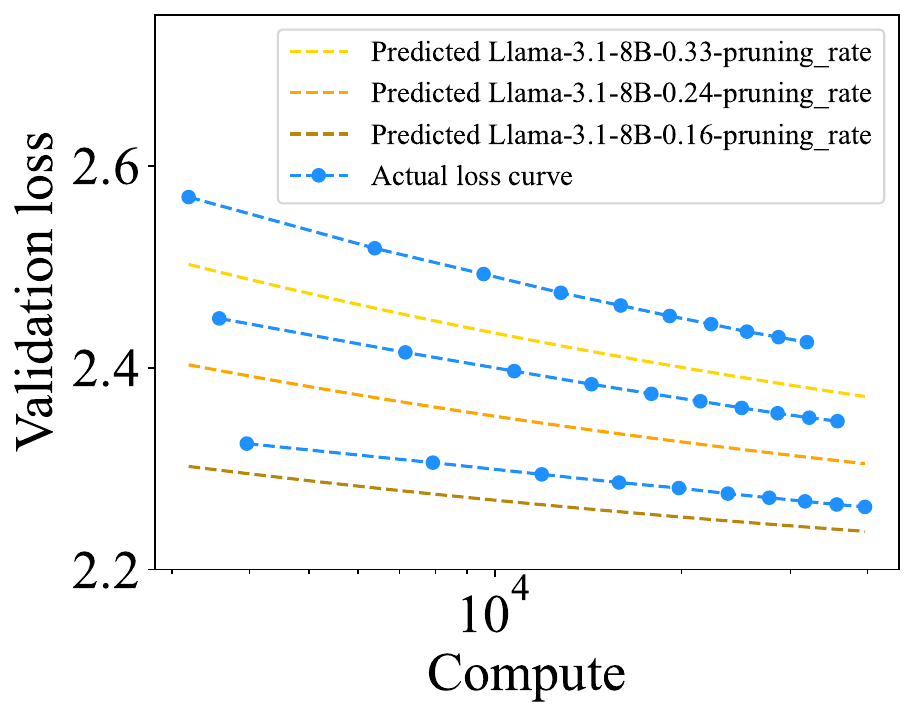}
            \caption{P$^2$ Law is fitted using checkpoints from smaller LLMs and used to predict the loss curves of larger LLMs. (Llama-3 series models pruned by depth pruning)}
        \end{subfigure}
        \hfill
        \begin{subfigure}[b]{0.3\linewidth}
            \includegraphics[width=\linewidth]{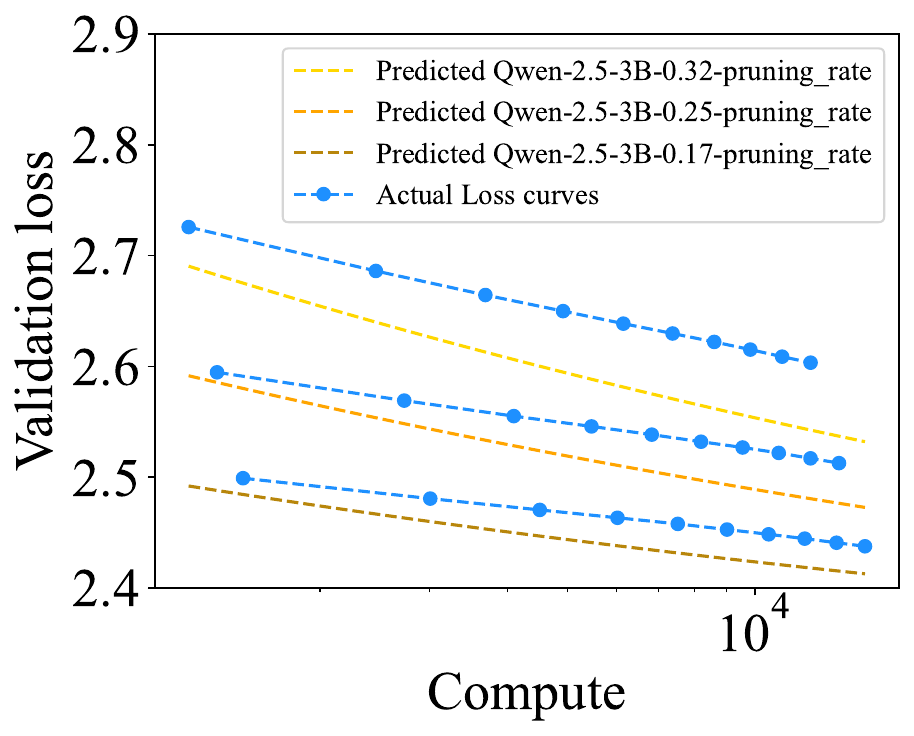}
            \caption{P$^2$ Law is fitted using checkpoints from smaller LLMs and used to predict the loss curves of larger LLMs. (Qwen-2.5 series models pruned by depth pruning)}
        \end{subfigure}
        \hfill
        \begin{subfigure}[b]{0.301\linewidth}
            \includegraphics[width=\linewidth]{new_Figures/qwen_width_model.pdf}
            \caption{P$^2$ Law is fitted using checkpoints from smaller LLMs and used to predict the loss curves of larger LLMs. (Qwen-2.5 series models pruned by width pruning)}
        \end{subfigure}
    \end{minipage}
    \caption{Generalization of the P$^2$ Law on model size.}
    \label{fig:model_size}
\end{figure*}

\begin{figure*}[t]
    \centering
    \begin{minipage}{\linewidth}
        \centering
        \begin{subfigure}[b]{0.45\linewidth}
            \includegraphics[width=\linewidth]{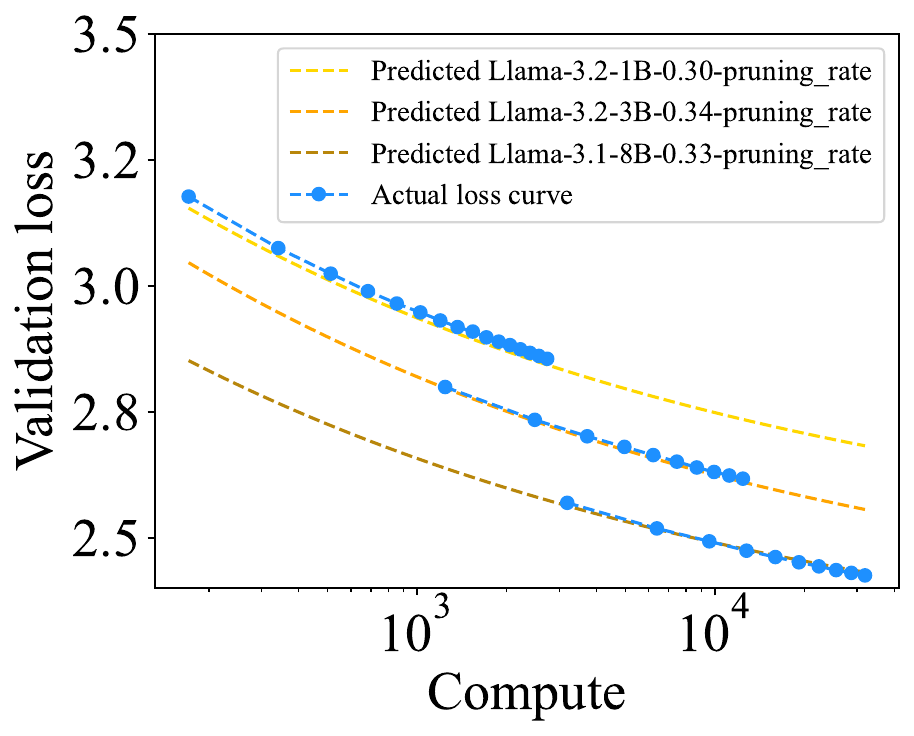}
            \caption{P$^2$ Law is fitted using checkpoints from smaller pruning rates and used to predict the loss curves of larger ones. (Llama-3 series models pruned by depth pruning)}
        \end{subfigure}
        \hfill
        \begin{subfigure}[b]{0.45\linewidth}
            \includegraphics[width=\linewidth]{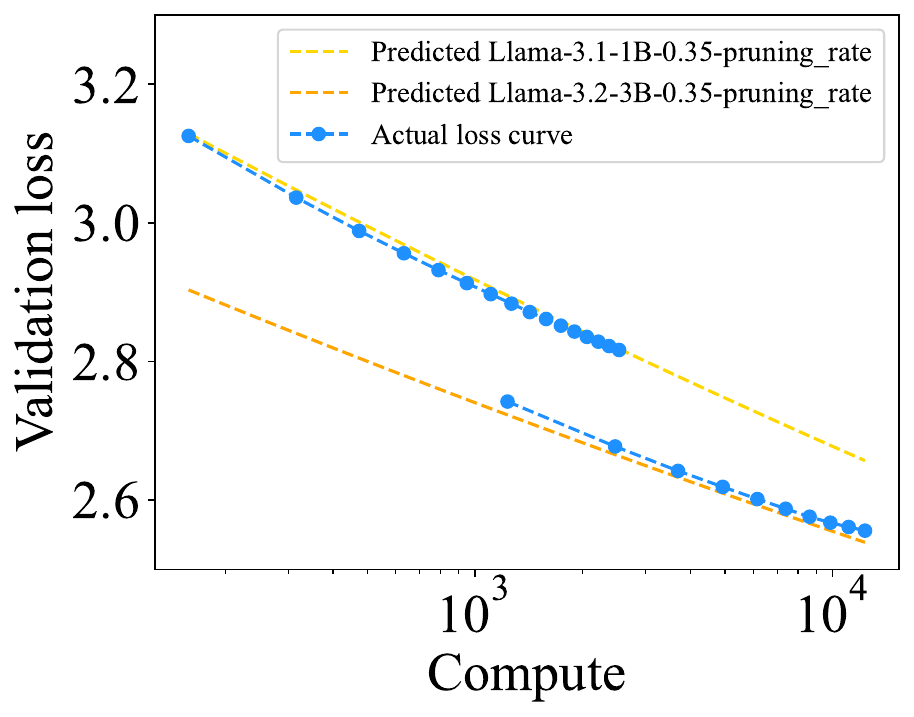}
            \caption{P$^2$ Law is fitted using checkpoints from smaller pruning rates and used to predict the loss curves of larger ones. (Llama-3 series models pruned by width pruning)}
        \end{subfigure}
    \end{minipage}
    \caption{Generalization of the P$^2$ Law for Llama-3 series models on pruning rate.}
    \label{fig:pruning_rate_llama}
\end{figure*}

\begin{figure*}[t]
    \centering
    \begin{minipage}{\linewidth}
        \centering
        \begin{subfigure}[b]{0.45\linewidth}
            \includegraphics[width=\linewidth]{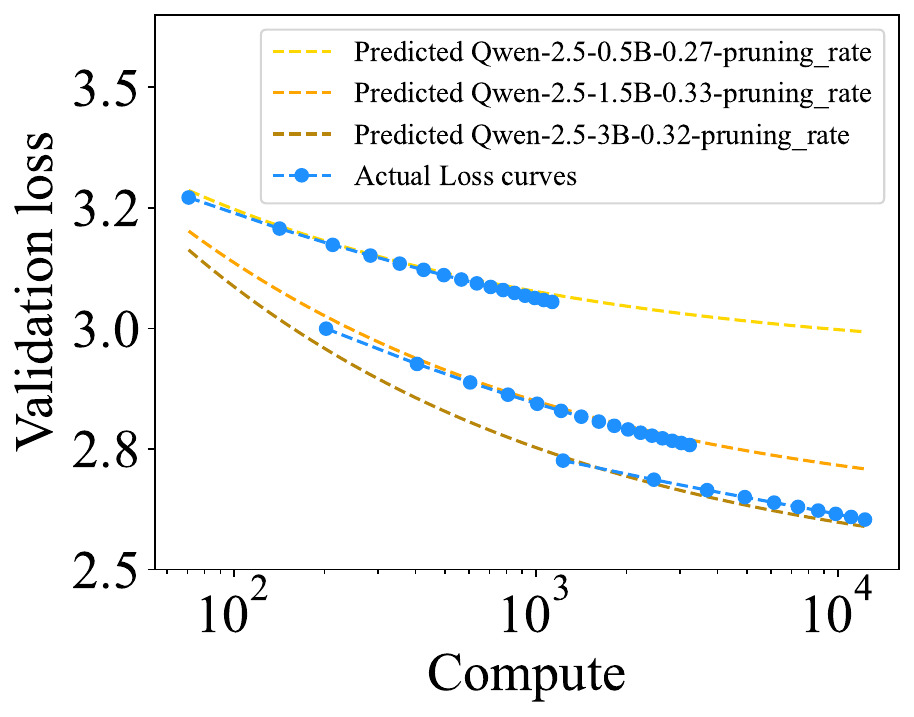}
            \caption{P$^2$ Law is fitted using checkpoints from smaller pruning rates and used to predict the loss curves of larger ones. (Qwen-2.5 series models pruned by depth pruning)}
        \end{subfigure}
        \hfill
        \begin{subfigure}[b]{0.45\linewidth}
            \includegraphics[width=\linewidth]{new_Figures/qwen_width_rate.pdf}
            \caption{P$^2$ Law is fitted using checkpoints from smaller pruning rates and used to predict the loss curves of larger ones. (Qwen-2.5 series models pruned by width pruning)}
        \end{subfigure}
    \end{minipage}
    \caption{Generalization of the P$^2$ Law for Qwen-2.5 series models on pruning rate.}
    \label{fig:pruning_rate_qwen}
\end{figure*}

We present the additional dataset size generalization predicted loss curves in the Figure~\ref{fig:llama_depth_data},~\ref{fig:llama_width_data},~\ref{fig:qwen_depth_data},~\ref{fig:qwen_width_data} and~\ref{fig:semi_data}, model size generalization predicted loss curves in the Figure~\ref{fig:model_size} and pruning rate generalization predicted loss curves in the Figure~\ref{fig:pruning_rate_llama} and~\ref{fig:pruning_rate_qwen}.

\begin{table*}[t]
\centering
\renewcommand\arraystretch{0.8}
\resizebox{\textwidth}{!}{
\begin{tabular}{c|c|ccc|ccc|ccc}
\toprule
 \multirow{2}{*}{LLM} & \multirow{2}{*}{Parameterizations}   & \multicolumn{3}{c|}{Depth pruning}      & \multicolumn{3}{c|}{Width pruning}      & \multicolumn{3}{c}{2:4 semi-structured pruning}        \\
 &    & $R^2$     & Huber loss & ASD      & $R^2$     & Huber loss & ASD      & $R^2$     & Huber loss & ASD      \\
 \midrule
\multirow{3}{*}{Llama-3 series}  
 & $\mathcal{L}_1$ & \textbf{0.9717} & \textbf{0.000016} & \textbf{0.000619} & \textbf{-1.2985} & \textbf{0.000177} & \textbf{0.000592} & \textbf{0.8126} & \textbf{0.000056} & \textbf{0.001466} \\
 & $\mathcal{L}_4$ &0.9339&0.000035&0.001482&-1.3660&0.000203&0.000814&0.7157&0.000112&0.002117 \\
 & $\mathcal{L}_5$ &0.6535&0.000198&0.001822&-1.7948&0.000345&0.000729&-0.3809&0.000638&0.003687 \\
 \midrule
\multirow{3}{*}{Qwen-2.5 series} 
 & $\mathcal{L}_1$ & \textbf{0.9781} & \textbf{0.000011} & \textbf{0.000524} & \textbf{0.9891} & \textbf{0.000010} & \textbf{0.000648} & \textbf{0.9995} & \textbf{0.000000} & \textbf{0.000191} \\
 & $\mathcal{L}_4$ &0.7730&0.000192&0.004085&0.9838&0.000015&0.001126&0.9960&0.000002&0.000550 \\
 & $\mathcal{L}_5$ &0.8283&0.000134&0.002648&0.9694&0.000040&0.001007&0.8360&0.000118&0.003925 \\
 \bottomrule
\end{tabular}
}
\caption{Comparison of law fitting results between OpenAI scaling law and Chinchilla scaling law.}
\label{tab:openai scaling law comparison}
\end{table*}

\begin{table*}[t]
\centering
\renewcommand\arraystretch{0.8}
\resizebox{\textwidth}{!}{
\begin{tabular}{c|c|ccccccc|ccccccc|cccccc}
\toprule
\multirow{2}{*}{LLM} & \multirow{2}{*}{Parameterizations} & \multicolumn{7}{c|}{Depth pruning} & \multicolumn{7}{c|}{Width pruning} & \multicolumn{6}{c}{2:4 semi-structured pruning} \\ 
& &$N_c$ &$D_c$ &$E$ &$\alpha$ &$\beta$ &$\gamma$ &$\delta$ & $N_c$ &$D_c$ &$E$ &$\alpha$ &$\beta$ &$\gamma$ &$\delta$ &$N_c$ &$D_c$ &$E$ &$\alpha$ &$\beta$ &$\delta$ \\
\midrule
\multirow{3}{*}{Llama-3 series} 
& $\mathcal{L}_1$ &0.02&5.94&0.14&-1.57&0.23&-1.08&0.29&0.05&5.86&-2.52&-1.68&0.08&-0.97&0.38&38.26&0.87&2.49&26.53&0.37&0.05 \\
& $\mathcal{L}_2$ &0.64&7.99&0.73&2.45&0.47&-1.08&-&0.00&3.53&0.20&-21.89&0.25&-0.97&-&0.53&0.89&2.19&0.92&0.41&- \\
& $\mathcal{L}_3$ &
-&5.93&0.54&-&0.30&-1.06&0.15&-&3.87&0.53&-&0.34&-0.98&-0.05&-&0.80&2.5&-&0.22&0.07 \\
\midrule
\multirow{3}{*}{Qwen-2.5 series} 
& $\mathcal{L}_1$ &0.01&4.32&0.20&-3.73&0.21&-1.17&0.22&-0.58&7.01&-1.89&0.38&0.10&-1.28&0.16&1.85&0.93&0.32&-0.12&0.10&0.17 \\
& $\mathcal{L}_2$ &0.02&4.78&0.62&4.08&0.32&-1.17&-&-0.01&5.84&-0.65&-1.58&0.18&-1.28&-&1.52&0.75&0.92&0.15&0.18&- \\
& $\mathcal{L}_3$ &-&4.77&0.87&-&0.36&-1.15&0.16&-&5.95&-0.91&-&0.16&-1.28&0.02&-&0.76&2.41&-&0.16&0.09 \\
\bottomrule
  
\end{tabular}
}
\caption{Parameter values of fitted parameterizations for P$^2$ Law fitting.}
\label{tab:fit parameter}
\end{table*}

\begin{table*}[t]
\centering
\renewcommand\arraystretch{0.8}
\resizebox{\textwidth}{!}{%
\begin{tabular}{c|c|c|c|c}
\toprule
LLM & Parameterizations & Depth pruning & Width pruning & 2:4 semi-structured pruning \\ 
\midrule
\multirow{3}{*}{Llama-3 series} 
& $\mathcal{L}_1$ &\checkmark&\checkmark&\checkmark \\
& $\mathcal{L}_2$ &$\times$&$\times$&$\times$  \\
& $\mathcal{L}_3$ &\checkmark&$\times$&\checkmark \\
\midrule
\multirow{3}{*}{Qwen-2.5 series} 
& $\mathcal{L}_1$ &\checkmark&\checkmark&\checkmark \\
& $\mathcal{L}_2$ &$\times$&$\times$&$\times$ \\
& $\mathcal{L}_3$&\checkmark&\checkmark&\checkmark \\
\bottomrule
\end{tabular}%
}
\caption{Compliance of $\mathcal{L}_1$, $\mathcal{L}_2$, and $\mathcal{L}_3$ with Condition 2.}
\label{tab:condition 2}
\end{table*}

\end{document}